\documentclass[sn-aps]{sn-jnl}

\usepackage{graphicx}%
\usepackage{multirow}%
\usepackage{amsmath,amssymb,amsfonts}%
\usepackage{amsthm}%
\usepackage{mathrsfs}%
\usepackage[title]{appendix}%
\usepackage[usenames,dvipsnames]{xcolor}
\usepackage{textcomp}%
\usepackage{manyfoot}%
\usepackage{booktabs}%
\usepackage{algorithm}%
\usepackage{algorithmicx}%
\usepackage{algpseudocode}%
\usepackage{listings}%
\usepackage{comment}
\usepackage{ulem}
\usepackage{CJKutf8}
\usepackage{tabularx}
\usepackage{siunitx}
\usepackage{docmute}

\graphicspath{{Figs_New/}{Supp/Figs/}}

\newcommand{\cn}{Conditioner}
\newcommand{\pn}{Predictor}
\newcommand{\tds}{trajectory-decoupled sampling}

\begin{document}

\title[Article Title]{Neural operator discovery from heterogeneous trajectories}

\author[1]{\fnm{Zituo} \sur{Chen}}%\email{iiauthor@gmail.com}
\equalcont{These authors contributed equally to this work.}
\author[1,2]{\fnm{Qiaofeng} \sur{Li}}%\email{iauthor@gmail.com}
\equalcont{These authors contributed equally to this work.}
\author[2]{\fnm{Jiaxin} \sur{Hu}}%\email{iiiauthor@gmail.com}
\author*[1]{\fnm{Sili} \sur{Deng}}\email{silideng@mit.edu}

\affil[1]{\orgdiv{Department of Mechanical Engineering}, \orgname{Massachusetts Institute of Technology}, \orgaddress{\street{77 Massachusetts Avenue}, \city{Cambridge}, \postcode{02139}, \state{MA}, \country{USA}}}
\affil[2]{\orgdiv{State Key Laboratory of Fluid Power \& Mechatronic Systems}, \orgname{Zhejiang University}, \orgaddress{\street{866 Yuhangtang Road}, \city{Hangzhou}, \postcode{310058}, \state{Zhejiang Province}, \country{China}}}

\abstract{

Neural operators provide data-driven mappings for modeling dynamical systems. Extending them to families of systems typically requires explicit conditioning variables such as physical parameters, geometries, or boundary conditions. In many real-world settings, these quantities are unobserved. Here, we formulate neural operator discovery (NOD) as the problem of learning both shared solution operators and system-specific variation directly from heterogeneous trajectories without access to labeled governing factors. We introduce a factorized latent-conditioning formulation that jointly learns a neural operator and a low-dimensional latent representation through factorized prediction, trajectory-decoupled sampling, and dimension selection. Across diverse systems, the learned latent representation captures the intrinsic dimensionality of system variation and organizes system instances in a smooth and approximately invertible latent structure aligned with the underlying governing factors. This organization enables generalization to previously unseen system instances, including zero-shot extrapolation across regimes and stable long-horizon prediction. These results establish an interpretable paradigm for operator learning in the absence of explicit factor supervision.

}


\maketitle

Across the physical sciences, trajectory data are often collected from systems that share a common governing law but differ across experimental conditions, geometries, or boundary settings. In many experimental datasets and legacy simulation archives, the factors that distinguish individual system instances are unobserved, unrecorded, or not directly measurable. The resulting data therefore consist of heterogeneous trajectories generated by multiple underlying configurations without explicit annotation of the variables driving their variation. A central challenge is to learn predictive models that capture both the shared dynamics of a system family and the differences between system instances directly from trajectory data, without access to labeled governing factors.

%\zc{One established approach to this challenge is symbolic regression, which recovers governing equations together with per-instance coefficients directly from trajectory data without requiring factor labels~\cite{brunton2016discovering,jacobs2024hypersindy,dascoli2024odeformer}. Expressivity is, however, bounded by the chosen primitive library, and inference cost grows sharply with system complexity (Supplementary Section~S3.6), in practice restricting symbolic regression to low-dimensional systems and ruling out discovery for the high-dimensional spatiotemporal dynamics governed by partial differential equations.}
One established approach to this challenge is symbolic regression, which seeks to recover governing equations together with system-specific coefficients directly from trajectory data without requiring factor labels~\cite{brunton2016discovering,jacobs2024hypersindy,dascoli2024odeformer}. However, its expressivity is constrained by the chosen primitive library, while computational cost grows rapidly with system complexity (Supplementary Section~S3.6). In practice, these limitations restrict symbolic regression largely to low-dimensional systems and hinder its application to high-dimensional spatiotemporal dynamics governed by partial differential equations.

Neural operator learning, by contrast, provides fast and accurate data-driven surrogates for such spatiotemporal dynamics~\cite{kovachki2023neural,azizzadenesheli2024neural,li2020fourier,lu2021learning}. Prevailing formulations either model a single dynamical system or, when extended to families, learn the solution operator in a supervised manner with factors as explicit inputs (\textit{parametric neural operators}, PNO)~\cite{zhou2024unisolver,sun2025towards,wu2024transolver,fang2024learning}. This assumption is violated in the setting considered here, where the factors distinguishing system instances are unobserved.%The PNO assumption is violated in our problem.

%Partial solutions have been proposed. 
Alternative approaches partially address this limitation. Foundation-model-style surrogates (\textit{autoregressive neural operators}, ANO)~\cite{cao2024vicon,hao2024dpot,chen2025flow,mccabe2023multiple} avoid explicit parameter supervision by conditioning prediction on long histories of observed states, referred to here as the \textit{past-state window}. However, these representations are high-dimensional and unstructured with respect to the underlying governing factors. As a result, autoregressive conditioning accumulates rollout error during self-prediction~\cite{arora2023exposurebiasmattersimitation,chen2026latentgenerativesolversgeneralizable} and does not yield an identifiable representation of system variation.
%by conditioning each prediction on a long window of observed past states, the \textit{past-state window}. However, this window is high-dimensional and unstructured around factors. ANO conditioning therefore accumulates error during self-rollouts~\cite{arora2023exposurebiasmattersimitation,chen2026latentgenerativesolversgeneralizable}, and recovers no identifiable system variation.

A separate line of work infers latent representations from trajectory data and uses them to condition prediction. Existing approaches, however, trade off expressivity, scalability, or interpretability. Variational autoencoders compress low-complexity dynamics into latent representation under Gaussian priors~\cite{Lu_2020}, restricting the capacity of the learned solution operator. Gradient-based meta-learning methods recover system descriptors through inner-loop adaptation~\cite{li2023metalearning}, but backpropagation through the inner loop becomes prohibitively expensive for high-dimensional spatiotemporal systems.

We observe that both the past-state window used in ANO and the inferred latent representation used in latent-conditioned models serve the same role: encoding the system-specific variation underlying a shared solution operator. Motivated by this perspective, we formulate the problem of learning both a shared solution operator and system-specific variation from heterogeneous trajectories without parameter supervision as \textbf{Neural Operator Discovery} (NOD).

\begin{figure*}[htbp]
    \centering
    \includegraphics[width=\linewidth]{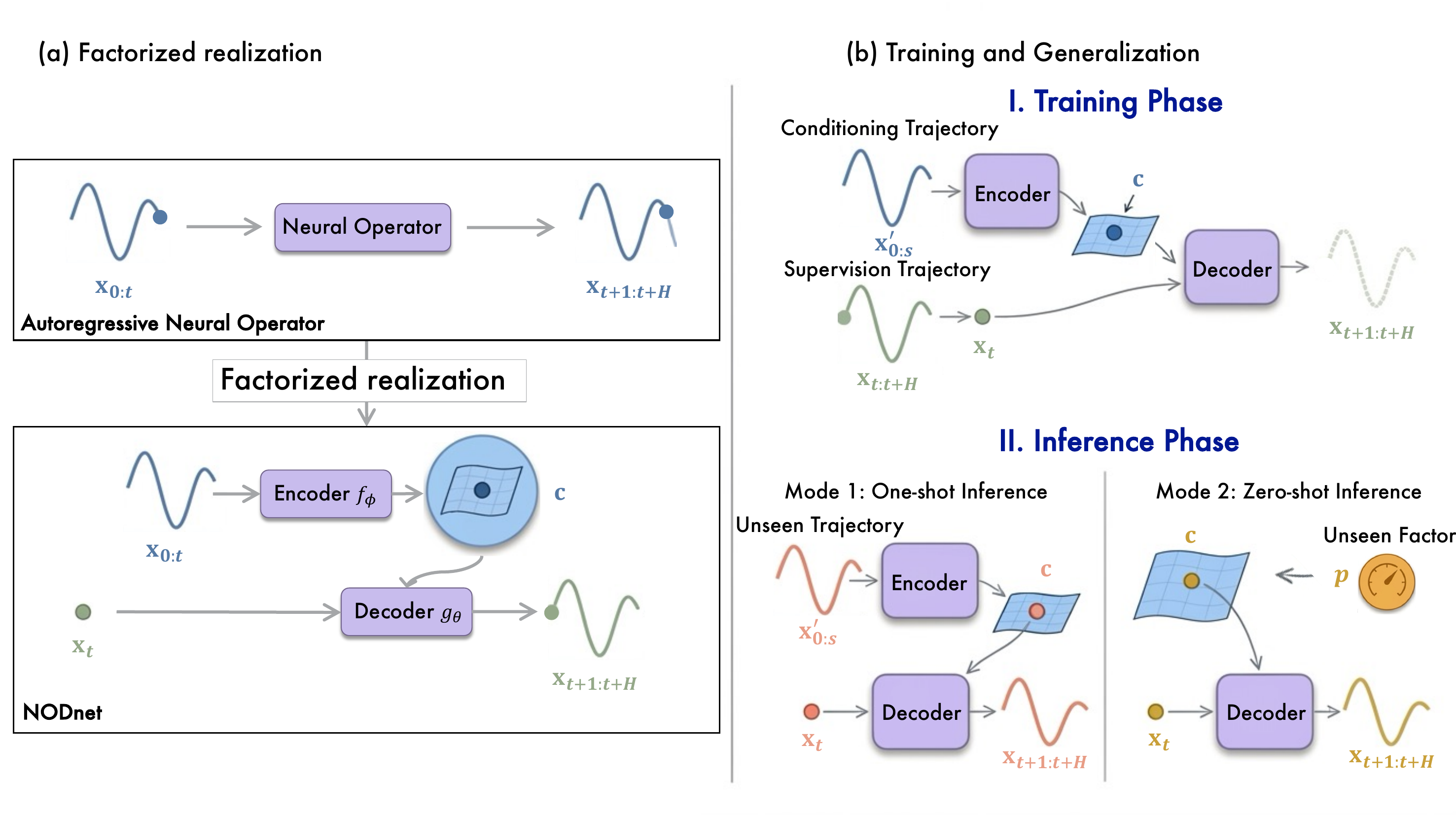}
    \caption{\textbf{Neural Operator Discovery.} (a) Comparison between conventional history-conditioned neural operators and Neural Operator Discovery. Conventional autoregressive neural operators condition prediction directly on a past-state trajectory window $\mathbf{x}_{0:t}$. In NODnet, the trajectory window is first mapped to a low-dimensional latent representation $\mathbf{c}$ through an encoder $f_\phi$, and a conditional decoder $g_\theta$ then predicts future dynamics from initial state $\mathbf{x}_{t}$ conditioned on $\mathbf{c}$. (b) Training and inference with NODnet. During training, trajectory-decoupled sampling pairs two trajectories $\mathbf{x}'_{0:s}$ and $\mathbf{x}_{t:t+H}$ from the same system instance but different initial conditions, using one trajectory for latent inference and the other for supervision. During inference, prediction can proceed either by inferring the latent representation from an observed trajectory (one-shot inference) or by directly sampling the latent space (zero-shot inference). Both modes generalize to system instances unseen during training, denoted as ``unseen trajectory'' in one-shot inference and ``unseen factor'' in zero-shot inference. }
    \label{fig:algo}
\end{figure*}

To address this problem, we consider a factorized realization consisting of two components (Fig.~\ref{fig:algo}a): a trajectory encoder $f_\phi$, which maps a past-state window from a system instance to a low-dimensional latent representation $\mathbf{c}$, and a state-evolution operator $g_\theta$, which predicts future states conditioned on this representation. Two complementary constraints are imposed on $\mathbf{c}$. First, the representation is restricted to be low-dimensional, with its dimension selected as the smallest value sufficient to capture variation across system instances. Second, the conditioning trajectory and prediction trajectory are sampled from different initial conditions of the same system instance (Fig.~\ref{fig:algo}b), encouraging the encoder to retain only cross-trajectory invariants. Together, these constraints promote latent representations that capture governing factors while suppressing trajectory-specific features.

Under this formulation, the learned representations organize system instances in a structured latent space $\mathcal{C}$ that reflects the underlying physical variation. Across multiple systems, this latent space exhibits a smooth and approximately invertible correspondence with the true governing factor space $\mathcal{P}$, despite being learned without parameter supervision. The resulting model supports both accurate long-horizon prediction and generalization to previously unseen system instances. In particular, new systems can be handled either by inferring the latent representation from a single observed trajectory (one-shot inference) or by sampling directly in the latent space to obtain the corresponding latent representation $\mathbf{c}$ (zero-shot inference), as illustrated in Fig.~\ref{fig:algo}b.

\section*{Results}

This section first illustrates the NODnet structure, then validates it on three families of partial differential equations: the Burgers' equation with varying viscosity, which tests the capture of sharp, regime-dependent dynamics under shocks of $\mathcal{O}(\nu)$ width; the cylinder flow wake on an unstructured mesh, which tests scalability under sparse supervision; and the FitzHugh--Nagumo reaction--diffusion system with two independent kinetic factors, which tests recovery of multi-factor variation. Four additional case studies in the Supplementary Information extend the evaluation to geometric, boundary-condition, and resolution variation, spanning convolutional, recurrent, Fourier Neural Operator, and Neural-ODE implementations of $f_\phi$ and $g_\theta$ (Supplementary Table~S1).

\subsection*{Method overview}

We consider families of dynamical systems governed by a shared equation form but differing in system-specific factors such as physical coefficients, boundary conditions, or geometric parameters. Let $p_i \in \mathcal{P}$ denote the factors associated with system instance $i$, where $\mathcal{P}\subset\mathbb{R}^d$ is the underlying factor space. The dynamics evolve according to an operator
\begin{equation}
\frac{\partial \mathbf{x}}{\partial t} = \mathcal{G}(\mathbf{x}_t; p_i),
\end{equation}
where the governing operator $\mathcal{G}$ and the factors $p_i$ are both unknown.

Each system instance generates trajectories $\tau=\mathbf{x}_{0:T}$ on the computation domain $\Omega$ with a time window $[0,T]$, yielding a heterogeneous dataset $\mathcal{D}={\tau_{ij}}$, where $i$ denotes system instance, $j$ denotes trajectory initial state $\mathbf{x}_0$. In the setting considered here, only the trajectories and their grouping by system instance are available; the governing factors themselves are unobserved. Importantly, the grouping information does not directly specify the underlying factors. For example, a two-dimensional factor space sampled at five values along each dimension produces 25 distinct system instances, while the observed data contain only trajectories grouped by instance rather than explicit factor coordinates.

Our goal is to learn both a shared solution operator in discrete form,
\begin{equation}
\mathbf{x}_{t+1} = \mathcal{G}'(\mathbf{x}_t; p_i),
\end{equation}
and a low-dimensional latent representation that captures the underlying system-specific variation. To achieve this, our formulation combines three elements: factorized prediction, trajectory-decoupled sampling, and dimension selection.

\paragraph*{Factorized prediction.} 

Conventional autoregressive neural operators predict future states directly from a window of past observations:
\begin{equation}
\mathbf{x}_{t+1} = \mathcal{G}'_{\psi}(\mathbf{x}_t;\mathbf{x}_{0:t}),
\end{equation}
where the past-state window $\mathbf{x}_{0:t}$ implicitly carries both dynamical information and system-specific variation. In NODnet, we instead factorize prediction into two components: a latent representation inferred from past trajectories, and a conditional state-evolution operator:
\begin{equation}
\mathbf{c} = f_\phi(\mathbf{x}_{0:t}),
\hat{\mathbf{x}}_{t+1} = g_\theta(\mathbf{x}_t;\mathbf{c}).
\label{eq:decompose}
\end{equation}
Here, the encoder $f_\phi$ maps a trajectory segment to a low-dimensional latent representation $\mathbf{c}$ that captures system-specific variation, while the decoder $g_\theta$ predicts future states conditioned on this representation. Training minimizes mean-squared error of prediction:
\begin{equation}
\mathcal{L} = \|\hat{\mathbf{x}}_{t+1}-\mathbf{x}_{t+1}\|_2.
\end{equation}

\paragraph*{Trajectory-decoupled sampling.} 

To prevent the latent representation from encoding trajectory-specific initial-condition information, we train the encoder and decoder using trajectories generated from different initial conditions of the same system instance. Specifically, the encoder infers a latent representation
\begin{equation}
\mathbf{c}=f_\phi(\mathbf{x}'_{0:s}),
\end{equation}
from a conditioning trajectory $\mathbf{x}'_{0:s}$, while the decoder predicts the evolution of a separate trajectory $\mathbf{x}_{t:t+1}$ from the same system instance:
\begin{equation}
\hat{\mathbf{x}}_{t+1}=g_\theta(\mathbf{x}_t;\mathbf{c}).
\end{equation}

Because the conditioning and prediction trajectories share governing factors but differ in initial conditions, the encoder is encouraged to retain only cross-trajectory invariants rather than transient trajectory-specific features. 

\paragraph*{Dimension selection.} To ensure that the latent representation captures only system-specific variation, we select its dimension as the smallest value sufficient to explain differences across system instances. %\zc{The selection follows a two-stage rule applied to a sweep over $d = 0, 1, 2, \dots$. Stage~1 identifies a candidate set from the training-loss curve as the smallest dimension at which the step to $d{+}1$ no longer yields a clear improvement relative to the across-seed envelope, together with the dimension immediately above. Stage~2 selects $d^*$ as the smallest candidate whose $d{+}1$ run yields no new factor-aligned structure in the learned latent space, verified by principal component analysis on the per-instance latent means.} 
The selection follows a two-stage procedure over a sweep of representation dimensions $d = 0, 1, 2, \dots$. In the first stage, candidate dimensions are identified from the training-loss curve as the smallest dimensions beyond which increasing $d$ no longer yields improvements exceeding the variability across random seeds. In the second stage, the final dimension $d^*$ is selected as the smallest candidate whose corresponding $d{+}1$ model does not produce additional factor-aligned structure in the learned latent space, assessed using principal component analysis of the per-instance latent representations. Applied to the three main case studies, this procedure yields $d^*=(1,1,2)$, matching the number of independent governing factors in each system family. Full details, threshold criteria, seed configurations, and latent-space analysis are provided in Supplementary Sections~S2.1--S2.3.

\subsection*{Evaluation overview}

\paragraph*{Inference modes of NODnet.} Once trained, the learned latent space organizes system instances according to their underlying physical variation, enabling two modes of inference. In zero-shot inference, a latent representation is sampled directly from the learned latent space and used to condition the decoder for prediction. In one-shot inference, the latent representation is first inferred from a short trajectory segment using the encoder, after which prediction proceeds using the same conditional decoder.

For Burgers' equation and cylinder flow wake, we evaluate one-shot inference, where the latent representation must be inferred from observed trajectories. For the FitzHugh--Nagumo reaction--diffusion system, we evaluate both one-shot and zero-shot inference to examine the structure and generalization properties of the learned latent space.

\paragraph*{Baselines and fair comparison.}
We compare against two baselines that isolate different conditioning strategies while maintaining matched model capacity within each case study. The autoregressive baseline (ANO) conditions predictions directly on a window of past states, without learning an explicit latent representation of system instances. The single-system baseline (SNO) removes the encoder entirely, corresponding to a shared neural operator trained across all trajectories without distinguishing between instances. The predictive capacity of SNO is matched to the decoder component of our model, while ANO is matched to the full model capacity. These baselines isolate the contribution of learning a low-dimensional latent representation for system-specific variation, independent of architectural scaling. A parameter-supervised reference (PNO), in which the true governing factors are provided as input, is reported in Supplementary Section~S2.4 as a non-comparable upper-bound benchmark.

\subsection*{Burgers' Equation}

\begin{figure*}[htbp]
    \centering
    \includegraphics[width=\linewidth]{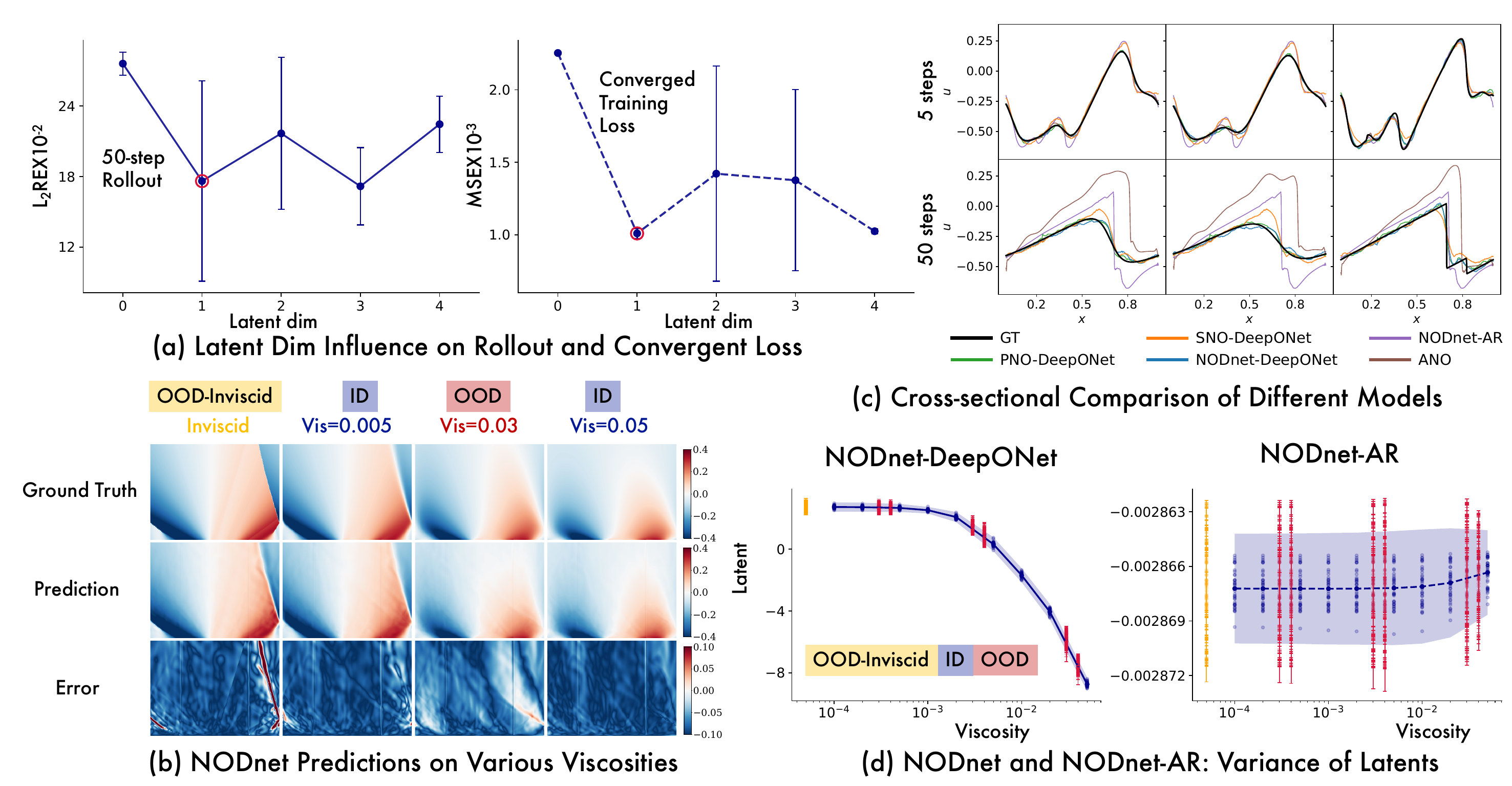}
   \caption{\textbf{Burgers' equation results.} (a) Dimension selection. The 50-step rollout error (left, $L_2$RE $\times 10^{-2}$) and converged training MSE (right, $\times 10^{-3}$) are evaluated across latent dimensions and three random seeds. The dimension-selection procedure identifies $d^*{=}1$, consistent with viscosity being the only varying governing factor (Supplementary Section~S2.1).  (b) Predictions across viscosity regimes, including the inviscid limit ($\nu{=}0$), in-distribution viscosities ($\nu{=}0.005, 0.05$), and an unseen viscosity within the training range ($\nu{=}0.03$). Rows show ground truth, prediction, and absolute error.  (c) Comparison of long-horizon prediction profiles for the parameter-supervised upper bound (PNO), the single-system baseline (SNO), the latent-conditioned model (NODnet), the autoregressive latent-conditioned variant (NODnet-AR), and the autoregressive baseline (ANO). Results are shown at rollout horizons $H{=}5$ (top) and $H{=}50$ (bottom). (d) Relationship between the learned latent representation and viscosity. The full-field predictor learns a smooth monotonic latent organization extending continuously to the inviscid limit, whereas the autoregressive variant collapses to a nearly constant representation and fails to separate viscosity regimes. }
    \label{fig:Burgers_Result}
\end{figure*}

As a one-dimensional special case of the Navier--Stokes equations, Burgers' equation describes the interplay between nonlinear convection and viscous diffusion. As viscosity decreases, solutions develop shock fronts of $\mathcal{O}(\nu)$ width, producing sharp regime-dependent dynamics that are difficult to resolve. The governing equation is
\begin{equation}
\frac{\partial u}{\partial t} + u\frac{\partial u}{\partial x} = \nu\frac{\partial^2u}{\partial x^2},
\end{equation}
where $u$ is the fluid velocity and $\nu$ is the kinematic viscosity, the only varying factor across the system family.

We evaluate generalization across three settings: unseen initial conditions at trained viscosities (ID), unseen viscosities within the training range (OOD), and the inviscid limit $\nu{=}0$, which tests extrapolation to a singular regime. Training data cover nine viscosities in the viscous regime.

Sweeping the latent representation dimension from $d{=}0$ to $4$ reveals an optimum at $d^*{=}1$, consistent with the fact that viscosity is the only varying factor across the system family (Fig.~\ref{fig:Burgers_Result}a). Predictions remain accurate across all three evaluation splits (Fig.~\ref{fig:Burgers_Result}b). At $\nu{=}0$, the ground truth corresponds to the entropy solution of the inviscid Burgers' equation~\cite{oleinik1957discontinuous,kruvzkov1970first}, which classical numerical integration typically reaches only by switching from parabolic to hyperbolic discretization schemes. In contrast, the learned latent representation varies smoothly and monotonically with viscosity, extending continuously to the inviscid regime (Fig.~\ref{fig:Burgers_Result}d, left). One-shot inference from a single unseen trajectory recovers a consistent latent representation for the inviscid system instance.

We next examine how decoder supervision influences factor identification. Under shock dynamics, most of the spatial domain remains smooth, while the shock front occupies only a small region whose position depends strongly on viscosity. We therefore compare two variants that share the same encoder but differ in prediction strategy. The first (NODnet-DeepONet) predicts the full spatiotemporal field in a single forward pass using a DeepONet-style decoder~\cite{lu2021learning}, receiving supervision across all $(x,t)$ locations simultaneously. The second (NODnet-AR) replaces the decoder with an autoregressive stepper trained only on one-step transitions. In the autoregressive setting, the training loss is dominated by smooth-region errors that vary weakly with viscosity, providing limited gradient signal for separating system instances. As a result, the autoregressive variant fails to distinguish viscosities and collapses to a nearly constant latent representation, whereas the full-field predictor learns a monotonic latent organization aligned with viscosity (Fig.~\ref{fig:Burgers_Result}d, Table~\ref{tab:performance}).

This effect is also reflected in long-horizon prediction performance. The autoregressive baseline achieves low one-step error but accumulates substantial rollout drift, producing the largest 50-step prediction error among all models (Fig.~\ref{fig:Burgers_Result}c, Table~\ref{tab:performance}). By contrast, the latent-conditioned full-field predictor maintains the lowest 50-step error across the ID, OOD, and inviscid splits, and remains within a bounded gap of the parameter-supervised reference reported in Supplementary Section~S2.4. The no-conditioning baseline converges to a family-averaged solution because it receives no information distinguishing one viscosity regime from another.

\subsection*{Cylinder Flow Wake}

The cylinder flow wake provides a challenging setting for evaluating Neural Operator Discovery under sparse and geometrically complex data. In the laminar Reynolds-number regime considered here ($\mathrm{Re} \in [75, 400]$), vortex shedding patterns and drag vary smoothly with viscosity. The governing equations are the 2D incompressible Navier--Stokes equations:
\begin{equation}
    \begin{split}
        \frac{\partial\mathbf{u}}{\partial t} + \left(\mathbf{u}\cdot\nabla\right)\mathbf{u} &= \nu \nabla^2\mathbf{u} - \nabla p/\rho\\
        \mathbf{u}\mid_{\partial\Omega_{in}} &= u_\infty
    \end{split}
    \label{ns}
\end{equation}
where $\nu$ is the kinematic viscosity, $u_\infty$ is the inlet velocity, and $\mathbf{u}=(u,v)$ is the velocity field. Across the system family, viscosity $\nu$ is the only varying governing factor, while variations in $u_\infty$ modify the initial flow development through the boundary condition.

This case probes scalability along two dimensions: the flow is defined on an unstructured mesh, and the training data are sparse, consisting of only 16 trajectories spanning four viscosity values and four inlet velocities. The ID split evaluates generalization to unseen $u_\infty$ at trained viscosities, while the OOD split evaluates extrapolation to viscosities outside the training range (Fig.~\ref{fig:Methods}II(c)).

\begin{figure*}[htbp]
    \centering
    \includegraphics[width=\linewidth]{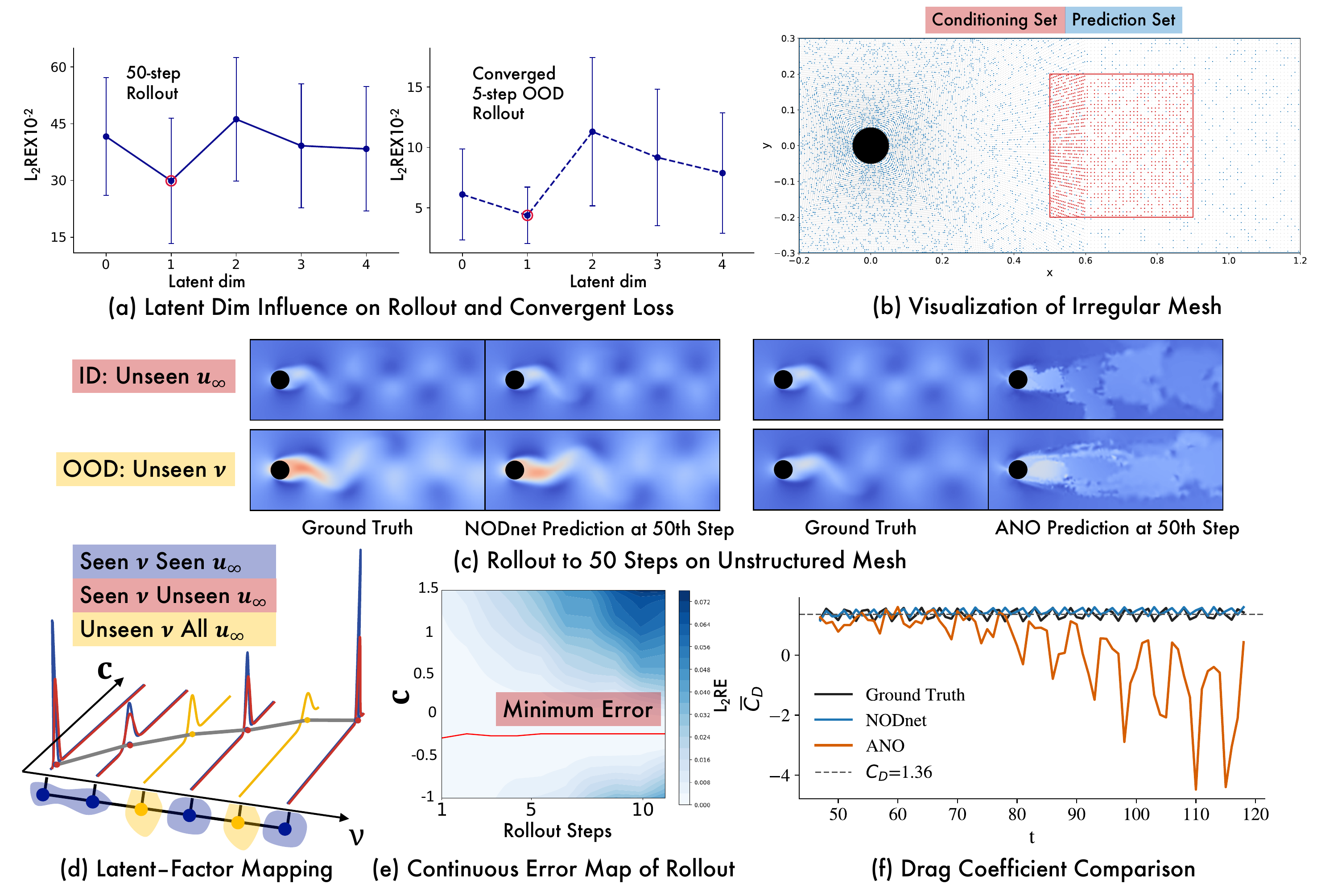}
    \caption{\textbf{Cylinder flow wake results.} (a) Dimension selection based on long-horizon training and OOD rollout errors identifies $d^*{=}1$ (Supplementary Section~S2.2).  (b) Irregular-mesh prediction setting. The encoder infers the latent representation from a conditioning set (red), while the decoder predicts flow evolution on a separate prediction set (blue).  (c) 50-step rollouts on the ID split (unseen $u_\infty$) and OOD split (unseen $\nu$). (d) Relationship between the inferred latent coordinate and viscosity. System instances organize along a smooth monotonic curve in latent space, with a convex loss landscape across prediction horizons. (e) Continuous rollout-error landscape as the latent representation value is perturbed. The error-minimizing latent value remains aligned with the inferred representation as rollout horizons increase. (f) Drag coefficient $C_D$ at $\mathrm{Re}{=}300$. The latent-conditioned model recovers the reference drag coefficient $\bar{C}_D{=}1.36$ and the correct vortex shedding frequency, whereas the autoregressive baseline exhibits frequency drift and develops nonphysical drag behavior. } 
    \label{fig:CFW_Result}
\end{figure*}

Sweeping the representation dimension from $d{=}0$ to $4$ reveals an optimum at $d^*{=}1$, consistent with viscosity being the only governing factor varying across the system family (Supplementary Section~S2.2). Larger representations degrade long-horizon rollout performance, indicating that additional dimensions capture trajectory-specific features rather than meaningful system variation. At $d{=}1$, the inferred representation varies monotonically with viscosity (Fig.~\ref{fig:CFW_Result}d), suggesting that the latent space organizes system instances according to the underlying physical structure.

To examine how the learned representation influences prediction, we perturb the inferred latent coordinate and evaluate the resulting rollout error. The error landscape remains smooth across rollout horizons, with prediction error minimized near the inferred latent representation (Fig.~\ref{fig:CFW_Result}e). On the OOD split, representations inferred from previously unseen trajectories remain consistent with the learned structure (Fig.~\ref{fig:CFW_Result}d), and one-shot inference achieves comparable performance. Together, these results suggest that the learned latent space generalizes smoothly beyond the training viscosity range.

To evaluate robustness under sparse and geometrically complex data, we train the model using only 16 trajectories defined on an unstructured mesh with 4,869 nodes per prediction step. Both the encoder and decoder are implemented using Point Transformer backbones~\cite{zhao2021point}, enabling latent conditioning and prediction directly on irregular geometries. Additional architectural details and ablation studies on encoder--decoder capacity allocation are provided in Supplementary Section~S2.2 and Table~S4.

Sparse supervision presents a second challenge: the model must simultaneously learn a generalizable solution operator and distinguish between different viscosity regimes from limited trajectories. To improve robustness, the encoder and decoder operate on spatially separated regions, where the encoder infers the latent representation from a conditioning window while the decoder predicts flow evolution on a distinct prediction domain (Fig.~\ref{fig:CFW_Result}b). We additionally introduce a supervised contrastive regularization term~\cite{khosla2020supervised} that encourages separation between latent representations associated with different system instances during early training. Together, these mechanisms stabilize latent-space organization under sparse supervision.

Despite the limited training data, the latent-conditioned model remains stable across both ID and OOD splits, achieving the lowest 50-step rollout error among the compared models (Table~\ref{tab:performance}); a comparison to the viscosity-supervised reference is reported in Supplementary Section~S2.4. The autoregressive baseline remains competitive at short horizons but accumulates rollout drift under distribution shift (Fig.~\ref{fig:CFW_Result}c). The no-conditioning baseline performs well at very short horizons, where distinguishing viscosity is less critical, but deteriorates at longer horizons (Table~\ref{tab:performance}), highlighting the importance of inferring system-specific variation.

Beyond rollout error metrics, we evaluate physical consistency through the drag coefficient at $\mathrm{Re}{=}300$ (Fig.~\ref{fig:CFW_Result}f). The latent-conditioned model recovers the reference drag coefficient $\bar{C}_D{=}1.36$~\cite{henderson1995details,wiliamson1996vortex,schafer1996benchmark} and Strouhal number $0.205$~\cite{henderson1995details} ($\mathrm{St}=0.2037$), whereas the autoregressive baseline develops frequency drift ($\mathrm{St}=0.2451$) and nonphysical drag behavior.

\subsection*{FitzHugh-Nagumo Reaction-Diffusion}

\begin{figure*}[htbp]
    \centering
    \includegraphics[width=\linewidth]{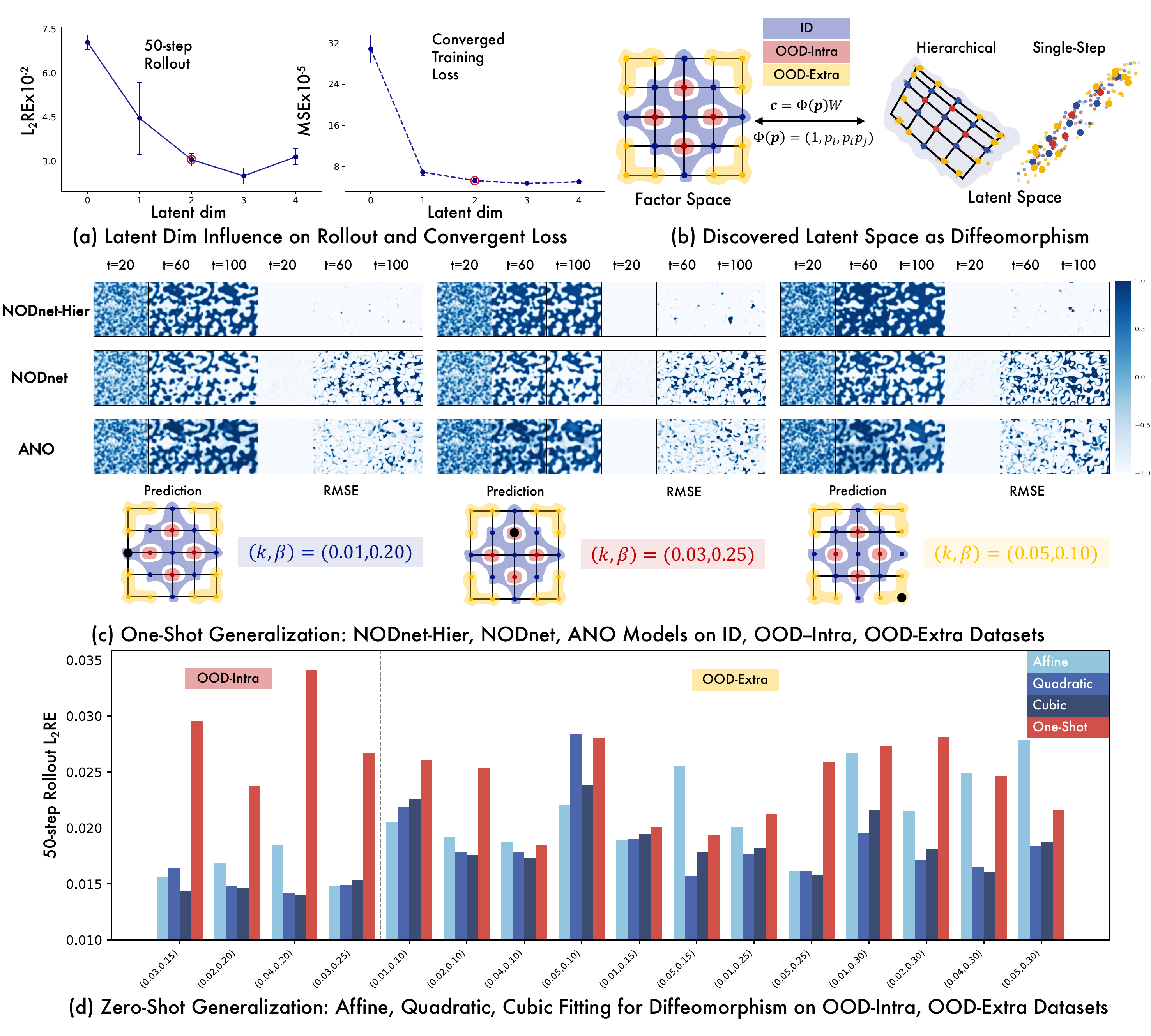}
    \caption{\textbf{FitzHugh--Nagumo reaction--diffusion results.} (a) Dimension selection across three random seeds identifies $d^*{=}2$, consistent with the two governing factors $(k,\beta)$ (Supplementary Section~S2.3).  (b) Learned latent-space organization for the 9 training system instances together with a quadratic mapping between latent and parameter spaces. Representations inferred from unseen system instances remain consistent with the learned structure. (c) Predictions and error maps for the hierarchical latent-conditioned model, the single-step latent-conditioned model, and the autoregressive baseline from the same initial condition. (d) Zero-shot prediction error as a function of the mapping complexity between latent and parameter spaces. Quadratic mappings improve prediction accuracy relative to affine mappings, while cubic mappings provide negligible additional benefit. One-shot inference is shown as a reference.}
    \label{fig:DiffusionReaction_Result}
\end{figure*}

The FitzHugh--Nagumo reaction--diffusion system~\cite{fitzhugh1961impulses,nagumo1962active} models nerve impulse propagation through excitable wave dynamics, where the qualitative regime of the system (quiescent, oscillatory, traveling pulses, or spiral patterns) is jointly determined by the stimulus parameter $k$ and the recovery rate $\beta$. The governing equations are
\begin{equation}
    \begin{split}
        \frac{\partial u}{\partial t} &= \gamma_u \nabla^2 u + u - u^3 + k - v\\
        \frac{\partial v}{\partial t} &= \gamma_v \nabla^2 v + \beta\left(u - v\right)
    \end{split}
    \label{diff-rxn}
\end{equation}
where $u$ is the membrane potential, $v$ is the recovery variable, and periodic boundary conditions are imposed.

This system provides a setting for evaluating whether the learned latent space captures interpretable system-level variation and supports generalization beyond the training regime. Training uses 9 $(k,\beta)$ pairs. The ID split evaluates unseen initial conditions at these parameter pairs, the OOD-Intra split evaluates 4 unseen parameter pairs inside the training hull, and the OOD-Extra split evaluates 12 pairs outside the training range (Fig.~\ref{fig:Methods}III(c)).

\begin{table}[!htbp]
    \centering
    \caption{\textbf{Generalization performance across the three dynamical systems.} Errors are reported as $L_2$ relative error ($\times 10^{-2}$, mean$\pm$std) at 1-, 5-, and 50-step rollout horizons. Latent-conditioned models (NODnet), no-conditioning baselines (SNO), and autoregressive baselines (ANO) are compared across in-distribution and out-of-distribution settings. {The parameter-supervised reference (PNO), provided for context, is reported in Supplementary Section~S2.4.} Reported statistics for latent-conditioned and no-conditioning models are averaged over three random seeds, with variability computed across both seeds and evaluation trajectories (Supplementary Sections~S2.1--S2.3). Model definitions and evaluation splits are provided in Methods.}
    \begin{tabular}{clccc}
        \toprule
        \multirow{2}{*}{Model} & \multirow{2}{*}{Subset} & \multicolumn{3}{c}{$L_2$ error ($\times 10^{-2}$)}\\
        \cmidrule(lr){3-5}
         & & 1-step & 5-step & 50-step \\
        \midrule
        \multicolumn{5}{l}{\textit{Burgers' Equation}} \\
        \midrule
        SNO-DeepONet (5.0M)               & ID            & $5.14{\pm}2.50$ & $12.78{\pm}9.15$ & $27.62{\pm}25.80$ \\
        & OOD           & $5.38{\pm}2.47$ & $14.05{\pm}9.05$ & $30.35{\pm}25.84$ \\
        & OOD-inviscid  & $5.29{\pm}1.49$ & $12.90{\pm}4.85$ & $24.98{\pm}9.38$ \\
        \textbf{NODnet-DeepONet} (10M)       & ID            & $4.46{\pm}1.87$ & $7.56{\pm}3.60$ & $\mathbf{17.62{\pm}13.07}$ \\
        & OOD           & $4.59{\pm}1.85$ & $7.52{\pm}3.65$ & $\mathbf{17.71{\pm}14.09}$ \\
        & OOD-inviscid  & $4.74{\pm}1.42$ & $9.17{\pm}4.09$ & $\mathbf{20.34{\pm}9.05}$ \\
        NODnet-AR (10M)                      & ID            & $3.47{\pm}3.68$ & $12.02{\pm}10.92$ & $61.48{\pm}36.14$ \\
        & OOD           & $4.09{\pm}4.02$ & $13.76{\pm}11.56$ & $64.42{\pm}36.38$ \\
        & OOD-inviscid  & $2.15{\pm}0.98$ & $8.95{\pm}4.71$   & $55.18{\pm}20.97$ \\
        ANO$^{\downarrow}$ (9.8M)         & ID            & $\mathbf{0.92{\pm}0.67}$ & $\mathbf{6.19{\pm}3.41}$ & $62.88{\pm}34.34$ \\
        & OOD           & $\mathbf{0.99{\pm}0.73}$ & $\mathbf{6.49{\pm}3.53}$ & $64.73{\pm}34.31$ \\
        & OOD-inviscid  & $\mathbf{1.31{\pm}0.74}$ & $\mathbf{6.72{\pm}2.89}$ & $64.86{\pm}29.31$ \\
        \midrule
        \multicolumn{5}{l}{\textit{Cylinder Flow Wake}} \\
        \midrule
        SNO (3.5M)               & ID  & $\mathbf{1.47{\pm}0.65}$ & $\mathbf{3.46{\pm}1.93}$ & $36.09{\pm}15.17$ \\
        & OOD & $\mathbf{1.63{\pm}0.84}$ & $6.11{\pm}3.76$ & $34.53{\pm}13.76$ \\
        \textbf{NODnet} (15M)       & ID  & $1.69{\pm}0.79$ & $4.00{\pm}2.60$ & $\mathbf{29.89{\pm}16.61}$ \\
        & OOD & $1.72{\pm}0.86$ & $\mathbf{4.38{\pm}2.34}$ & $\mathbf{31.04{\pm}15.11}$ \\
        ANO$^{\downarrow}$ (14M) & ID  & $2.21{\pm}0.82$ & $5.33{\pm}2.90$ & $31.63{\pm}12.26$ \\
        & OOD & $1.99{\pm}0.95$ & $4.52{\pm}2.32$ & $36.31{\pm}9.43$ \\
        \midrule
        \multicolumn{5}{l}{\textit{FitzHugh-Nagumo Reaction-Diffusion}} \\
        \midrule
        SNO-Hier (2.5M)               & ID        & $1.31{\pm}0.09$ & $1.78{\pm}0.10$ & $7.33{\pm}1.63$ \\
        & OOD-Intra & $1.28{\pm}0.06$ & $1.75{\pm}0.07$ & $7.05{\pm}1.24$ \\
        & OOD-Extra & $1.38{\pm}0.11$ & $1.96{\pm}0.16$ & $9.31{\pm}2.28$ \\
        \textbf{NODnet-Hier} (3.1M)      & ID        & $0.91{\pm}0.06$ & $\mathbf{1.20{\pm}0.10}$ & $\mathbf{2.52{\pm}0.67}$ \\
        & OOD-Intra & $0.90{\pm}0.04$ & $\mathbf{1.19{\pm}0.09}$ & $\mathbf{3.05{\pm}0.93}$ \\
        & OOD-Extra & $1.03{\pm}0.19$ & $\mathbf{1.61{\pm}0.66}$ & $\mathbf{5.49{\pm}4.38}$ \\
        NODnet (3.1M)                    & ID        & $\mathbf{0.79{\pm}0.03}$ & $2.21{\pm}0.15$ & $22.56{\pm}2.81$ \\
        & OOD-Intra & $\mathbf{0.79{\pm}0.02}$ & $2.18{\pm}0.11$ & $21.83{\pm}2.41$ \\
        & OOD-Extra & $\mathbf{0.79{\pm}0.04}$ & $2.27{\pm}0.16$ & $23.70{\pm}3.41$ \\
        ANO$^{\downarrow}$ (2.8M)     & ID        & $1.06{\pm}0.20$ & $2.07{\pm}0.55$ & $46.70{\pm}5.73$ \\
        & OOD-Intra & $0.97{\pm}0.13$ & $1.85{\pm}0.36$ & $44.68{\pm}2.80$ \\
        & OOD-Extra & $1.34{\pm}0.24$ & $2.69{\pm}0.58$ & $50.79{\pm}8.05$ \\
        \bottomrule
    \end{tabular}
    \label{tab:performance}
\end{table}

Sweeping the representation dimension from $d{=}0$ to $4$ reveals an optimum at $d^*{=}2$, consistent with the two independent governing factors $(k,\beta)$. Although prediction error continues to decrease slightly at $d{=}3$, principal component analysis of the learned representations shows that the third dimension remains effectively unused (Supplementary Section~S2.3). The resulting latent space organizes system instances according to the underlying parameter variation (Fig.~\ref{fig:DiffusionReaction_Result}b). A quadratic mapping between the latent space and the parameter space generalizes to held-out system instances with $R^2{=}0.9937$ and a Jacobian condition number of $13.6{\pm}7.0$, indicating that the learned correspondence remains smooth and stable beyond the training set (Eq.~S7).

We next examine how supervision structure influences factor identification. At matched model capacity, replacing the hierarchical decoder in NODnet-Hier with a single-step decoder in NODnet increases long-horizon rollout error by approximately an order of magnitude (Table~\ref{tab:performance}). The hierarchical decoder supervises prediction simultaneously at lead times $1$, $3$, and $9$, feeding longer-horizon dynamical information back into the encoder and stabilizing recovery of the latent organization (Fig.~\ref{fig:DiffusionReaction_Result}b, right). This parallels the Burgers' system, where the supervision geometry of the decoder strongly influences the ability to identify governing factors.

The learned latent space further enables zero-shot generalization to previously unseen system instances. To evaluate this behavior, we fit analytical mappings of increasing complexity between the latent space and the parameter space and generate predictions directly from sampled latent representations. The mappings include: affine (first-order), quadratic (second-order), and cubic (third-order), with details in Supplementary Information~S2.3. A quadratic mapping improves prediction accuracy relative to an affine fit, while higher-order polynomial fits provide negligible additional benefit (Fig.~\ref{fig:DiffusionReaction_Result}d). The comparable performance of one-shot and zero-shot inference (Fig.~\ref{fig:DiffusionReaction_Result}c) indicates that the latent space captures physically meaningful system variation that extrapolates beyond the training regime.

Baseline comparisons further support the role of latent conditioning in long-horizon prediction. Within the $(k,\beta)$ training hull, the latent-conditioned hierarchical model achieves nearly uniform performance across the ID and OOD-Intra splits, with the lowest 5- and 50-step rollout error among the compared models (Table~\ref{tab:performance}). The no-conditioning baseline fails to distinguish between parameter regimes and accumulates substantially larger long-horizon error. The autoregressive baseline performs worse still, as it conditions prediction only on past-state histories and accumulates rollout drift over time (Fig.~\ref{fig:DiffusionReaction_Result}c). Notably, even without hierarchical prediction, the latent-conditioned model achieves roughly half the 50-step prediction error of the autoregressive baseline, indicating that latent conditioning together with trajectory-decoupled sampling accounts for much of the performance gain.

\section*{Discussion}

We introduced Neural Operator Discovery as the problem of learning both a shared solution operator and a representation of system-specific variation directly from heterogeneous trajectories without parameter supervision. Across the systems studied here, the learned latent space aligns consistently with the underlying governing factors, while long-horizon prediction outperforms autoregressive and no-conditioning baselines (Table~\ref{tab:performance}) and remains within a bounded gap of the parameter-supervised reference reported in Supplementary Section~S2.4. These results suggest that latent conditioning can provide an interpretable and scalable alternative to explicit parameter supervision in operator learning.

A central observation of this work is that the learned latent space exhibits smooth structure across system instances and supports both one-shot and zero-shot generalization. In particular, the FitzHugh--Nagumo system demonstrates extrapolation beyond the convex hull of the training regime, despite the model never observing those parameter combinations during training. Although the method is trained only to minimize prediction error, two aspects of the formulation appear important for this behavior. First, trajectory-decoupled sampling encourages the encoder to retain only information shared across trajectories from the same system instance, suppressing dependence on transient initial-condition features. Second, restricting the latent representation to the minimal dimension sufficient to distinguish system instances limits the capacity available for encoding trajectory-specific variation. Together, these constraints encourage a compact latent organization that reflects system-level structure.

The results also suggest that factor identification depends strongly on the supervision geometry of the decoder. In both the Burgers' and FitzHugh--Nagumo systems, decoders receiving longer-horizon or full-field supervision recover more structured latent representations and achieve more stable long-term rollouts than autoregressive predictors trained only on local transitions. This indicates that accurate system identification may require supervision signals that remain sensitive to governing-factor variation over extended temporal horizons.

The Neural Operator Discovery setting also connects naturally to broader questions of identifiability in latent-variable models. Khemakhem et al.~\cite{khemakhem2020variational} establish identifiability in nonlinear ICA through auxiliary variables, while Locatello et al.~\cite{locatello2020weakly} analyze weakly supervised disentanglement under shared latent factors. Hyv\"arinen et al.~\cite{hyvarinen2019nonlinear} identify latent structure through temporal contrastive learning. In our setting, trajectories grouped by system instance provide an implicit supervisory signal, while latent dimension selection introduces an additional model-selection component absent from most existing approaches. Supplementary Sections~S3.2--S3.6 further discuss connections to latent-variable models for scientific data~\cite{chen2022automated,chen2024constructing}, gradient-based meta-learning~\cite{li2023metalearning}, symbolic regression, and generative modeling~\cite{goodfellow2014generativeadversarialnetworks}.

Several limitations remain. The present study considers continuously varying factors in low-dimensional Euclidean spaces; discrete system classes, bifurcating dynamics, and topologically nontrivial factor manifolds remain unexplored. All experiments are conducted on 1D and 2D systems, and extension to large-scale 3D dynamics remains an important direction for future work. The datasets considered here are simulation-generated; robustness under experimental noise, sparse sensing, and missing observations requires further investigation. In addition, the minimum number of system instances required for reliable latent-space formation remains unclear, particularly in sparse-data regimes. Finally, while the learned latent spaces exhibit smooth and approximately invertible correspondence with governing factors across the systems studied here, a formal theoretical characterization of these properties remains open.

\section*{Methods}

This section summarizes the implementations used across the three primary case studies, including dataset generation and encoder and decoder architectures. The Burgers' system uses a CNN encoder paired with a DeepONet-style decoder; the cylinder flow wake uses Point Transformer backbones for both encoder and decoder; and the FitzHugh--Nagumo system uses a recurrent encoder with a hierarchical U-CNN decoder. Full architectural details, hyperparameters, and dimension-selection procedures are provided in Supplementary Sections~S2.1--S2.3.

\subsection*{Burgers' Equation}

\begin{figure*}[htbp]
    \centering
    \includegraphics[width=\linewidth]{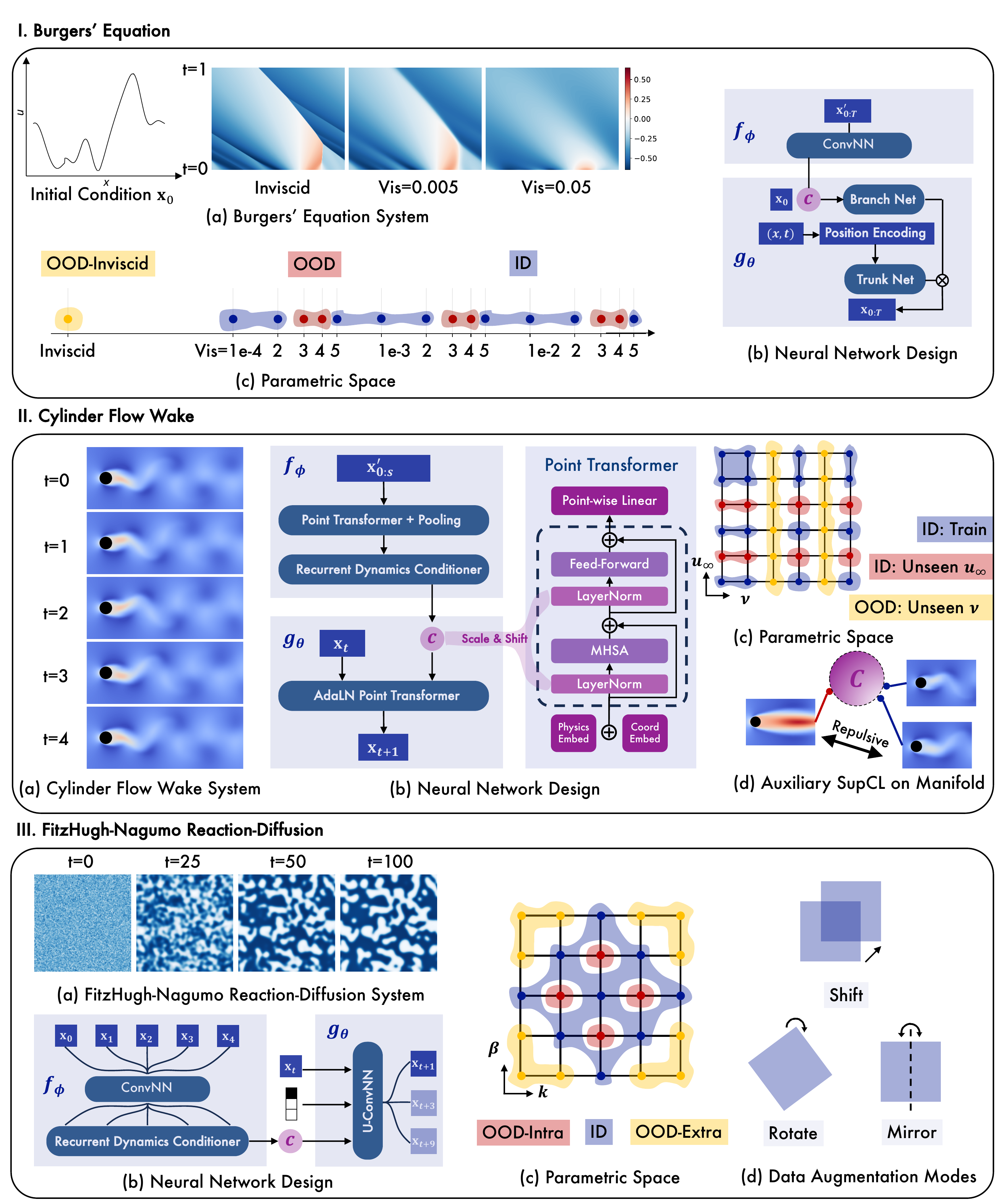}
    \caption{\textbf{System descriptions and learning configurations.} \textbf{I. Burgers' equation:} (a) sampled spatiotemporal fields across viscosity regimes,  (b) CNN encoder and DeepONet-style decoder with positional encoding, (c) training and evaluation splits in viscosity space. 
    \textbf{II. Cylinder flow wake:} (a) sampled trajectory on the irregular mesh, (b) Point Transformer encoder and AdaLN-conditioned Transformer decoder, (c) training and evaluation splits across viscosity and inlet velocity, (d) supervised contrastive regularization of the latent space under sparse supervision.
    \textbf{III. FitzHugh--Nagumo reaction--diffusion:} (a) sampled trajectory, (b) recurrent encoder and hierarchical U-CNN decoder, (c) parameter-space splits across $(k,\beta)$ pairs, (d) symmetry-based data augmentation under O(2) transformations. }
    \label{fig:Methods}
\end{figure*}

Burgers' equation is solved using spatial and temporal discretizations satisfying the CFL stability condition ($\Delta x{=}5\times10^{-4}$, $\Delta t{=}5\times10^{-3}$). Simulations are generated on a $4001\times1001$ grid and downsampled to $401\times101$ for learning. The training set spans nine viscosity values in the viscous regime, while evaluation includes both interpolation within the training range and extrapolation to the inviscid limit $\nu{=}0$ (Fig.~\ref{fig:Methods}I(c)). For each system instance, 50 trajectories are generated from different initial conditions.

Following the factorized formulation in Eq.~\ref{eq:decompose}, the encoder infers a latent representation from a conditioning trajectory, while the decoder predicts the full spatiotemporal field from an initial condition and the inferred representation.

\paragraph*{Encoder}

The encoder is implemented as a convolutional neural network that processes the conditioning trajectory as a $(T,X)$ image and outputs a scalar latent representation through global pooling and a small multilayer perceptron (MLP).

\paragraph*{Decoder}

The decoder is implemented using a DeepONet-style architecture that predicts the solution field at arbitrary spatiotemporal coordinates in a single forward pass:
\begin{equation}
    \hat{u}(t, x) = u_0(x) + \alpha_{\text{res}}\,\bigl\langle\,b(\mathbf{u}_0, c),\,\tau\bigl(\gamma(x, t)\bigr)\,\bigr\rangle,
\end{equation}
Branch and trunk networks are implemented as MLPs, and the coordinate inputs are sinusoidally positional encoded to improve representation of sharp shock fronts (Supplementary Section~S2.1). Training minimizes mean-squared prediction error over randomly sampled spatiotemporal query points. The NODnet-AR architecture can also be found in Supplementary Section~S2.1.

% \paragraph*{Baselines}

% To evaluate the role of latent conditioning, we compare against three baselines. The autoregressive latent-conditioned variant replaces the full-field decoder with a U-Net autoregressive stepper conditioned on the inferred latent representation. The parameter-supervised baseline replaces the inferred latent representation with the true viscosity while keeping the decoder unchanged. The autoregressive baseline conditions prediction directly on past-state windows without latent conditioning, and its total parameter count matches the full latent-conditioned model. 

\subsection*{Cylinder Flow Wake}

The dataset is generated from high-fidelity simulations of incompressible flow past a cylinder on the 2D domain $\Omega=[-9,21]\times[-6,6]$. Inlet velocity takes values $u_\infty \in {15,16,17,18,19,20}$ and viscosity $\nu \in {0.005,0.008,0.011,0.014,0.017,0.020}$. With cylinder diameter fixed at $0.1$, the resulting system family spans Reynolds numbers from $75$ to $400$, and the simulations are validated against reference statistics~\cite{henderson1995details}. To evaluate learning under sparse supervision, training uses only 16 trajectories sampled across $(\nu,u_\infty)$ combinations (Fig.~\ref{fig:Methods}II(c)). Generalization is evaluated on two settings: unseen inlet velocities at trained viscosities, and unseen viscosities outside the training range.

The decoder predicts the next-step velocity field autoregressively from the current state and a latent representation inferred by the encoder from a separate conditioning trajectory generated under the same viscosity but different initial conditions. Conditioning trajectories are chosen adjacent to the prediction window so that the prediction setting remains comparable to history-conditioned autoregressive neural operators.

\paragraph*{Encoder and decoder}

Both encoder and decoder are implemented using Transformer architectures designed for irregular meshes~\cite{zhao2021point,alkin2024universal} (Fig.~\ref{fig:Methods}II(b)). The encoder processes a localized spatiotemporal conditioning window to infer the latent representation from wake dynamics, while the decoder predicts flow evolution over the target prediction region using AdaLN-Zero conditioning~\cite{peebles2023scalable}. Rotary Position Embedding (RoPE)~\cite{su2023roformerenhancedtransformerrotary} is used to preserve relative spatial information on the irregular mesh. Additional architectural details and ablation studies are provided in Supplementary Section~S2.2.

\paragraph*{Latent space regularization}

To stabilize latent space organization under sparse supervision, we introduce a supervised contrastive regularization term~\cite{khosla2020supervised} that encourages separation between representations associated with different system instances (Fig.~\ref{fig:Methods}II(d)). The contribution of this regularization is gradually annealed during training so that the prediction objective dominates optimization at later stages.

\subsection*{FitzHugh-Nagumo Reaction-Diffusion}
\label{sec:FNRD}

The FitzHugh--Nagumo reaction--diffusion dataset is generated on a $128^2$ grid across 25 distinct system instances: 9 for training, 4 for the OOD-Intra split (interpolated parameter pairs), and 12 for the OOD-Extra split (extrapolated parameter pairs) (Fig.~\ref{fig:Methods}III(c)). For each system instance, 50 trajectories are generated from Gaussian random-field initial conditions, with 40 used for training and 10 reserved for evaluation of initial-condition generalization.

The decoder predicts the next-step fields $(\mathbf{u},\mathbf{v})'_{t+1}$ autoregressively from the current state $(\mathbf{u},\mathbf{v})'_t$ and a latent representation inferred by the encoder from a separate conditioning trajectory generated under the same $(k,\beta)$ pair but different initial conditions.

\paragraph*{Encoder}

To process high-dimensional spatiotemporal trajectories, the encoder incorporates a recurrent inductive bias:
\begin{equation}
\begin{aligned}
\mathbf{h}_i &= \Psi(\mathbf{h}_{i-1}, \psi(\mathbf{x}_i)), \quad i = 0, \dots, s,\\
c &= o(\mathbf{h}_s),
\end{aligned}
\end{equation}
where $\Psi(\cdot,\cdot)$ denotes a recurrent unit (GRU~\cite{chung2014empirical} or LSTM~\cite{hochreiter1997long}) operating on learned per-frame representations $\psi(\mathbf{x}_i)$. The final hidden state is projected to the latent representation $c$ (Fig.~\ref{fig:Methods}III(b)).

\paragraph*{Decoder}

Because the dynamics are high-dimensional and evolve over long temporal horizons, the decoder predicts future states autoregressively using a hierarchical prediction strategy inspired by hierarchical temporal aggregation~\cite{pangu} and multi-output forecasting~\cite{green2024stratifyunifyingmultistepforecasting}. Three steppers with lead times of 1, 3, and 9 steps share a U-shaped CNN backbone (Fig.~\ref{fig:Methods}III(b)). Longer-horizon supervision stabilizes rollout prediction and improves recovery of system-level variation by feeding long-timescale dynamical information back into the latent-conditioning process.

Representing prediction horizons hierarchically also reduces the number of network function evaluations required for long rollouts from $\mathcal{O}(N)$ to $\mathcal{O}(\log_3 N)$ for the $N$-th prediction step. For example,
$$\hat{\mathbf{x}}_{25} = g_9^{(2)}\circ g_3^{(2)}\circ g_1^{(1)}(\mathbf{x}_0,c).$$

\paragraph*{Symmetry-aware data augmentation}

Under periodic boundary conditions, the FitzHugh--Nagumo system is equivariant to spatial translation, mirroring, and rotation (Fig.~\ref{fig:Methods}III(d)). Applying these transformations enlarges the effective dataset by approximately $10\times$ while preserving the underlying system instance. This augmentation encourages the encoder to identify governing factors independently of the spatial orientation or position of excitation patterns and improves robustness across initial conditions.

\bibliography{Refs}

\section*{Acknowledgments}
Z.C. and S.D. acknowledge financial support from MIT Sea Grant. Q.L. and J.H. are supported by the Major Research Plan of the National Natural Science Foundation of China (Grant No.~92470109) and the Zhejiang Provincial Team of Leading Talents in Innovation and Entrepreneurship (Grant No.~2024R01002). The authors acknowledge the MIT Office of Research Computing and Data (ORCD) for providing high-performance computing resources that have contributed to the research results reported within this paper. A portion of this research was performed using computational resources sponsored by the U.S. Department of Energy's Office of Critical Minerals and Energy Innovation and located at the National Laboratory of the Rockies. We thank Huaibo Chen, Benjamin Koenig, and Nicolas Tricard for advice on numerical methods for data generation in the early stage of this project.

\section*{Author contributions}
Z.C.: conceptualization, methodology, investigation, software, visualization, writing--original draft, writing--review and editing.
Q.L.: conceptualization, methodology, software (early prototyping), writing--review and editing.
J.H.: investigation, software, visualization.
S.D.: supervision, funding acquisition, writing--review and editing.

\section*{Competing interests}
The authors declare no competing interests.

\section*{Additional information}
\bmhead{Supplementary information} Supplementary Sections~S1--S3 provide additional benchmark cases (bistable oscillator, wall-bouncing system, mass-spring chain, KPP equation), extended simulation and architecture details for each case study, the drag-force computation procedure, efficiency comparison with iMODE, and alternative theoretical perspectives on the diffeomorphic latent space.

\bmhead{Data availability} The Burgers' dataset used in this work is available at https://doi.org/10.5281/zenodo.20372988. The FitzHugh-Nagumo reaction-diffusion dataset and cylinder flow wake dataset can be reproduced using MATLAB or OpenFOAM with the simulation parameters detailed in Supplementary Section~S3. 

\bmhead{Code availability} The source code for all seven case studies (three in the main text, four in Supplementary Information) is available at https://doi.org/10.5281/zenodo.20406331. 

\clearpage
\begingroup

\setcounter{section}{0}
\setcounter{subsection}{0}
\setcounter{subsubsection}{0}
\setcounter{figure}{0}
\setcounter{table}{0}
\setcounter{equation}{0}

\renewcommand{\thesection}{S\arabic{section}}
\renewcommand{\thefigure}{S\arabic{figure}}
\renewcommand{\thetable}{S\arabic{table}}
\renewcommand{\theequation}{S\arabic{equation}}

% Give the reset supplementary counters unique hyperlink destinations.
\providecommand{\theHsection}{}
\providecommand{\theHsubsection}{}
\providecommand{\theHfigure}{}
\providecommand{\theHtable}{}
\providecommand{\theHequation}{}
\renewcommand{\theHsection}{supp.\arabic{section}}
\renewcommand{\theHsubsection}{supp.\arabic{section}.\arabic{subsection}}
\renewcommand{\theHfigure}{supp.\arabic{figure}}
\renewcommand{\theHtable}{supp.\arabic{table}}
\renewcommand{\theHequation}{supp.\arabic{equation}}

\newcommand{\ngs}{NODnet}
\renewcommand{\cn}{encoder}
\renewcommand{\pn}{decoder}
\renewcommand{\tds}{trajectory-decoupled sampling}

\begin{center}
{\Large\bfseries Supplementary Information\par}
\vspace{0.75em}
{\Large Neural operator discovery from heterogeneous trajectories\par}
\vspace{1em}
Zituo Chen, Qiaofeng Li, Jiaxin Hu, and Sili Deng
\end{center}
\vspace{1.5em}

% The supplementary source remains independently compilable.  In this
% combined build, docmute ignores its preamble and document wrapper; its
% separate title and bibliography are suppressed in favor of the combined
% title above and the single reference list generated from both parts.
\let\maketitle\relax
\renewcommand{\bibliographystyle}[1]{}
\renewcommand{\bibliography}[1]{}

\maketitle

\section{Supplementary Results}\label{sec:supp_results}

Table~\ref{tab:supp_seven_cases} summarises the seven case studies reported across the main text and this Supplementary Information. Each case stresses a different aspect of the minimal conditioning principle and is implemented with a different backbone pair; the recovered $d^*$ matches the number of physically independent governing factors in every case.

\begin{table*}[htbp]
    \centering
    \caption{Seven NODnet case studies. Section, observation structure, governing factors, recovered latent dimension $d^*$, and the encoder/decoder backbones. Distinct backbones used across the seven cases: CNN, U-CNN, DenseNet, DeepONet, Point Transformer, recurrent (LSTM/GRU), Fourier Neural Operator, Neural ODE.}
    \scriptsize
    \setlength{\tabcolsep}{4pt}
    \resizebox{\textwidth}{!}{%
    \begin{tabular}{lllclll}
        \toprule
        Case & Section & Domain & Factors & $d^*$ & Encoder & Decoder \\
        \midrule
        Bistable oscillator    & S1.1     & 2-DoF ODE        & $(k_1, k_3)$    & 2 & DenseNet              & DenseNet \\
        Wall-bouncing ball     & S1.2     & 2D billiard      & $(W, H)$        & 2 & DenseNet              & DenseNet \\
        Mass-spring chain      & S1.3     & 10-DoF ODE       & $(c, k)$        & 2 & DenseNet              & Neural ODE \\
        KPP reaction-diffusion & S1.4     & 1D PDE           & $(r, D)$        & 2 & FNO + LSTM            & FNO \\
        Burgers' equation      & Main / S2.1 & 1D PDE        & $\nu$           & 1 & CNN                   & DeepONet (PE trunk) \\
        Cylinder-flow wake     & Main / S2.2 & 2D N--S, mesh & $\nu$           & 1 & Point Transformer     & Point Transformer (AdaLN) \\
        FitzHugh--Nagumo RD    & Main / S2.3 & 2D PDE        & $(k, \beta)$    & 2 & Recurrent (LSTM/GRU)  & Hierarchical U-CNN \\
        \bottomrule
    \end{tabular}}
    \label{tab:supp_seven_cases}
\end{table*}

The four subsections below complement the main text by stressing aspects of the design that the Burgers', FN-RD, and cylinder-wake cases do not: low-dimensional ODE families (bistable oscillator), geometric variation (wall-bouncing system), boundary-condition generalization (mass-spring chain), and resolution-agnostic discovery (KPP). In every case, the same design, namely a compact encoder producing a latent at the gauged minimum dimension and a decoder conditioned on it, is used; only the backbone architectures change to match the observation structure.

\subsection{Bistable system: dimension gauging on a low-dimensional ODE family}

This case establishes the elbow signal and the emergent diffeomorphism on the simplest setting: a two-parameter family with no spatial structure. We consider a bistable oscillator with two stable equilibrium states (Fig.~\ref{fig:Bistable}(a)), whose simplest form is characterized by a cubic polynomial stiffness featuring a negative linear coefficient and a positive cubic coefficient. 
\begin{equation}
\begin{bmatrix}
\dot{x} \\
\ddot{x}
\end{bmatrix}
=
\underbrace{
\begin{bmatrix}
0 & 1 \\
k_1 &-c
\end{bmatrix}
}_{\text{linear part}}
\begin{bmatrix}
x \\
\dot{x}
\end{bmatrix}
+
\underbrace{
\begin{bmatrix}
0 \\
-k_3 x^3
\end{bmatrix}
}_{\text{nonlinear part}}
\end{equation}

Bistable systems exhibit excitation-dependent nonlinear behaviors: small excitations result in intra-well oscillations, where the system oscillates around one equilibrium state, while large excitations trigger inter-well oscillations characterized by transitions between the two stable states. In this example, we demonstrate that \ngs{} is able to learn the shared dynamics of the system family without knowing the actual factors.

\begin{figure}[htbp]
    \centering
    \includegraphics[width=\linewidth]{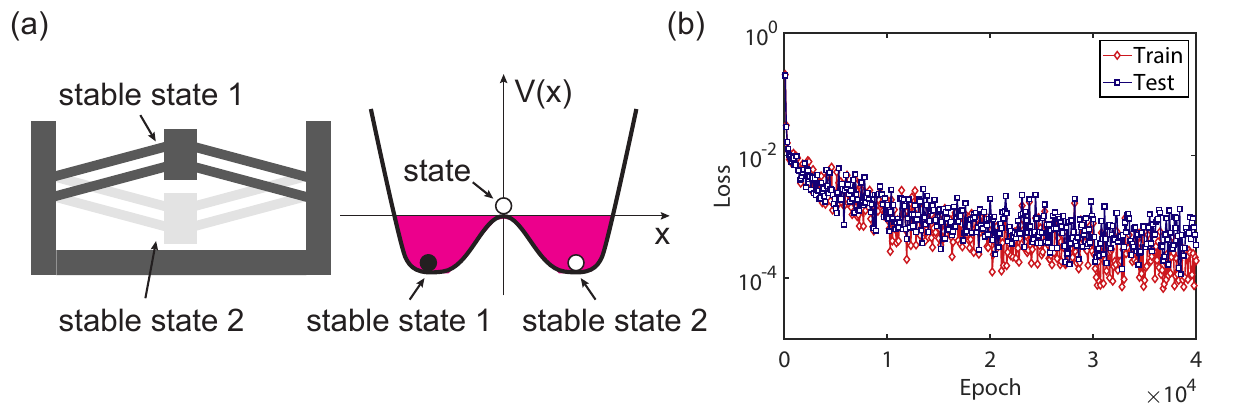}
    \caption{(a) An illustrative bistable system. Two potential wells typically exist in the potential function of a bistable system, marking two stable states. The system thus exhibits local intra-well and global inter-well motion patterns; (b) Sample training and test loss histories for the bistable system.}
    \label{fig:Bistable}
\end{figure}

The training dataset contains 112 system instances, each taking a factor combination between $k_1=[-1.0:-0.1:-0.4] \ \si{N/m }$ and $k_3=[2.0:0.2:5.0] \ \si{N / m^3}$. For each system instance 10 trajectories, with initial positions $x_0=[0.1:0.1:1.0] \ \si{m}$, initial velocities $\dot{x}_0 = 0 \ \si{m/s}$, and time marching stepsize $\Delta t=0.1 \ s$ are generated. We randomly sample trajectories of length 500 within the long 1000-step trajectories to mimic random initial conditions for conditioning sequences. Sample train and test results are shown in Fig.~\ref{fig:Bistable}(b). The test trajectories are generated in the same way as the training sequences but of a different set of factors. After training, the physically meaningful latent space is formed.

We sweep the dimension of the latent $\mathbf{c}$ and track the converged training loss. The loss curve forms an ``elbow'' shape as the latent dimension increases; the ``elbow'' point marks the optimal dimension for $\mathbf{c}$, equal to the true number of independent factors of the system.

The $\mathbf{c}$-space is empirically diffeomorphic to the true factor space. We validate this by training a neural ODE~\cite{chen2018neural} that transports the learnt $\mathbf{c}$ to the true factor $\boldsymbol{\phi}$~\cite{lee2003smooth}, i.e., finding weights $\boldsymbol{\Theta}$ of a neural network $\mathbf{v}$ such that
\begin{align}
    &\frac{\mathrm{d}\mathbf{z}(t)}{\mathrm{d} t} = \mathbf{v}_{\boldsymbol{\Theta}}(\mathbf{z}) \nonumber \\
    &\mathbf{z}(0) = \mathbf{c}_i,~\mathbf{z}(1) = \boldsymbol{\phi}_i,~i=1,\ldots,N_s
\end{align}

The learnt NODE realises a continuous and differentiable transformation from the $\mathbf{c}$ space into the true factor space, instantiating the diffeomorphism. The latent space is physically meaningful and fully interpretable, with each point indexing a system instance and enabling one-shot generalization to unseen system instances for response prediction.

\subsection{Wall-bouncing system: geometry as the governing factor}

This case tests whether the minimal conditioning principle extends to a governing factor that is \emph{geometric} rather than a coefficient in an equation. A ball bounces elastically inside a rectangular region whose width and height vary across system instances (Fig.~\ref{fig:Wall}(a)); the two factors control the domain boundary rather than a differential term, and the encoder sees only trajectories, not the domain dimensions. The \ngs{} architecture is shown in Fig.~\ref{fig:Wall}(b): the \cn{} receives trajectories, and the \pn{} receives the concatenation of an initial condition and the latent produced by the \cn{}, trained against trajectory targets. 

We utilize a dataset comprising 25 system instances. Each instance corresponds to a specific combination of factors $\textrm{width}=[2.1,2.3,2.5,2.7,2.9] \ \si{m}$ and $\textrm{height}=[2.1,2.3,2.5,2.7,2.9]  \ \si{m}$. For each system instance, 25 trajectories are generated with different initial velocities. Trajectory time sampling stepsize $\Delta t=0.1\ \si{s}$. We divide the dataset into a training set ($13$ system instances) and a test set ($12$ system instances), separated by odd/even numbering (Fig.~\ref{fig:Wall}(c)).

We employ DenseNet\cite{huang2017densely} architecture for both \cn{} and \pn{} (see Fig.~\ref{fig:Wall}(b)). The learnt latent space for training and testing are shown in Figs.~\ref{fig:Wall}(d) and (e). After constructing a diffeomorphism between the latent space and true geometric factor space, the transported latents are shown in Figs.~\ref{fig:Wall}(f) and (g), with true (width, height) pairs indicated by black dots. Figure~\ref{fig:Wall}(h) illustrates the learned diffeomorphism mapping from the latents to the true geometric factors. 

% corresponding test results are presented in Fig.~\ref{fig:Wall}(d). Training and testing results for the diffeomorphism are displayed in Fig.~\ref{fig:Wall}(e) and~(f),  Figure~\ref{fig:Wall}(g) illustrates the learned diffeomorphism mapping the latent variables to the true geometric factors. 

\begin{figure}[htbp]
    \centering
    \includegraphics[width=\linewidth]{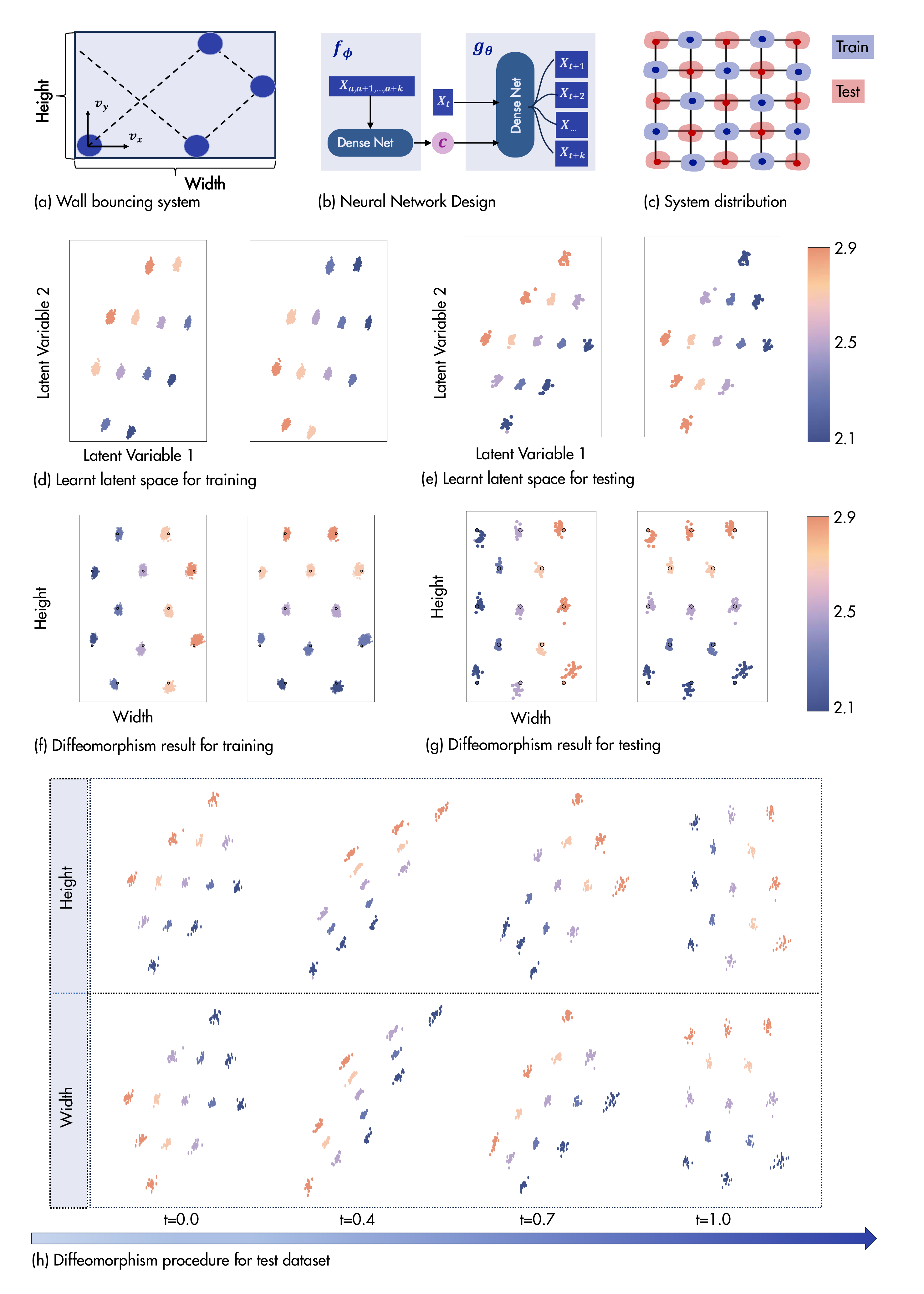}
    \caption{(a) An illustrative wall bouncing system; (b) The adopted \ngs{} architecture in the wall bouncing system case; (c) The distribution of system instances with respect to geometric factors; (d) The learnt latent space for training; (e) The learnt latent space for testing; (f) The diffeomorphism result for training; (g) The diffeomorphism result for testing; (h) Illustration of the diffeomorphism procedure from latent space to true geometric factor space for test dataset.}
    \label{fig:Wall}
\end{figure}

\subsection{Mass-spring chain: boundary-condition generalization via a differential decoder}

This case tests whether the minimal conditioning principle, together with a decoder built on a differential operator, supports generalization across \emph{unseen boundary conditions} at test time. The system is a 10-DoF mass-spring chain (Fig.~\ref{fig:Massspring}(a)). The \pn{} is a neural ordinary differential equation (NODE, Fig.~\ref{fig:Massspring}(b)), which exposes the boundary through its state rather than through its parameters. Because the latent produced by the \cn{} carries only the governing factors (damping and stiffness), and not the boundary specification, changing the boundary at test time leaves the learned operator family intact.

The training dataset contains 16 system instances, each taking a factor combination between $c=[0.1,0.3,0.5,0.7] \ \si{N \cdot s/m}$ and $k=[2.0,3.0,4.0,5.0] \ \si{N / m}$. For each system instance 11 trajectories with random initializations and time marching stepsize $\Delta t=0.1$ s are generated.

First we train and test on the same boundary conditions. We train with trajectories of 20 time steps; after training, we generate trajectories of 50 time steps. The results are shown in Fig.~\ref{fig:Massspring}(d). \ngs{} performs accurate extrapolation outside of the training time span. This proves that \ngs{} with NODE as \pn{} is able to perform accurate long-term rollouts.

The learnt latent space (training denoted as circles, testing as asterisks) is shown in Fig.~\ref{fig:Massspring}(c). Two principal directions exist in the latent space, respectively corresponding to damping and stiffness increases of the system family.

Then we train with the right boundary \textit{fixed} and test with the right boundary \textit{free}. Test is performed on system instances excluded in the training dataset. When trajectories with the \textit{fixed} boundary condition are available for the test system instances, we simply pass these trajectories through \cn{} to generate the latents. These variables together with the NODE are directly applied to the \textit{free} boundary condition for test-time prediction. The results are shown in Fig.~\ref{fig:Massspring_test}(a).

When only trajectories with the \textit{free} boundary condition are available for the test system instances, the \pn{} can be used as a \cn{}, similar to the approach in \cite{li2023metalearning}.
\begin{equation}
    \mathbf{c} = \mathop{\arg\min}_{\mathbf{c}} \Vert \mathbf{X} -\textrm{NODE}(\mathbf{x}_0,\mathbf{c}) \Vert_2
    \label{eq:msc_op}
\end{equation}
where $\mathbf{X}$ is a trajectory of the test system instance with free boundary condition. After solving Eq.~\eqref{eq:msc_op}, the NODE parameterized by its solution $\mathbf{c}$ is capable of trajectory prediction given new initial conditions. The results are shown in Fig.~\ref{fig:Massspring_test}(b).

\begin{figure}[htbp]
    \centering
    \includegraphics[width=\linewidth]{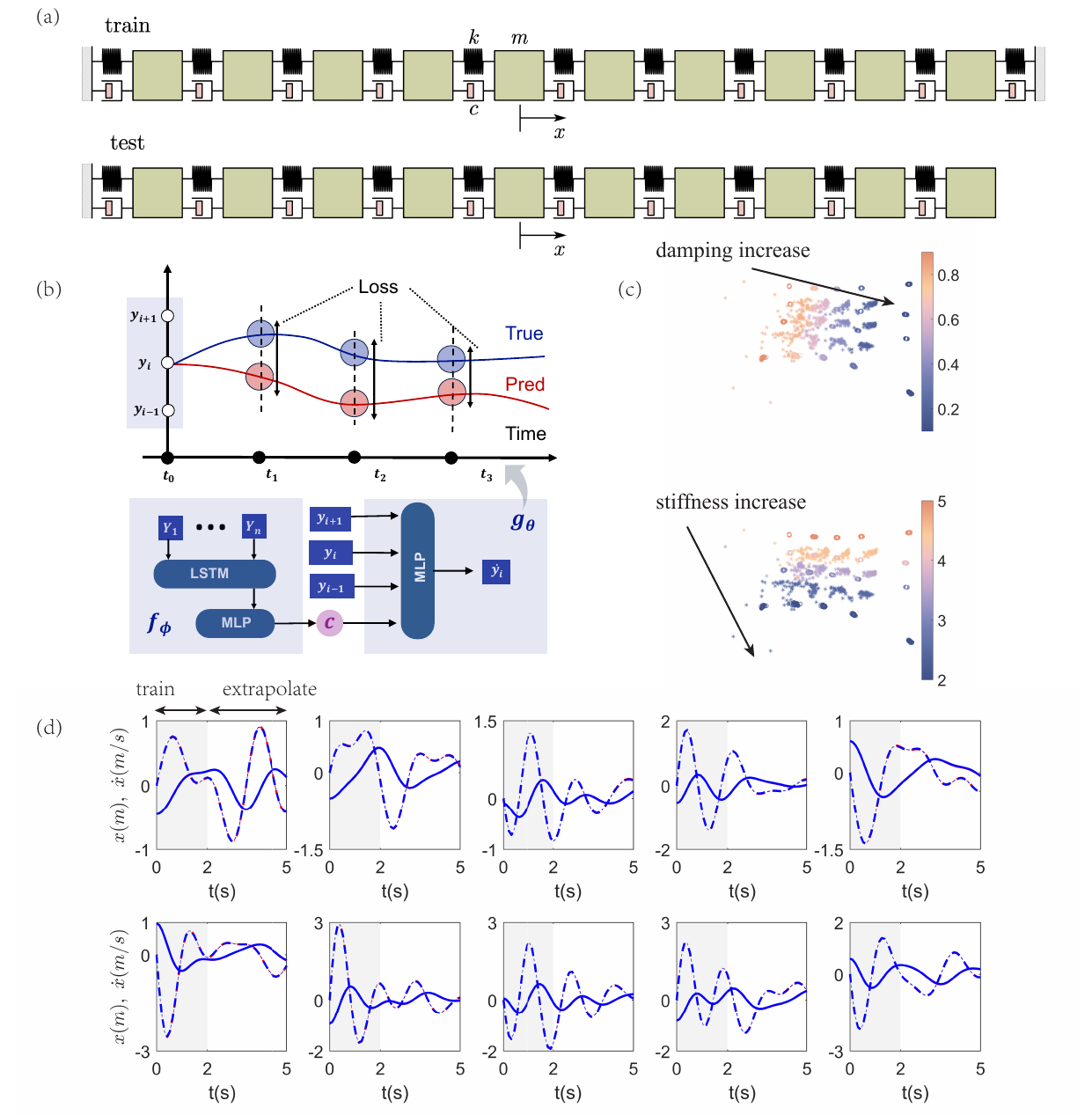}
    \caption{(a) An illustrative 10-DoF mass-spring chain system. We train with fixed boundary condition, test with free boundary condition on the right; (b) The \ngs{} architecture for the mass-spring-chain case. NODE is adopted as the \pn{}; (c) The learnt latent space. Clearly 2 principal directions exist corresponding to damping and stiffness increases of the physical system family; (d) Train and extrapolation results. With NODE as \pn{}, \ngs{} is able to perform accurate long-time rollouts outside of the training time span.}
    \label{fig:Massspring}
\end{figure}

\begin{figure}[htbp]
    \centering
    \includegraphics[width=\linewidth]{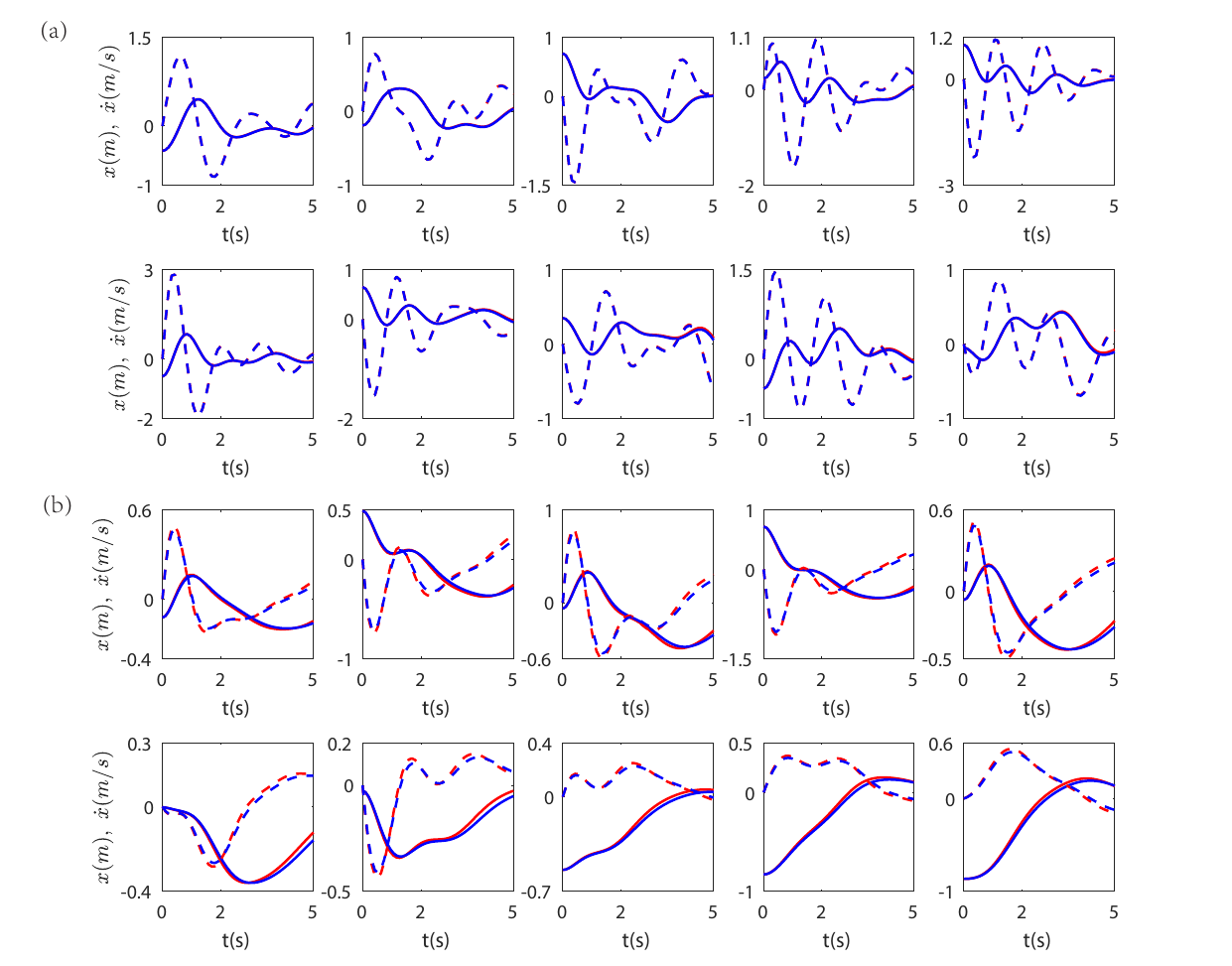}
    \caption{(a) Boundary condition generalization when conditioning data is of the same boundary condition as the training data. In this scenario, the \cn{} can be used to generate the latent $\mathbf{c}$; (b) Boundary condition generalization when conditioning data is of a different boundary condition from the training data. In this scenario, the \cn{} cannot be used. $\mathbf{c}$ has to be generated via solving an optimization problem (Eq.~\eqref{eq:msc_op}).}
    \label{fig:Massspring_test}
\end{figure}

\subsection{Kolmogorov-Petrovsky-Piskunov (KPP) equation: resolution-agnostic discovery with FNO backbones}

This case tests whether the minimal conditioning principle holds when the encoder and decoder are both Fourier Neural Operators (FNOs), and whether the resulting operator family transfers across grid resolutions without retraining. Two questions follow. First, does the encoder, built on FNO layers, still discover the correct latent dimension on a resolution-agnostic representation? Second, does a latent inferred on a low-resolution trajectory parametrize the decoder correctly when the decoder is evaluated on a high-resolution grid? A positive answer to both would mean the principle is compatible with resolution generalization that label-conditioned methods cannot deliver without re-training.

The system governing equation is 
\begin{equation}
    \frac{\partial u}{\partial t}=D \frac{\partial^{2} u}{\partial x^{2}}+r u(1-u)
\end{equation}
In this case, the training dataset contains 20 system instances, each taking a factor combination between $r=[0.01,0.02,0.03,0.04,0.05]$ and $D = [0.005, 0.010, 0.015, 0.020, 0.025]$. %The results are shown in Fig.~\ref{fig:KPP}.

The \ngs{} architecture and overall pipeline are shown in Figs.~\ref{fig:KPP}(a) and (b). A time sequence of the solution field $\{\mathbf{u}_j^c\in$$\mathcal{R}^{x_N}$$\} \ (j=1,2,...,N)$ is passed through a modified FNO paralleled by the spatial coordinates, where we keep the Fast Fourier Transform but discard the Inverse Fast Fourier Transform. The resulting vectors are passed through an LSTM to process the temporal evolution, outputting the latent vector $\mathbf{c} \in \mathcal{R}^{2}$, which acts as the system indicator. The \pn{} is also an FNO. Its input is a matrix formed by four vectors in $\mathcal{R}^{x_N}$: the spatial coordinate vector $\mathbf{x}$, the solution field vector $\mathbf{u}_t$ at time $t$, and two copies of the latent vector $\mathbf{c}$ broadcast across the spatial grid. The output of \pn{} is the solution field at the next time step.

As shown in Fig.~\ref{fig:KPP}, we train \ngs{} on low-resolution grid and directly test on high-resolution grid. The 0-shot high-resolution prediction results are illustrated in Fig.~\ref{fig:KPP}(c), with small errors existing near the boundaries. The learnt latent space is shown in Fig.~\ref{fig:KPP}(d). Two principal directions exist corresponding to reaction and diffusion coefficients increases.

\begin{figure}[htbp]
    \centering
    \includegraphics[width=\linewidth]{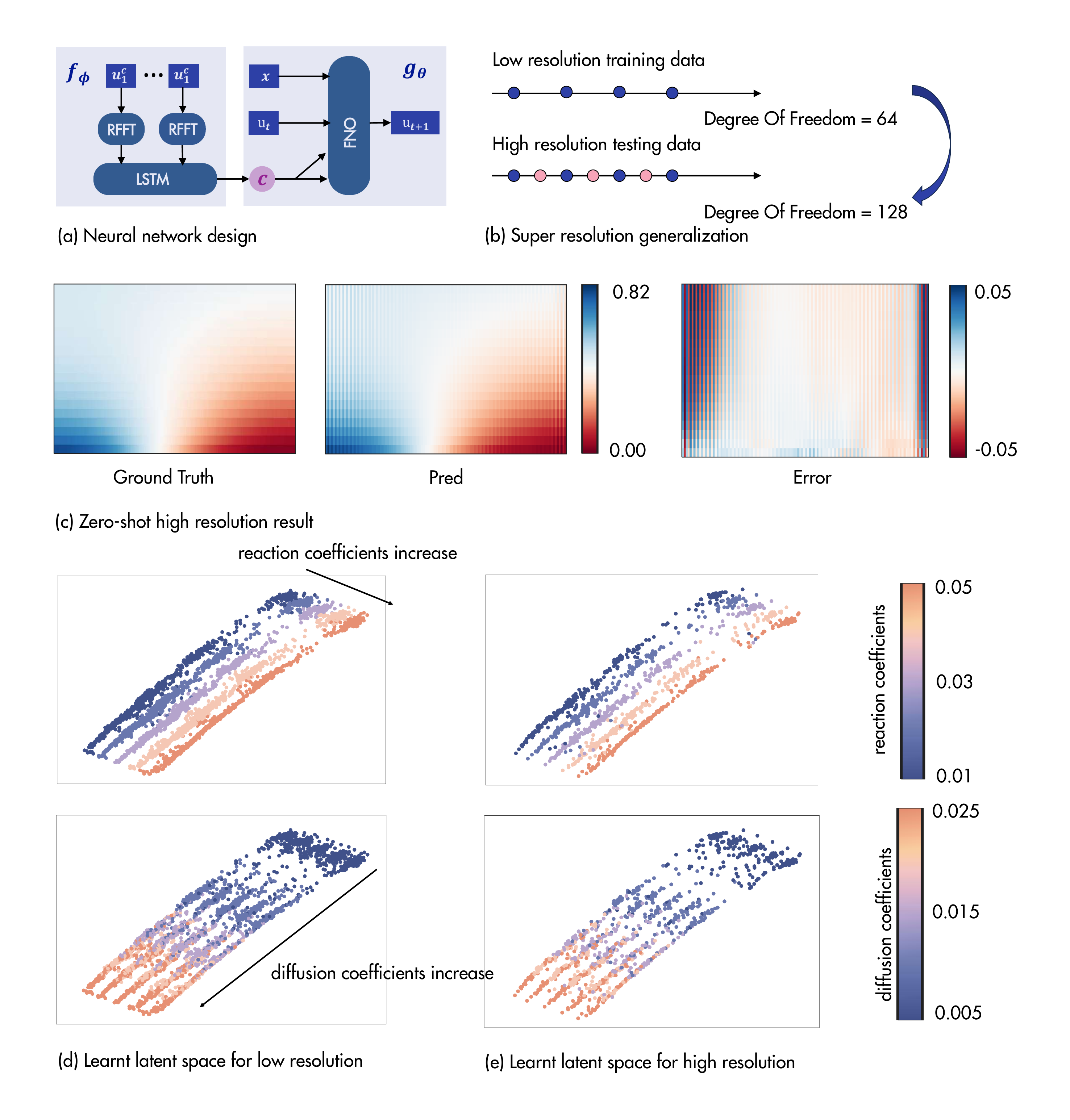}
    \caption{(a) The \ngs{} architecture for the KPP system case; (b) {\ngs{} is trained on low-resolution grid and directly tested on high-resolution grid}; (c) Zero-shot high-resolution test result; (d) The learnt latent space. Two principal directions exist corresponding to reaction and diffusion coefficients increases.}
    \label{fig:KPP}
\end{figure}

%%%%%%%%%%%%%%%%%%
\section{Extended Results}\label{sec:extended_results}

This section reports detailed numerical results, ablations, and case-specific design rationale for the three main-text cases. Each subsection collects the simulation setup, decoder choice, latent-dimension sweep, and split-by-split tables that the main text references but does not display in full.

\subsection{Details of Burgers' equation}\label{sec:supp_burgers}

\begin{comment}
The training data is generated with central finite difference and Godunov schemes for viscous and inviscid cases respectively. A uniform spatial discretization $\Delta x = 0.02 \ \si{m}$ and time marching time step $\Delta t = 0.001 \ \si{s}$ are maintained across all training cases. For test cases, we consider $\nu = 0.0001, \ 0.005$ and $\nu = 0.2$. The spatial discretization and time marching time step for the former are $\Delta x = 0.002 \ \si{m}$ and $\Delta t = 0.0001 \ \si{s}$; for the latter, $\Delta x = 0.02 \ \si{m}$ and $\Delta t = 0.0001 \ \si{s}$. All temporal sequences are downsampled by 100x to form the training dataset and testing dataset.

LSTM and DenseNet architectures are adopted for \cn{} and \pn{} respectively. A comparison of adopting FNO and DeepONet as \pn{} is presented (see Fig.~\ref{fig:Burgers'}), demonstrating that DenseNet achieves superior performance.

\begin{figure}[htbp]
    \centering
    \includegraphics[width=\linewidth]{figure_bistable.pdf}
    \caption{}
    \label{fig:Burgers'}
\end{figure}
\end{comment}

\paragraph{Simulation details}
The dataset is generated by solving the one-dimensional viscous Burgers' equation $\partial_t u + u \partial_x u = \nu \partial_{xx} u$ on a periodic domain $\Omega = [-1, 1]$ with a spatial resolution of 4001 grid points. For the viscous cases ($\nu > 0$), an IMEX scheme is employed, with explicit central-difference advection with implicit spectral diffusion under adaptive CFL-constrained timestepping. For the inviscid case ($\nu = 0$), a Godunov scheme with MUSCL reconstruction (minmod limiter) and TVD Runge-Kutta 2nd-order time integration is used. The solution is integrated up to $t = 5$ s and sampled every 50 steps, yielding 101 temporal snapshots. The spatial output is downsampled to 401 points. The viscosity coefficient $\nu$ takes 16 values: $\nu = 0$ and $\nu \in {1, 2, 3, 4, 5} \times 10^{-4, -3, -2}$, giving 16 distinct simulation instances per initial condition. 

\paragraph{Choice of decoder}\label{sec:supp_burgers_pe_ablation}
Two requirements shape the decoder choice. First, the decoder must supervise the encoder with signals from across the entire $(x, t)$ field, not from the immediate next step: under shock dynamics the field is smooth almost everywhere and the shock front (where $\nu$ actually appears) occupies a small fraction of the domain, so one-step supervision is dominated by smooth-region MSE that varies little with $\nu$ and leaves the encoder no $\nu$-dependent gradient. We therefore use a single-shot full-field decoder, namely DeepONet (branch-and-trunk MLP), shared between NODnet and PNO. Second, within the DeepONet the trunk must express the shock fronts at arbitrary continuous $(x, t)$; coordinate-input MLPs smooth them out under their low-frequency spectral bias~\cite{rahaman2019spectral,tancik2020fourier}, so we equip the trunk with sinusoidal positional encoding (PE).

Given the initial frame $\mathbf{u}_0$ and the latent $c$, the branch produces a latent vector $b(\mathbf{u}_0, c) \in \mathbb{R}^{512}$; the trunk takes the PE-encoded coordinates $\gamma(x, t) \in \mathbb{R}^{26}$ and produces a basis $\tau(\gamma(x, t)) \in \mathbb{R}^{512}$; the prediction is $\hat{u}(t, x) = u_0(x) + \alpha_{\text{res}}\,\langle b, \tau\rangle + b_{\text{out}}$, with $\alpha_{\text{res}}$ a learnable residual scale and $b_{\text{out}}$ a learnable bias. Both branch and trunk are four-layer Softplus MLPs (hidden $800$ / $1100$ respectively). The PE follows the standard NeRF form~\cite{tancik2020fourier}: each coordinate $c \in \{x, t\}$ is mapped to $\gamma(c) = [c, \sin(2^k\pi c), \cos(2^k\pi c)]_{k=0}^{L_c-1}$, with $L_x{=}8$ matched to the spatial Nyquist of the 401-grid and $L_t{=}4$ matched to the temporal smoothness of Burgers' dynamics. Training uses $u$-MSE on $8\,192$ randomly sampled $(t, x)$ queries per case. ANO is a separate architecture (1D U-Net autoregressive stepper with a four-frame past-state window) used as the LB and is described in Methods of the main text.

Both requirements leave direct empirical signatures (Table~\ref{tab:supp_pe_ablation}, Fig.~\ref{fig:supp_pe_ablation_codes}). Removing PE preserves the supervision geometry but breaks the representational class: the trunk MLP smooths shocks, short-horizon prediction error grows by $4$--$5\times$, and the encoder's latent distribution collapses with all nine trained $\nu$ mapping to the same scalar within training noise. NODnet-AR (main text) preserves the representational class but breaks the supervision geometry: the same encoder and latent, but a U-Net autoregressive stepper supervises only one step at a time, where smooth-region MSE dominates the loss, and the latent again collapses (Fig.~2(d) of the main text). Both routes leave the encoder without a $\nu$-dependent gradient. Encoder identifiability therefore requires both, and the DeepONet+PE combination provides both.

\begin{table}[hbtp]
    \centering
    \caption{Removing-PE ablation on NODnet and PNO for Burgers' ($L_2$ relative error $\times 10^{-2}$ at 1-, 5-, and 50-step horizons). Same encoder, same loss, same training schedule; the only change is whether the trunk's $(x, t)$ input is PE-encoded. All four rows are single-seed (seed=1234) since this is a focused mechanism ablation; the with-PE row is therefore the seed=1234 reference of Table~1's NODnet-DeepONet, the row whose multi-seed mean shifts modestly to $17.62/17.71/20.34$ at $H{=}50$ (Table~\ref{tab:supp_dsweep}). The seed-pass does not change the qualitative gap to the no-PE row ($\sim 3\times$ at $H{=}50$) or the encoder-collapse signature in Fig.~\ref{fig:supp_pe_ablation_codes}.}
    \begin{tabular}{lcccc}
        \toprule
        Model & Subset & 1-step & 5-step & 50-step \\
        \midrule
        NODnet (with PE)
            & ID            & $4.26{\pm}1.74$ & $6.44{\pm}2.93$  & $12.80{\pm}6.33$ \\
            & OOD           & $4.37{\pm}1.67$ & $6.24{\pm}2.67$  & $12.42{\pm}6.48$ \\
            & OOD-inviscid  & $4.59{\pm}1.34$ & $8.30{\pm}3.68$  & $16.90{\pm}5.24$ \\
        \midrule
        NODnet (no PE, raw $(x,t)$ trunk)
            & ID            & $20.36{\pm}5.89$ & $27.86{\pm}8.54$ & $43.11{\pm}26.34$ \\
            & OOD           & $20.42{\pm}5.97$ & $27.77{\pm}8.41$ & $44.93{\pm}25.80$ \\
            & OOD-inviscid  & $20.27{\pm}5.60$ & $29.48{\pm}7.24$ & $41.52{\pm}15.28$ \\
        \midrule
        PNO$^{\uparrow}$ (with PE)
            & ID            & $3.89{\pm}1.74$ & $6.04{\pm}2.58$  & $13.43{\pm}6.39$ \\
            & OOD           & $4.03{\pm}1.76$ & $5.79{\pm}2.43$  & $12.85{\pm}6.20$ \\
            & OOD-inviscid  & $4.19{\pm}1.19$ & $7.68{\pm}3.12$  & $16.90{\pm}5.70$ \\
        \midrule
        PNO$^{\uparrow}$ (no PE)
            & ID            & $14.69{\pm}4.61$ & $21.94{\pm}8.01$ & $30.91{\pm}14.09$ \\
            & OOD           & $14.63{\pm}4.75$ & $21.64{\pm}8.18$ & $30.59{\pm}14.06$ \\
            & OOD-inviscid  & $15.55{\pm}4.46$ & $24.26{\pm}7.61$ & $35.01{\pm}12.42$ \\
        \bottomrule
    \end{tabular}
    \label{tab:supp_pe_ablation}
\end{table}

\begin{figure}[hbtp]
    \centering
    \includegraphics[width=\linewidth]{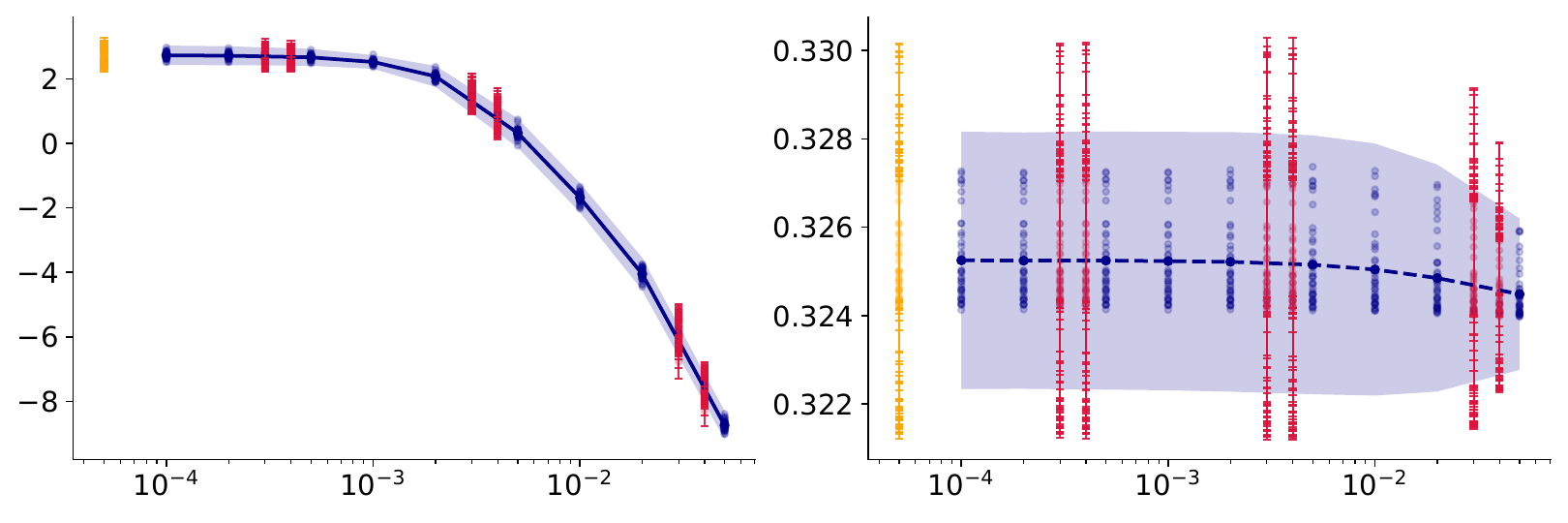}
    \caption{NODnet encoder latent distribution by viscosity, with and without PE on the DeepONet trunk (same CNN encoder, same $u$-only MSE, same training schedule on both panels). Left: with PE, the encoder learns a clean monotonic $\nu \to c$ curve (within-$\nu$ std small, inter-$\nu$ spread large), and the inviscid limit $\nu{=}0$ extends the curve smoothly upward. Right: without PE, the same encoder collapses to a constant latent regardless of $\nu$ (the entire $y$-axis range is $\sim 10^{-3}$, comparable to within-$\nu$ noise). Removing PE thus disables both the decoder's shock representation and, via gradient feedback, the encoder's $\nu$ discrimination.}
    \label{fig:supp_pe_ablation_codes}
\end{figure}

\paragraph{NODnet-AR architectural details}\label{sec:supp_burgers_arablation}
NODnet-AR (reported alongside NODnet in Table~1 of the main text) shares NODnet's CNN encoder and uses an alternative decoder: a 1D U-Net autoregressive stepper conditioned via FiLM injection of the latent at each resolution level, $\hat{\mathbf{u}}_{t+1} = \mathbf{u}_t + g_\theta(\mathbf{u}_t, c)$, rolled out step-by-step at inference. The U-Net has 5 resolution levels with base-channels $44$, matching the parameter count of the DeepONet decoder used by NODnet/PNO. PE does not apply to this decoder: the AR stepper operates on the discrete spatial grid rather than continuous coordinates. Training is $u$-MSE on adjacent $(\mathbf{u}_t, \mathbf{u}_{t+1})$ pairs with rollout depth $4$. Both NODnet and NODnet-AR receive identical conditioning information through the same encoder; the comparison isolates the decoder architectural choice. The autoregressive choice is the source of the encoder collapse seen in main-text Fig.~2d: one-step AR supervision restricts the loss to local increments dominated by smooth-region MSE, eliminating the $\nu$-dependent gradient the encoder needs to separate viscosities (mechanism detailed under \emph{Choice of decoder} above).

\paragraph{PNO calibration: $\sqrt{\nu}$ versus $\log\nu$}
The PNO$^{\uparrow}$ replaces the learned latent with an analytic transform $c = \alpha\sqrt{\nu}+\beta$ of the true viscosity, with $(\alpha, \beta)$ calibrated so that the nine trained viscosities map to $[-2, +2]$ and the inviscid case $\nu{=}0$ maps to $\beta \approx -2.19$, a finite latent just below the smallest trained latent and inside the smooth extension of the calibrated interval. The $\sqrt{\nu}$ form is chosen over the more common $\log\nu$: $\log$ sends $\nu{=}0$ to $-\infty$ and forces an arbitrary $\varepsilon$-clamp that lands the inviscid latent far outside calibration, whereas $\sqrt{\nu}$ remains finite at the singular limit by construction. Inviscid prediction under either NODnet or $\sqrt{\nu}$-PNO is therefore reached by continuation along the calibrated latent curve, in contrast to the textbook $\log+\varepsilon$ encoding which sends $\nu{=}0$ to a latent value never seen during training.

\paragraph{Latent-dimension sweep (supports Fig.~2(a))}
The minimal-conditioning principle predicts that the latent dimension $d$ should saturate at the number of independent governing factors, so $d^*{=}1$ for Burgers' since viscosity is the only governing factor. To probe both sides of the elbow, we train NODnet at $d \in \{0, 1, 2, 3, 4\}$ over three independent seeds $\{1234, 5678, 9012\}$ under identical training settings (same CNN encoder for $d{\geq}1$, same DeepONet+PE decoder, same $u$-only MSE, $200$ epochs each); the $d{=}0$ run drops the encoder entirely and conditions the decoder only on the initial state, recovering the naive single-system neural-operator baseline that ignores per-instance variation. The expected pattern is two-sided: at $d{<}d^*$ the decoder receives no factor information and falls back to family-averaged dynamics; at $d{>}d^*$ test MSE surges because spare dimensions absorb initial-condition-specific structure that does not generalise, the structural-slack mechanism described in the Discussion of the main text. Numerical values are in Table~\ref{tab:supp_dsweep}. The two-stage rule from the main-text Methods is applied as follows. \textit{Stage~1.} Seed-mean Train MSE values $2.25, 1.01, 1.42, 1.38, 1.02$ ($\times 10^{-3}$) for $d=0,\ldots,4$ give step-wise relative changes $-55\%, +41\%, -3\%, -26\%$. The $d{=}1{\to}d{=}2$ change of $+0.41$ is comparable to the $d{=}2$ across-seed envelope ($\sigma_s = 0.74$), so the candidate set is $\{1, 2\}$. \textit{Stage~2.} The trained $d{=}2$ encoder's per-instance latent means populate a spurious second axis (PCA variance ratios with PC2 carrying within-instance shock-front noise rather than $\nu$, mirroring the CFW $d{=}2$ pattern in Fig.~\ref{fig:supp_d2_latents_cfw}), while $d{=}1$ produces a clean monotonic $\nu$-curve (main-text Fig.~2d). Stage~2 therefore picks $d^*{=}1$. The non-monotonicity of long-horizon error past $d^*$ ($d{=}3$ marginally below $d{=}1$ on $H{=}50$ within $\pm 1\sigma_s$, Table~\ref{tab:supp_dsweep}) is consistent with structural slack scattering across multiple axes at $d{>}2$; the latent space verification rules out using long-horizon argmin as the selection criterion.

\begin{table}[hbtp]
    \centering
    \caption{Latent-dimension sweep on NODnet for Burgers' ($L_2$ relative error $\times 10^{-2}$ at $H{=}1, 5, 50$; last-50-epoch train MSE $\times 10^{-3}$). Cells are reported as $\mu \pm \sigma_w \scriptscriptstyle{\pm \sigma_{s,s}}$, where $\sigma_w$ is within-eval std on the test cases (single-seed) and $\sigma_s$ is across-seed std over $\{1234, 5678, 9012\}$. Identical training to the main-text NODnet row except for the latent dimension $d$; the $d{=}0$ row drops the encoder and conditions the decoder only on the initial state. Bold marks the lowest seed-mean in each column; the two-stage rule (Sec.~S2.1 paragraph above) selects $d^*{=}1$ as the smallest plateau dimension passing the latent space verification.}
    \begin{tabular}{cllcccc}
        \toprule
        $d$ & Params & Subset & 1-step & 5-step & 50-step & Train MSE \\
        \midrule
        0 & 5.03M  & ID            & $5.14{\pm}2.49\ensuremath{\,{\scriptscriptstyle\pm 0.21_s}}$ & $12.78{\pm}9.15\ensuremath{\,{\scriptscriptstyle\pm 0.16_s}}$ & $27.62{\pm}25.78\ensuremath{\,{\scriptscriptstyle\pm 0.99_s}}$ & $2.25\ensuremath{\,{\scriptscriptstyle\pm 0.01_s}}$ \\
          &        & OOD           & $5.38{\pm}2.47\ensuremath{\,{\scriptscriptstyle\pm 0.16_s}}$ & $14.05{\pm}9.04\ensuremath{\,{\scriptscriptstyle\pm 0.08_s}}$ & $30.35{\pm}25.83\ensuremath{\,{\scriptscriptstyle\pm 0.87_s}}$ & --- \\
          &        & OOD-inviscid  & $5.29{\pm}1.46\ensuremath{\,{\scriptscriptstyle\pm 0.29_s}}$ & $12.90{\pm}4.83\ensuremath{\,{\scriptscriptstyle\pm 0.42_s}}$ & $24.98{\pm}9.35\ensuremath{\,{\scriptscriptstyle\pm 0.76_s}}$ & --- \\
        \midrule
        1 & 10.08M & ID            & $\mathbf{4.46{\pm}1.84}\ensuremath{\,{\scriptscriptstyle\pm 0.31_s}}$ & $\mathbf{7.56{\pm}3.14}\ensuremath{\,{\scriptscriptstyle\pm 1.76_s}}$ & $17.62{\pm}9.90\ensuremath{\,{\scriptscriptstyle\pm 8.52_s}}$ & $\mathbf{1.01}\ensuremath{\,{\scriptscriptstyle\pm 0.03_s}}$ \\
          &        & OOD           & $\mathbf{4.59{\pm}1.82}\ensuremath{\,{\scriptscriptstyle\pm 0.32_s}}$ & $\mathbf{7.52{\pm}3.02}\ensuremath{\,{\scriptscriptstyle\pm 2.05_s}}$ & $17.71{\pm}10.54\ensuremath{\,{\scriptscriptstyle\pm 9.36_s}}$ & --- \\
          &        & OOD-inviscid  & $\mathbf{4.74{\pm}1.39}\ensuremath{\,{\scriptscriptstyle\pm 0.27_s}}$ & $\mathbf{9.17{\pm}3.83}\ensuremath{\,{\scriptscriptstyle\pm 1.44_s}}$ & $20.34{\pm}6.99\ensuremath{\,{\scriptscriptstyle\pm 5.74_s}}$ & --- \\
        \midrule
        2 & 10.08M & ID            & $4.82{\pm}2.10\ensuremath{\,{\scriptscriptstyle\pm 0.38_s}}$ & $9.58{\pm}5.23\ensuremath{\,{\scriptscriptstyle\pm 2.85_s}}$ & $21.67{\pm}15.56\ensuremath{\,{\scriptscriptstyle\pm 6.47_s}}$ & $1.42\ensuremath{\,{\scriptscriptstyle\pm 0.74_s}}$ \\
          &        & OOD           & $4.99{\pm}2.09\ensuremath{\,{\scriptscriptstyle\pm 0.42_s}}$ & $9.97{\pm}5.10\ensuremath{\,{\scriptscriptstyle\pm 3.64_s}}$ & $22.56{\pm}15.75\ensuremath{\,{\scriptscriptstyle\pm 7.97_s}}$ & --- \\
          &        & OOD-inviscid  & $5.04{\pm}1.46\ensuremath{\,{\scriptscriptstyle\pm 0.33_s}}$ & $10.67{\pm}4.31\ensuremath{\,{\scriptscriptstyle\pm 2.05_s}}$ & $22.69{\pm}8.40\ensuremath{\,{\scriptscriptstyle\pm 3.75_s}}$ & --- \\
        \midrule
        3 & 10.08M & ID            & $4.69{\pm}2.01\ensuremath{\,{\scriptscriptstyle\pm 0.24_s}}$ & $8.29{\pm}4.07\ensuremath{\,{\scriptscriptstyle\pm 1.54_s}}$ & $\mathbf{17.18{\pm}10.84}\ensuremath{\,{\scriptscriptstyle\pm 3.29_s}}$ & $1.38\ensuremath{\,{\scriptscriptstyle\pm 0.63_s}}$ \\
          &        & OOD           & $4.88{\pm}2.00\ensuremath{\,{\scriptscriptstyle\pm 0.32_s}}$ & $8.53{\pm}4.04\ensuremath{\,{\scriptscriptstyle\pm 1.91_s}}$ & $\mathbf{17.39{\pm}11.38}\ensuremath{\,{\scriptscriptstyle\pm 3.54_s}}$ & --- \\
          &        & OOD-inviscid  & $4.92{\pm}1.41\ensuremath{\,{\scriptscriptstyle\pm 0.08_s}}$ & $9.55{\pm}4.03\ensuremath{\,{\scriptscriptstyle\pm 0.93_s}}$ & $\mathbf{19.55{\pm}6.65}\ensuremath{\,{\scriptscriptstyle\pm 2.09_s}}$ & --- \\
        \midrule
        4 & 10.08M & ID            & $4.89{\pm}1.97\ensuremath{\,{\scriptscriptstyle\pm 0.22_s}}$ & $9.15{\pm}3.81\ensuremath{\,{\scriptscriptstyle\pm 1.04_s}}$ & $22.46{\pm}13.77\ensuremath{\,{\scriptscriptstyle\pm 2.40_s}}$ & $1.02\ensuremath{\,{\scriptscriptstyle\pm 0.02_s}}$ \\
          &        & OOD           & $5.01{\pm}1.97\ensuremath{\,{\scriptscriptstyle\pm 0.23_s}}$ & $9.15{\pm}3.70\ensuremath{\,{\scriptscriptstyle\pm 0.99_s}}$ & $22.88{\pm}14.36\ensuremath{\,{\scriptscriptstyle\pm 2.16_s}}$ & --- \\
          &        & OOD-inviscid  & $5.11{\pm}1.51\ensuremath{\,{\scriptscriptstyle\pm 0.18_s}}$ & $10.47{\pm}4.26\ensuremath{\,{\scriptscriptstyle\pm 0.91_s}}$ & $23.60{\pm}9.25\ensuremath{\,{\scriptscriptstyle\pm 2.17_s}}$ & --- \\
        \bottomrule
    \end{tabular}
    \label{tab:supp_dsweep}
\end{table}

\subsection{Details of flow past a cylinder}\label{sec:supp_cfw}

% Critical transition can be analyzed in the framework of Grad-CAM like structure. By formulating the difference of upper and lower half of the figure as the indicator for the instability of the flow past cylinder, we could examine this indicator's sensitivity with respect to the whole input field (with irregular geometry) through the one bp process thanks to the point-wise transformer network. We can also examine the activation heat map of each attention layer.

% Critical transition with meta-learning
% 3rd order spring critical transition -- PRL

% First step would be to visualize the indicator's temporal trajectory. We find that our model is very accurate in this 1st order quantity (in H1 space, doubtful) although our model is purely trained on 0th order loss. 

\paragraph{Simulation details}
The dataset is a simulation on an irregular mesh on a 2D domain $\Omega=[-9,21]\times[-6,6]$ and boundary of the cylinder on a circle with radius $0.5$ center at origin, with 159,720 points in total. The time span is $1.2$ s, and integration timestep is $0.00001$ s. Data is downsampled every 1000 steps, forming 120 steps. $\nu$ species a system, which takes the value of $[0.005,0.008,0.011,0.014,0.017,0.020]$. For each $\nu$, six $u_\infty$ are simulated, which are $[15,16,17,18,19,20]$. The dataset consists of 36 trajectories in total. There are 16 trajectories in the training dataset, which are $[\nu,u_\infty]_{\text{train}}=[0.005,0.008,0.014,0.02]\times[15,17,19,20]$; so the training data is limited. The first test set consists of 8 trajectories $[\nu,u_\infty]_{\text{test,1}}=[0.005,0.008,0.014,0.02]\times[16,18]$. It tests the model capability to generate correct $\boldsymbol{\eta}$ and correct prediction on unseen velocities for trained $\nu$. The second test set consists of another 12 trajectories $[\nu,u_\infty]_{\text{test,2}}=[0.011,0.017]\times[15,16,17,18,19,20]$. It tests the model to generalize to unseen systems with different $\nu$ from the training set, given arbitrary velocities.

\paragraph{Formulation of SupCL}

\begin{equation}
\ell_i = - \frac{1}{|P(i)|} \sum_{p \in P(i)} \log \frac{\exp\bigl(\text{sim}(\mathbf{z}_i, \mathbf{z}_p)/\tau\bigr)}{\sum_{\substack{a=1 \\ a \neq i}}^{N} \exp\bigl(\text{sim}(\mathbf{z}_i, \mathbf{z}_a)/\tau\bigr)},
\end{equation}
where $\text{sim}(\mathbf{z}_i, \mathbf{z}_j)$ is a similarity function. We employ $L_2$ distance as the similarity metric, a choice motivated by the simplicity of our one-dimensional latent space. In this formulation, $\tau$ denotes the temperature parameter and $N$ represents the total number of samples within the batch. To gradually shift the optimization focus, we anneal the temperature over time, reducing the model's reliance on the supervised contrastive loss (SupCL) and transitioning toward the standard training objective. Additionally, we inject a small amount of random noise into the generated $\phi$ to mitigate mode collapse.
\paragraph{Rollout augmentation}
We train on 2 step rollout result (1 step prediction with 1 step rollout $\mathbf{x}_{t+2} = \mathbf{f}^{(2)}(\mathbf{x}_t,\boldsymbol{\eta})$) to increase the training stability.

\paragraph{Drag force computation}
\begin{equation}
\begin{split}
    \boldsymbol{\sigma} &= -p \mathbf{I} + \mu (\nabla \mathbf{u} + \nabla \mathbf{u}^\top),\\
    F_D &= \int_{\partial \Omega_c} \boldsymbol{\sigma}\mathbf{n}\cdot \hat{\mathbf{e}}_x , ds, \\ &=
    -\int_{\partial \Omega_c} p n_x  ds
    +
    \mu \int_{\partial \Omega_c}
    \left[(\nabla \mathbf{u} + \nabla \mathbf{u}^\top)\mathbf{n}\right]_x
    ds.
\end{split}
\end{equation}
In our dataset, velocity predictions are available on scattered spatial points. To compute surface forces consistently across ground truth, NODnet, and Baseline, the scattered velocity field is interpolated onto a narrow band of points surrounding the cylinder surface to approximate velocity gradients and surface normals; since pressure is not directly predicted, we recover it by solving the pressure Poisson equation derived from incompressible Navier–Stokes with Neumann boundary condition on the cylinder surface consistent with the normal momentum balance. This ensures that the reconstructed pressure field yields physically consistent surface traction. This procedure ensures that drag evaluation is consistent with the governing equations and avoids biases arising from missing pressure information.

\paragraph{Strouhal number computation}
The Strouhal number $\mathrm{St} = f D / u_\infty$ (with $f$ the vortex shedding frequency, $D$ the cylinder diameter, $u_\infty$ the inlet velocity) is the dimensionless analogue of the shedding frequency in the laminar wake. We extract $f$ from the lift coefficient time series $C_L(t)$ via FFT and take the peak frequency in the band consistent with vortex shedding ($\sim 0.2\, u_\infty/D$ at the Reynolds numbers considered here). $C_L$ is computed from the same surface integral as $C_D$ but along the transverse direction $\hat{\mathbf{e}}_y$. Together, $\bar{C}_D$ and $\mathrm{St}$ characterize the two canonical features of the laminar wake: the time-averaged momentum transfer and the periodic vortex emission. A predictor that recovers both is reproducing the dynamics; a predictor that drifts in either is failing in a physically interpretable way.

\paragraph{Architectural inductive bias}

We adopt the same autoregressive encoder and decoder setting as the FN-RD system. Our parametric studies (4-step rollout averaged $L_2$ prediction error, Table~\ref{tab:param_study}) indicate that a ``Heavy-Encode'' architecture, in which the encoder $f_\phi$ has significantly higher capacity than the decoder $g_\theta$, yields superior training stability and generalization. This is consistent with the minimal conditioning principle: identifying which system one is observing is the more demanding task, while evolving the state, once the governing factors are compactly available, is relatively simpler. We additionally use a multi-step rollout augmentation strategy, training on two-step rollout outputs ($\mathbf{x}_{t+2}=g_\theta^{(2)}(\mathbf{x}_t,c)$, $\hat{\mathbf{y}}=(\mathbf{x}_{t+1},\mathbf{x}_{t+2})$), to enforce temporal stability and mitigate numerical drift during long-horizon inference.

\begin{table}[hbtp]
    \centering
    \caption{Architecture parametric experiments for the cylinder flow wake case. 4-step rollout averaged $L_2$ prediction error across train and test sets.}
    \begin{tabular}{cccccc}
        \toprule
        Mode & Parameter & Train & Unseen $u_\infty$ & Unseen $\nu$ & Performance\\
        \hline
        Balanced & 23M & \textbf{0.023} & 0.084 & 0.042 & Overfit
        \\
        Heavy Encode & \textbf{15M} & 0.027 & \textbf{0.030} & \textbf{0.028} & \textbf{Generalize}
        \\
        Heavy Decode & 42M & 0.025 & 0.089 & 0.052 & Overfit
        \\
        \bottomrule
    \end{tabular}
    \label{tab:param_study}
\end{table}

\paragraph{Latent-dimension sweep}
The cylinder wake is parameterized by viscosity $\nu$ and inlet velocity $u_\infty$, but at the Reynolds numbers considered here ($u_\infty \in [15,20]$, $\nu \in [0.005,0.020]$) the wake dynamics are governed effectively by the single dimensionless group $\mathrm{Re}=u_\infty D/\nu$. The minimal-conditioning principle therefore predicts $d^*{=}1$. To probe both sides of the elbow, we evaluate the heavy-encode model with $d \in \{0, 1, 2, 3, 4\}$ on the same Train / Unseen-$u_\infty$ / Unseen-$\nu$ splits used in Table~\ref{tab:param_study}, at horizons $H \in \{1,5,50\}$, with three independent training seeds $\{1234, 5678, 9012\}$ at every $d$. The $d{=}0$ run drops the encoder (3.46M params) and conditions the decoder only on the initial state, recovering the naive single-system neural-operator baseline. For $d \geq 1$ we use the same encoder/decoder shape and $\sim$15M parameter budget; only latent dimension $d$ differs (contributing $\sim$1k parameters per increment). The training schedule uses a warmup+cosine latent-noise schedule (constant $5{\times}10^{-2}$ for the first 200 epochs, then cosine decay to $0$ over the remaining 800 epochs; $1{\times}10^{-1}$ peak for $d{=}3$ to overcome an encoder-collapse failure mode at the lower noise level), and at $d{=}1$ the supervised-contrastive weight is downweighted to $0.3{\times}\text{rep\_force}$ because the full-strength contrastive loss is geometrically over-constrained on a 1-D latent (36 system labels cannot be maximally separated on a line). Numerical values are in Table~\ref{tab:supp_dsweep_cfw} (the same sweep is visualised in main-text Fig.~3a).

\textit{Why CFW needs a long-horizon (or OOD) gauge.} Unlike Burgers' and FN-RD, the $1$-step Train L$_2$RE for CFW does not elbow at the predicted $d^*{=}1$: seed-mean values $1.47, 1.69, 2.49, 1.91, 2.07$ ($\times 10^{-2}$) for $d=0,\ldots,4$ give step-wise relative changes $+15\%, +47\%, -23\%, +8\%$, with $d{=}0$ the lowest. This is the encoder-free overfit signature: at $d{=}0$ the decoder receives a $19\,\text{M}$-parameter budget free of any latent-conditioning bottleneck and absorbs all family variability into per-step memorisation of the training distribution, beating $d{=}1$ on the very horizon at which the bottleneck has the least room to pay off. The cost of that flexibility surfaces as soon as the model has to roll out (errors compound system-agnostically) or generalise (no latent to interpolate over): on Train $H{=}50$ the seed-means are $36.09, 29.89, 39.40, 38.42, 31.52$ (step-wise $-17\%, +32\%, -2\%, -18\%$), and on OOD $H{=}5$ they are $6.11, 4.38, 11.30, 9.17, 7.88$ (step-wise $-28\%, +158\%, -19\%, -14\%$). Both long-horizon gauges minimise at $d{=}1$.

Applying the two-stage rule. \textit{Stage~1.} On the Train $H{=}50$ gauge, $d{=}1$ is the minimum and $d{=}1{\to}d{=}2$ jumps by $+32\%$ ($\sim 9.5 \times \sigma_s$, well outside seed envelope), so the candidate set is $\{1, 2\}$. \textit{Stage~2.} The trained $d{=}2$ encoder populates a clear second axis (PCA variance ratios $(0.74, 0.26)$, Fig.~\ref{fig:supp_d2_latents_cfw}), absorbing within-system / IC-window variability rather than carrying a second governing factor; the $d{=}1$ encoder produces a clean monotonic $\nu$-curve with signal-to-noise $\sim 36$ (Fig.~\ref{fig:supp_d1_latents_cfw}). Stage~2 selects $d^*{=}1$.

Three patterns are consistent with the principle. (i) $d{=}0$ wins short-horizon ID ($H{=}1, 5$) and OOD-$H{=}1$ through encoder-free overfit, but $d{=}1$ wins all four long-horizon and OOD-$H{=}5$ cells: the encoder bottleneck pays off precisely at the rollout horizons where slack matters. (ii) $d{=}0$ at $H{=}50$ on OOD ($34.5\%$) sits below the entire $d{>}1$ curve and within the seed envelope of $d{=}1$ ($31.0\%$): the structural-slack mechanism in compact form. $d{=}0$ has no pathway for IC-specific noise to enter the latent and drift the rollout, whereas $d{>}1$ provides exactly such a pathway through its spare dimensions, so at long horizons ``no encoder'' beats ``encoder with slack''. (iii) The $d{>}1$ curve is non-monotonic but strictly above $d{=}1$ on Train $H{=}50$, with $d{=}2$ the worst (OOD-$H{=}5$ rises by a factor of $\sim 2.6$ over $d{=}1$, the ``surge past sufficient capacity'' signature seen in Burgers'), while $d{=}3, 4$ partially recover at long horizons by spreading the spare dimensions into a near-degenerate submanifold but never beat $d{=}1$ on the long-horizon Train metric.

\begin{table}[hbtp]
    \centering
    \caption{Latent-dimension sweep on the heavy-encode CFW model ($L_2$ relative error $\times 10^{-2}$, at $H{=}1, 5, 50$). Cells reported as $\mu \pm \sigma_w \scriptscriptstyle{\pm \sigma_{s,s}}$; $\sigma_w$ is within-eval std on the test cases (single-seed) and $\sigma_s$ is across-seed std over $\{1234, 5678, 9012\}$. ID = Train split; OOD = Unseen-$u_\infty$ and Unseen-$\nu$ pooled. \textbf{Bold marks the lowest seed-mean in each column}: $d{=}0$ wins short-horizon ID through encoder-free overfit; $d{=}1$ wins long-horizon and OOD-$H{=}5$ --- the elbow that motivates $d^*{=}1$.}
    \begin{tabular}{cllccc}
        \toprule
        $d$ & Params & Subset & $H{=}1$ & $H{=}5$ & $H{=}50$ \\
        \midrule
        0 & 3.46M  & ID  & $\mathbf{1.47{\pm}0.59}\ensuremath{\,{\scriptscriptstyle\pm 0.29_s}}$ & $\mathbf{3.46{\pm}1.76}\ensuremath{\,{\scriptscriptstyle\pm 0.81_s}}$ & $36.09{\pm}14.17\ensuremath{\,{\scriptscriptstyle\pm 5.43_s}}$ \\
         (no encoder) &  & OOD & $\mathbf{1.63{\pm}0.75}\ensuremath{\,{\scriptscriptstyle\pm 0.39_s}}$ & $6.11{\pm}3.76\ensuremath{\,{\scriptscriptstyle\pm 0.18_s}}$ & $34.53{\pm}13.43\ensuremath{\,{\scriptscriptstyle\pm 3.01_s}}$ \\
        \midrule
        1 & 15.24M & ID  & $1.69{\pm}0.79\ensuremath{\,{\scriptscriptstyle\pm 0.31_s}}$ & $4.00{\pm}2.60\ensuremath{\,{\scriptscriptstyle\pm 1.23_s}}$ & $\mathbf{29.89{\pm}16.61}\ensuremath{\,{\scriptscriptstyle\pm 6.11_s}}$ \\
          &        & OOD & $1.72{\pm}0.86\ensuremath{\,{\scriptscriptstyle\pm 0.36_s}}$ & $\mathbf{4.38{\pm}2.34}\ensuremath{\,{\scriptscriptstyle\pm 0.45_s}}$ & $\mathbf{31.04{\pm}15.11}\ensuremath{\,{\scriptscriptstyle\pm 4.51_s}}$ \\
        \midrule
        2 & 15.24M & ID  & $2.49{\pm}0.97\ensuremath{\,{\scriptscriptstyle\pm 0.47_s}}$ & $8.75{\pm}5.47\ensuremath{\,{\scriptscriptstyle\pm 4.77_s}}$ & $39.40{\pm}17.01\ensuremath{\,{\scriptscriptstyle\pm 7.35_s}}$ \\
          &        & OOD & $2.84{\pm}1.06\ensuremath{\,{\scriptscriptstyle\pm 0.60_s}}$ & $11.30{\pm}6.13\ensuremath{\,{\scriptscriptstyle\pm 5.41_s}}$ & $38.33{\pm}13.42\ensuremath{\,{\scriptscriptstyle\pm 6.36_s}}$ \\
        \midrule
        3 & 15.24M & ID  & $1.91{\pm}0.74\ensuremath{\,{\scriptscriptstyle\pm 0.43_s}}$ & $5.54{\pm}2.94\ensuremath{\,{\scriptscriptstyle\pm 1.86_s}}$ & $38.42{\pm}15.40\ensuremath{\,{\scriptscriptstyle\pm 6.17_s}}$ \\
          &        & OOD & $2.78{\pm}1.37\ensuremath{\,{\scriptscriptstyle\pm 0.91_s}}$ & $9.17{\pm}5.64\ensuremath{\,{\scriptscriptstyle\pm 2.08_s}}$ & $34.54{\pm}13.67\ensuremath{\,{\scriptscriptstyle\pm 4.36_s}}$ \\
        \midrule
        4 & 15.24M & ID  & $2.07{\pm}0.87\ensuremath{\,{\scriptscriptstyle\pm 0.36_s}}$ & $4.67{\pm}2.24\ensuremath{\,{\scriptscriptstyle\pm 1.44_s}}$ & $31.52{\pm}14.61\ensuremath{\,{\scriptscriptstyle\pm 6.93_s}}$ \\
          &        & OOD & $2.58{\pm}1.25\ensuremath{\,{\scriptscriptstyle\pm 0.72_s}}$ & $7.88{\pm}4.97\ensuremath{\,{\scriptscriptstyle\pm 2.11_s}}$ & $33.03{\pm}14.36\ensuremath{\,{\scriptscriptstyle\pm 3.30_s}}$ \\
        \bottomrule
    \end{tabular}
    \label{tab:supp_dsweep_cfw}
\end{table}

\paragraph{Why does $d{>}1$ regress? Latent-space inspection}
Figures~\ref{fig:supp_d1_latents_cfw}, \ref{fig:supp_d2_latents_cfw}, and~\ref{fig:supp_d3_latents_cfw} render the trained $d{=}1$, $d{=}2$, and $d{=}3$ encoders evaluated on all 36 cylinder-wake systems (16 train + 8 unseen-$u_\infty$ + 12 unseen-$\nu$, with 16 conditioning-window draws per system). The right panels colour each per-system mean by $\nu$. The $d{=}1$ encoder lays out the 36 systems as six tight clusters with one cluster per viscosity, monotonically ordered along the latent axis from low to high $\nu$ (inter-system range $2.10$ vs.\ mean within-system std $0.06$, signal-to-noise ratio $\sim$36); the six $u_\infty$ within each $\nu$ collapse to a single point up to noise, confirming that the encoder has discovered a single governing dimension dominated by $\nu$ in this $\mathrm{Re}$ regime. PCA on the 36 system means then yields:
\begin{itemize}
    \item $d{=}2$: variance ratios $(0.739, 0.261)$. PC2 retains over a quarter of the total variance --- nowhere near the empty-axis degeneracy seen in FN-RD's $d{=}3$ run (variance ratios $(0.80, 0.20, \mathbf{0.00})$, Sec.~\ref{sec:supp_fnrd}). Both available dimensions are actively populated.
    \item $d{=}3$: variance ratios $(0.685, 0.231, 0.083)$. Even the third PC carries an $8\%$ share, so all three dimensions remain populated; the encoder does not spontaneously collapse onto a 1-D submanifold.
\end{itemize}
This is the empirical signature of the structural-slack mechanism. The CFW case is parameterized by a single dominant dimensionless group ($\mathrm{Re}=u_\infty D/\nu$, hence $d^*{=}1$), but neither the SupCL contrastive force nor the rollout-MSE objective penalizes the extra latent axes for being non-empty. Unlike FN-RD, CFW lacks a strong O(2) augmentation that would multiply each system into many equivalent draws and pull degenerate axes to zero during training; the only within-system variability the encoder sees is the conditioning-window subsampling. Consequently, when given two or three latent dimensions, the encoder is free to spread within-system noise and IC-window fluctuations into the slack axes, producing the populated PC2/PC3 we observe. At inference time the decoder must integrate this extra non-governing structure across the 50-step rollout, which is exactly the OOD-degradation pattern in Table~\ref{tab:supp_dsweep_cfw}: the more slack the encoder is given to fill, the further the trajectory drifts from the true-$\mathrm{Re}$ flow. The non-monotonicity ($d{=}2$ being worst, $d{=}3,4$ partially recovering) is also consistent with this picture: at $d{=}3,4$ the slack is divided across multiple axes so that no single axis carries the full burden of IC absorption (PC3 contribution $0.083$, PC2 contribution drops from $0.261$ at $d{=}2$ to $0.231$ at $d{=}3$), reducing the per-axis impact on rollout but not eliminating it.

\begin{figure}[hbtp]
    \centering
    \includegraphics[width=\linewidth]{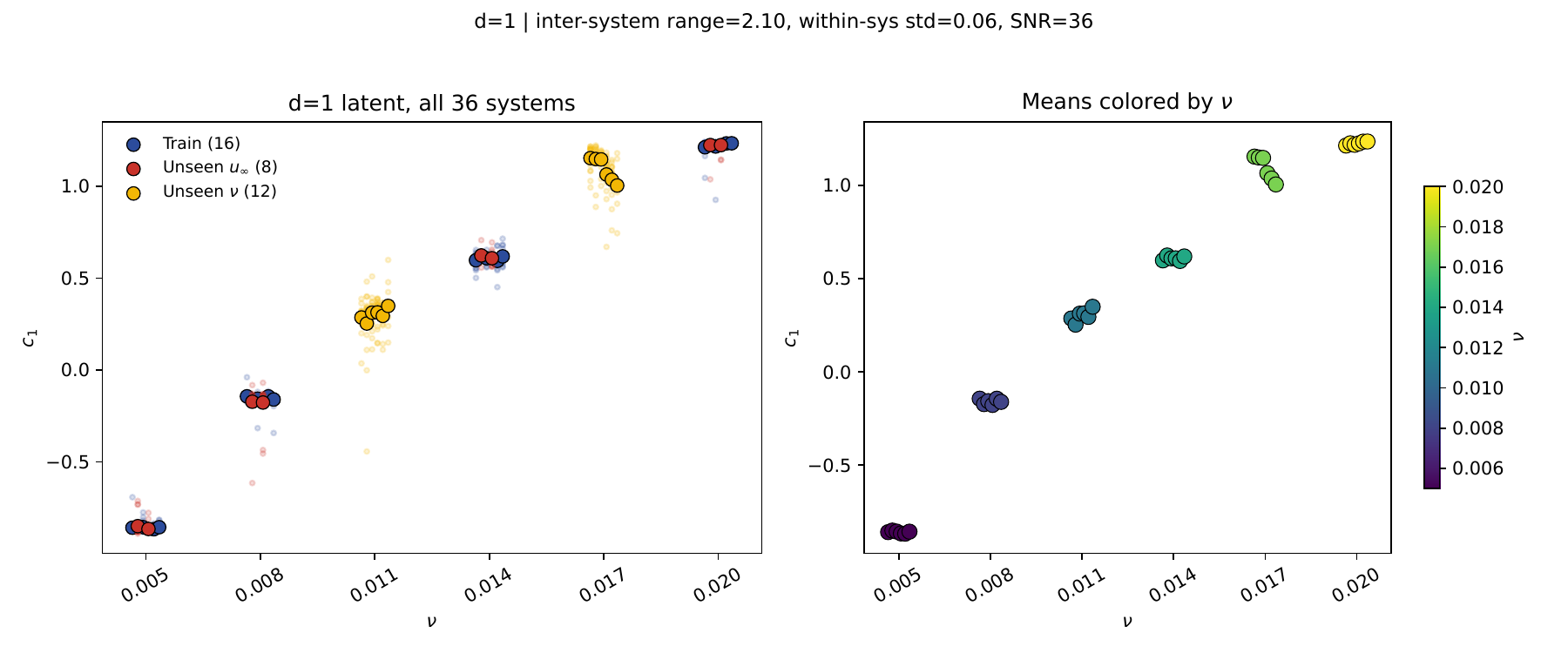}
    \caption{$d{=}1$ latent representation of all 36 CFW systems (the original heavy-encode checkpoint reported in the main text). \textbf{Left:} $\nu$ vs.\ $c_1$ scatter coloured by split (blue: 16 train, red: 8 unseen-$u_\infty$, orange: 12 unseen-$\nu$); faint dots are 16 conditioning-window draws per system, solid dots are per-system means. The $u_\infty$ axis is shown as a small horizontal jitter inside each $\nu$ column. \textbf{Right:} the same per-system means, coloured by $\nu$, on a viridis colourbar. The encoder produces a clean monotonic $\nu \to c_1$ map with $\sim$36$\times$ inter-system / within-system separation, and the six $u_\infty$ values within each $\nu$ collapse to one point --- the discovered 1-D latent aligns with $\nu$ in this $\mathrm{Re}$ regime, contrast directly with the populated slack axes in Figs.~\ref{fig:supp_d2_latents_cfw}, \ref{fig:supp_d3_latents_cfw}.}
    \label{fig:supp_d1_latents_cfw}
\end{figure}

\begin{figure}[hbtp]
    \centering
    \includegraphics[width=\linewidth]{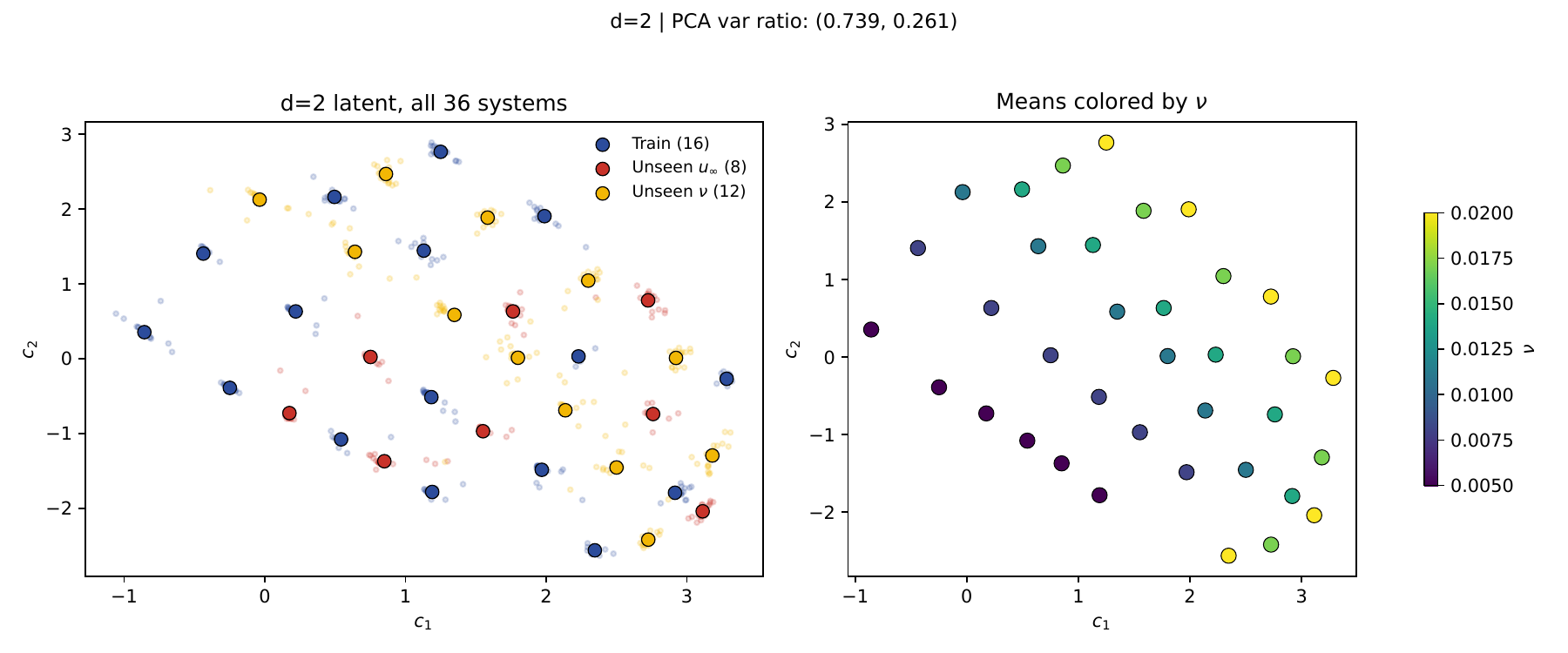}
    \caption{$d{=}2$ latent representation of all 36 CFW systems. \textbf{Left:} 2-D scatter coloured by split (blue: 16 train, red: 8 unseen-$u_\infty$, orange: 12 unseen-$\nu$), faint dots are 16 conditioning-window draws per system, solid dots are per-system means. \textbf{Right:} per-system means coloured by viscosity $\nu$. PCA on the 36 means yields variance ratios $(0.739, 0.261)$: the second axis is clearly populated, in contrast to the empty-axis collapse seen for FN-RD when given more dimensions than $d^*$. This is the structural-slack signature explaining the $H{=}5$ rollout surge in Table~\ref{tab:supp_dsweep_cfw}.}
    \label{fig:supp_d2_latents_cfw}
\end{figure}

\begin{figure}[hbtp]
    \centering
    \includegraphics[width=\linewidth]{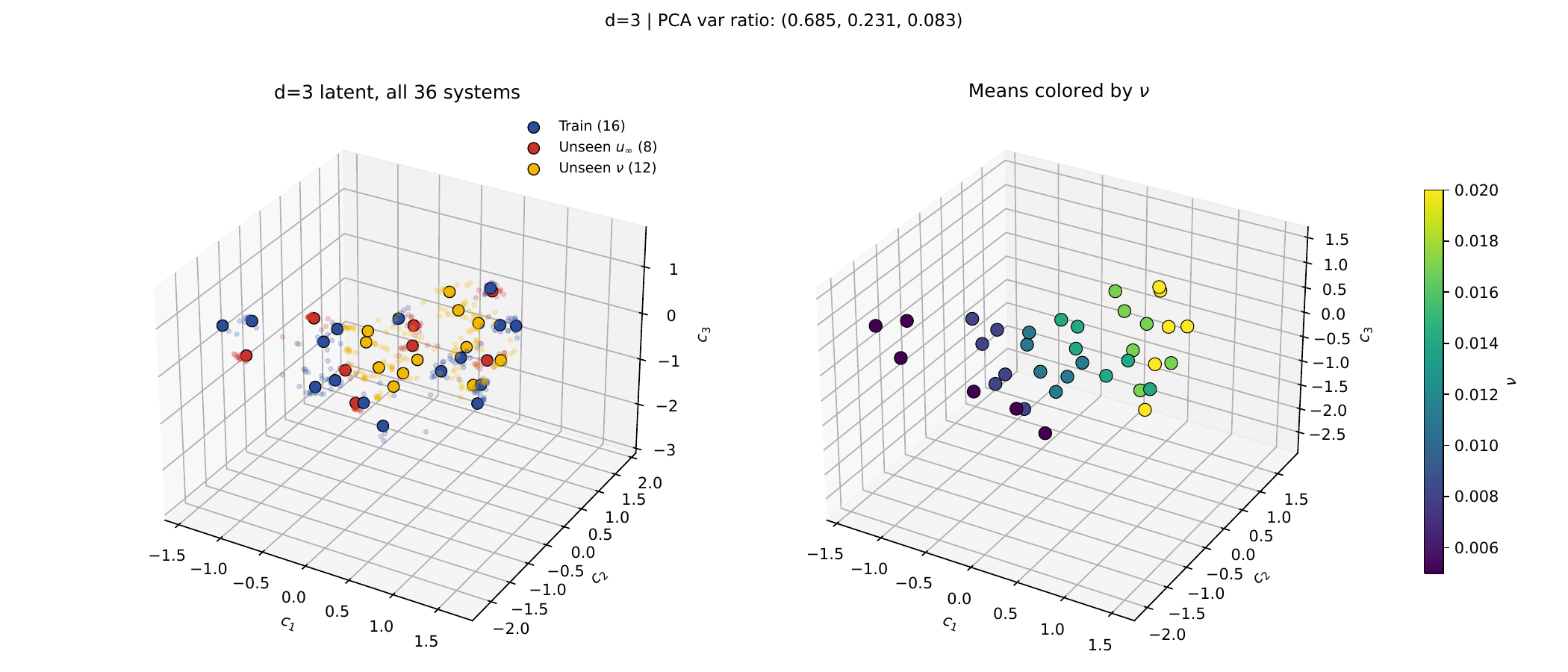}
    \caption{$d{=}3$ latent representation of all 36 CFW systems. Layout matches Fig.~\ref{fig:supp_d2_latents_cfw}: \textbf{left} coloured by split, \textbf{right} per-system means coloured by $\nu$. PCA on the 36 means yields variance ratios $(0.685, 0.231, 0.083)$. Unlike the FN-RD $d{=}3$ run (Sec.~\ref{sec:supp_fnrd}, variance ratios $(0.80, 0.20, 0.00)$, where the trained encoder discovers a 2-D submanifold matching $d^*{=}2$), the CFW encoder uses all three available axes; the third axis still holds $8\%$ of the inter-system variance. The slack is spread rather than collapsed --- consistent with $d^*{=}1$ being the only setting that recovers the original heavy-encode performance.}
    \label{fig:supp_d3_latents_cfw}
\end{figure}

\paragraph{What the SNO/NODnet convergence at $H{=}50$ on OOD does and does not imply.}
The seed-mean comparison $d{=}1$ vs.\ $d{=}0$ across horizons is asymmetric. At $H{=}1, 5$ on ID and $H{=}1$ on OOD, $d{=}0$ is at or below $d{=}1$: the encoder-free baseline absorbs all family variability into per-step memorisation of the training distribution, an overfit that the bottleneck at $d^*$ cannot match by construction. The advantage flips at OOD-$H{=}5$ ($d{=}1$ at $4.38\%$ vs.\ $d{=}0$ at $6.11\%$, a $28\%$ relative reduction) and at long horizons on both splits ($d{=}1$ at $29.9\%$ ID-$H{=}50$ vs.\ $d{=}0$ at $36.1\%$, a $17\%$ reduction; OOD-$H{=}50$ within seed envelope). Two readings support the structural-slack interpretation. \emph{First}, the asymmetric ordering --- $d{=}0$ winning short-horizon ID and $d{=}1$ winning long-horizon and OOD-$H{=}5$ --- is exactly what the bottleneck-as-regulariser claim predicts: $d{=}0$ has no factor-aware latent so it cannot interpolate across instances, but it also has no slack axis for IC noise to drift the rollout, so it catches up at $H{=}50$ through the absence of slack rather than the presence of factor information. \emph{Second}, NODnet's drag and Strouhal-number recovery at $\mathrm{Re}{=}300$ (main-text Fig.~3f) remains physically valid where SNO produces family-averaged drag with no per-instance shedding frequency. The rollout-error metric saturates at long horizons; the physical-observable metric continues to separate the two.

\subsection{Details of FHN reaction-diffusion network}\label{sec:supp_fnrd}

\paragraph{Simulation details}
The dataset is a simulation on a uniform 2D mesh grid in a domain $\Omega=[0,1]^2$ with $\Delta t=0.001$ s up to $t = 10$ s and sampled every 100 timesteps, using finite difference method with the 4th-order Runge-Kutta time integration. $k$ and $\beta$ are tuned to model different neural activities, with $k$ ranging from $0.01$ to $0.05$ and $\beta$ ranging from $0.1$ to $0.3$, giving $25$ distinct system instances.

\paragraph{Design of hierarchical mask}
The prediction rule applies smaller-lead-time steppers first, which suits a diffusion-reaction system whose transient phase happens early. This contrasts with the weather-forecasting setting~\cite{pangu}, where predictions over different horizons can be issued in any order without compounding the early-transient error.

\paragraph{Data augmentation}
To enhance the model's capability to generalize across initial conditions without enlarging the simulation budget, we apply data augmentation to replicate trajectories. The governing equations together with periodic boundary conditions exhibit O(2) equivariance: any spatiotemporal trajectory transformed via the O(2) action remains a valid solution of the same FN-RD system instance. We therefore augment each trajectory by shifting to the four corners, mirroring horizontally and vertically, and rotating by $90^\circ$, $180^\circ$, $270^\circ$, yielding a $10\times$ enlargement of the effective dataset. This enforces the encoder to identify factors that are invariant to spatial orientation while keeping the U-CNN decoder robust across diverse initial states.

\paragraph{Latent-dimension sweep}
The minimal-conditioning principle predicts $d^*{=}2$ for FN-RD, since $(k,\beta)$ are the two factors that vary across systems. To probe both sides of the elbow, we train the hierarchical NODnet with $d \in \{0, 1, 2, 3, 4\}$ over three independent seeds $\{42, 5678, 9012\}$ under identical training settings (same encoder for $d{\geq}1$, same hierarchical decoder with $\{1,3,9\}$-step modules, same MSE-on-rollout objective, same $\sim$650 epochs of data-augmented training); the $d{=}0$ run drops the encoder entirely, recovering the naive single-system neural-operator baseline that ignores per-instance variation. Numerical values are in Table~\ref{tab:supp_dsweep_dr2d}. Applying the two-stage rule from the main-text Methods. \textit{Stage~1.} Seed-mean Train MSE values $30.87, 6.92, 5.26, 4.74, 5.06$ ($\times 10^{-5}$) for $d=0,\ldots,4$ give step-wise relative changes $-78\%, -24\%, -10\%, +7\%$. The $d{=}3{\to}d{=}4$ change of $+0.32$ is comparable to the across-seed envelope ($\sigma_s = 0.35$ at $d{=}4$), and the $d{=}2{\to}d{=}3$ change of $-0.52$ is at the boundary of the envelope ($\sigma_s = 0.34$ at $d{=}2$, $0.12$ at $d{=}3$); the candidate set is therefore $\{2, 3\}$. \textit{Stage~2.} The trained $d{=}3$ encoder spontaneously collapses to a 2-D submanifold, with PCA on per-instance latent means returning variance ratios $(0.80, 0.20, 0.00)$ --- the third axis is empty to numerical precision (Sec.~\ref{sec:supp_fnrd}, Fig.~\ref{fig:supp_d3_latent_3d}). Stage~2 selects $d^*{=}2$. The failure modes below $d^*$ are graded on the long-rollout extrapolation split ($H{=}50$): $d{=}0$ collapses worst ($L_2$RE $9.31{\pm}2.28$, no encoder so no factor information at all), $d{=}1$ collapses partially ($L_2$RE $7.87{\pm}4.48$, a 1-D latent cannot resolve the $(k,\beta)$ pair), and $d{\geq}2$ are within a factor of $1.5$ of each other with $d{=}3, 4$ marginally lower on Train and OOD-Intra metrics.

\begin{table}[hbtp]
    \centering
    \caption{Latent-dimension sweep on the hierarchical NODnet for FN-RD ($L_2$ relative error $\times 10^{-2}$ at $H{=}1, 5, 50$; last-50-epoch train MSE $\times 10^{-5}$). Cells are reported as $\mu \pm \sigma_w \scriptscriptstyle{\pm \sigma_{s,s}}$, where $\sigma_w$ is within-eval std on the test cases (single-seed) and $\sigma_s$ is across-seed std over $\{42, 5678, 9012\}$. Splits: Train (= ID), Interpolated (= OOD-Intra), Extrapolated (= OOD-Extra). Identical training to the main-text NODnet-Hier row except for the latent dimension $d$; the $d{=}0$ row drops the encoder. \textbf{Bold marks the lowest seed-mean in each column}; the two-stage rule (Sec.~S2.3 paragraph above) selects $d^*{=}2$ as the smallest plateau dimension passing the latent space verification (the $d{=}3$ encoder spontaneously collapses to a 2-D submanifold).}
    \begin{tabular}{cllcccc}
        \toprule
        $d$ & Params & Subset & 1-step & 5-step & 50-step & Train MSE \\
        \midrule
        0 & 2.518M & Train         & $1.31{\pm}0.09\ensuremath{\,{\scriptscriptstyle\pm 0.01_s}}$ & $1.78{\pm}0.09\ensuremath{\,{\scriptscriptstyle\pm 0.03_s}}$ & $7.33{\pm}1.62\ensuremath{\,{\scriptscriptstyle\pm 0.21_s}}$ & $30.87\ensuremath{\,{\scriptscriptstyle\pm 2.73_s}}$ \\
          &        & Interpolated  & $1.28{\pm}0.06\ensuremath{\,{\scriptscriptstyle\pm 0.01_s}}$ & $1.75{\pm}0.07\ensuremath{\,{\scriptscriptstyle\pm 0.03_s}}$ & $7.05{\pm}1.22\ensuremath{\,{\scriptscriptstyle\pm 0.26_s}}$ & --- \\
          &        & Extrapolated  & $1.38{\pm}0.11\ensuremath{\,{\scriptscriptstyle\pm 0.02_s}}$ & $1.96{\pm}0.16\ensuremath{\,{\scriptscriptstyle\pm 0.04_s}}$ & $9.31{\pm}2.27\ensuremath{\,{\scriptscriptstyle\pm 0.21_s}}$ & --- \\
        \midrule
        1 & 3.108M & Train         & $0.94{\pm}0.05\ensuremath{\,{\scriptscriptstyle\pm 0.03_s}}$ & $1.25{\pm}0.09\ensuremath{\,{\scriptscriptstyle\pm 0.10_s}}$ & $2.98{\pm}1.24\ensuremath{\,{\scriptscriptstyle\pm 1.42_s}}$ & $6.92\ensuremath{\,{\scriptscriptstyle\pm 0.60_s}}$ \\
          &        & Interpolated  & $0.94{\pm}0.05\ensuremath{\,{\scriptscriptstyle\pm 0.03_s}}$ & $1.29{\pm}0.11\ensuremath{\,{\scriptscriptstyle\pm 0.13_s}}$ & $4.46{\pm}1.61\ensuremath{\,{\scriptscriptstyle\pm 1.22_s}}$ & --- \\
          &        & Extrapolated  & $1.13{\pm}0.18\ensuremath{\,{\scriptscriptstyle\pm 0.07_s}}$ & $1.76{\pm}0.46\ensuremath{\,{\scriptscriptstyle\pm 0.11_s}}$ & $7.87{\pm}4.48\ensuremath{\,{\scriptscriptstyle\pm 1.20_s}}$ & --- \\
        \midrule
        2 & 3.108M & Train         & $0.91{\pm}0.05\ensuremath{\,{\scriptscriptstyle\pm 0.02_s}}$ & $1.20{\pm}0.09\ensuremath{\,{\scriptscriptstyle\pm 0.04_s}}$ & $2.52{\pm}0.66\ensuremath{\,{\scriptscriptstyle\pm 0.07_s}}$ & $5.26\ensuremath{\,{\scriptscriptstyle\pm 0.34_s}}$ \\
          &        & Interpolated  & $0.90{\pm}0.03\ensuremath{\,{\scriptscriptstyle\pm 0.02_s}}$ & $1.19{\pm}0.07\ensuremath{\,{\scriptscriptstyle\pm 0.05_s}}$ & $3.05{\pm}0.91\ensuremath{\,{\scriptscriptstyle\pm 0.21_s}}$ & --- \\
          &        & Extrapolated  & $1.03{\pm}0.16\ensuremath{\,{\scriptscriptstyle\pm 0.10_s}}$ & $1.61{\pm}0.50\ensuremath{\,{\scriptscriptstyle\pm 0.43_s}}$ & $5.49{\pm}3.34\ensuremath{\,{\scriptscriptstyle\pm 2.84_s}}$ & --- \\
        \midrule
        3 & 3.109M & Train         & $0.89{\pm}0.05\ensuremath{\,{\scriptscriptstyle\pm 0.02_s}}$ & $\mathbf{1.13{\pm}0.09}\ensuremath{\,{\scriptscriptstyle\pm 0.03_s}}$ & $\mathbf{2.05{\pm}0.49}\ensuremath{\,{\scriptscriptstyle\pm 0.07_s}}$ & $\mathbf{4.74}\ensuremath{\,{\scriptscriptstyle\pm 0.12_s}}$ \\
          &        & Interpolated  & $0.87{\pm}0.03\ensuremath{\,{\scriptscriptstyle\pm 0.01_s}}$ & $\mathbf{1.13{\pm}0.06}\ensuremath{\,{\scriptscriptstyle\pm 0.03_s}}$ & $\mathbf{2.50{\pm}0.68}\ensuremath{\,{\scriptscriptstyle\pm 0.28_s}}$ & --- \\
          &        & Extrapolated  & $1.01{\pm}0.12\ensuremath{\,{\scriptscriptstyle\pm 0.07_s}}$ & $1.48{\pm}0.32\ensuremath{\,{\scriptscriptstyle\pm 0.25_s}}$ & $4.96{\pm}2.83\ensuremath{\,{\scriptscriptstyle\pm 2.59_s}}$ & --- \\
        \midrule
        4 & 3.110M & Train         & $\mathbf{0.88{\pm}0.04}\ensuremath{\,{\scriptscriptstyle\pm 0.01_s}}$ & $1.15{\pm}0.08\ensuremath{\,{\scriptscriptstyle\pm 0.01_s}}$ & $2.55{\pm}0.72\ensuremath{\,{\scriptscriptstyle\pm 0.24_s}}$ & $5.06\ensuremath{\,{\scriptscriptstyle\pm 0.35_s}}$ \\
          &        & Interpolated  & $\mathbf{0.87{\pm}0.03}\ensuremath{\,{\scriptscriptstyle\pm 0.01_s}}$ & $1.16{\pm}0.06\ensuremath{\,{\scriptscriptstyle\pm 0.01_s}}$ & $3.15{\pm}0.77\ensuremath{\,{\scriptscriptstyle\pm 0.27_s}}$ & --- \\
          &        & Extrapolated  & $\mathbf{0.98{\pm}0.15}\ensuremath{\,{\scriptscriptstyle\pm 0.05_s}}$ & $\mathbf{1.38{\pm}0.34}\ensuremath{\,{\scriptscriptstyle\pm 0.15_s}}$ & $\mathbf{4.13{\pm}2.38}\ensuremath{\,{\scriptscriptstyle\pm 1.54_s}}$ & --- \\
        \bottomrule
    \end{tabular}
    \label{tab:supp_dsweep_dr2d}
\end{table}

\paragraph{The marginal $d{=}3$ advantage}
Table~\ref{tab:supp_dsweep_dr2d} shows that $d{=}3$ is marginally lower than $d{=}2$ on every long-rollout split. The principled gauge above already identifies $d^*{=}2$ as the smallest plateau dimension; three observations explain why $d{=}3$ remains marginally lower without contradicting that gauge.

\textit{(i) The plateau dimensions show only marginal differences.} Seed-mean Train MSE at $d{=}2,3,4$ ($5.26, 4.74, 5.06 \times 10^{-5}$) varies by $\sim 10\%$ across $d$, smaller than the $d{=}1{\to}d{=}2$ elbow drop ($\sim 24\%$). On long-horizon test, the 50-step interpolation seed-means $3.05$ ($d{=}2$) and $2.50$ ($d{=}3$) overlap within seed envelope ($\sigma_s = 0.21$ at $d{=}2$, $0.28$ at $d{=}3$). The marginal $d{=}3$ advantage is one realisation of a flat plateau.

\textit{(ii) Optimization slack at $d{=}3$.} The $(k,\beta){\to}c$ map is genuinely curved (quadratic $\succ$ affine for the zero-shot diffeomorphism, Table~\ref{tab:zeroshot}). A 2-D latent must embed that curve exactly, while a 3-D ambient gives the optimizer a smoother landscape to find the same 2-D submanifold; $d{=}3$ buys optimization slack at no representational cost.

\textit{(iii) Structural slack appears past the plateau, at $d{=}4$.} Train MSE worsens from $4.74$ ($d{=}3$) to $5.06$ ($d{=}4$), the first upward step in the sweep, and the 50-step interpolation column tracks the same trend ($2.50 \to 3.15$). The shift is small in absolute terms because the $10\times$ O(2) data augmentation acts as an additional regularizer that suppresses the structural-slack signature within $d{\leq}3$, but the direction is consistent with the Burgers'-style ``test surges past sufficient capacity'' pattern.

\textit{Empirical confirmation that the $d{=}3$ representation is intrinsically 2-D.} Figure~\ref{fig:supp_d3_latent_3d} renders the per-system means of the trained $d{=}3$ model in its 3-D ambient space. PCA on the 25 system-mean latents returns singular values $\sigma = (0.4003, 0.1996, 0.0022)$, with $\sigma_3$ two orders of magnitude below $\sigma_1$, and the corresponding variance ratios $\sigma_i^2 / \sum_j \sigma_j^2 = (0.8009, 0.1991, \mathbf{0.0000})$, with the third entry below numerical resolution. Within numerical noise the $d{=}3$ model has discovered a strictly 2-D submanifold inside its 3-D latent space. This is the cleanest possible reconciliation with the principle: the model sees three available dimensions, but uses two --- exactly $d^*$ --- with the third dimension carrying no signal. The marginal performance gain of $d{=}3$ over $d{=}2$ is the optimization-slack effect described in (ii), not a representational one.

\begin{figure}[hbtp]
    \centering
    \includegraphics[width=\linewidth]{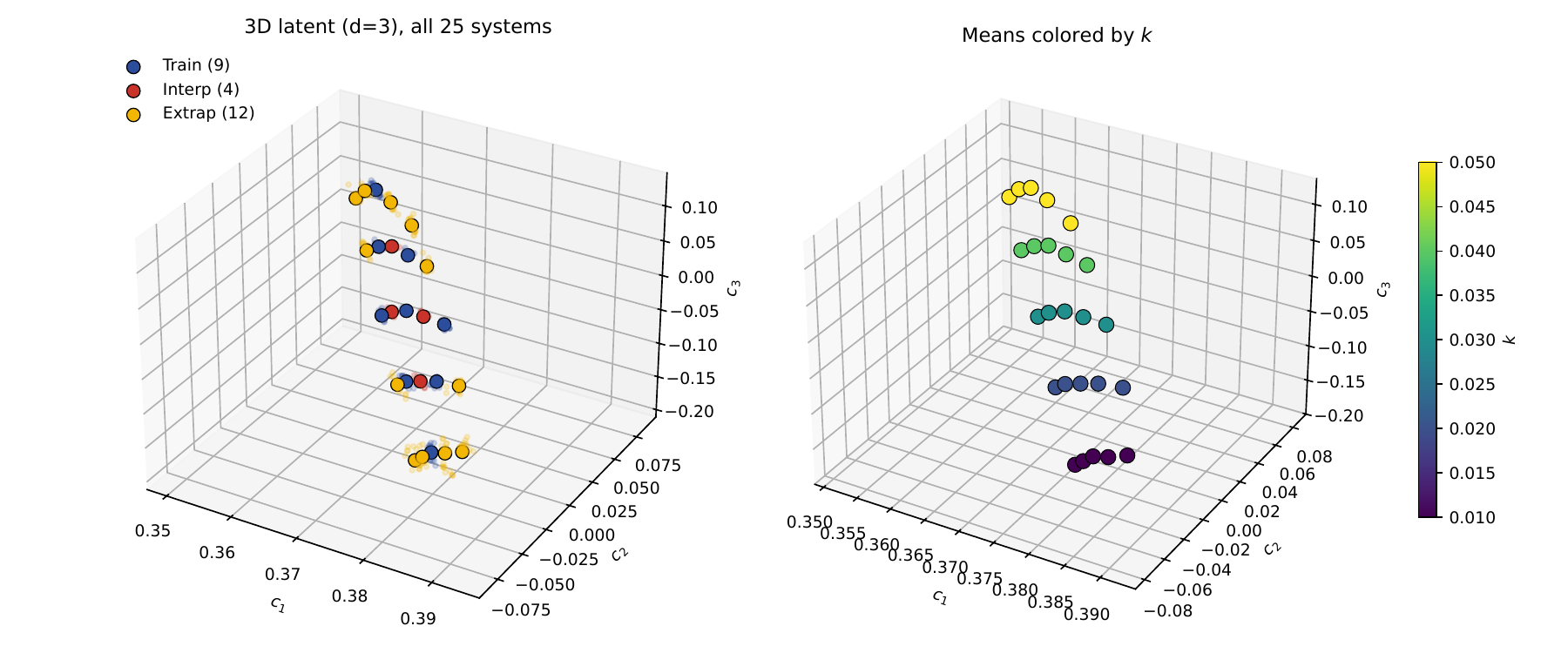}
    \caption{$d{=}3$ latent representation of all 25 FN-RD systems. \textbf{Left:} 3-D scatter in latent space, coloured by split (blue: 9 train SIs, red: 4 interpolated SIs, orange: 12 extrapolated SIs). Faint markers are 20 init+augmentation draws per system, solid markers are per-system means. \textbf{Right:} per-system means coloured by $k$, showing a smooth monotonic factor sweep along the manifold. PCA on the 25 means yields variance ratios $(0.8009, 0.1991, 0.0000)$: the third dimension is empty up to numerical noise, so the trained $d{=}3$ model has spontaneously embedded a 2-D submanifold (exactly $d^*{=}2$) inside its 3-D ambient.}
    \label{fig:supp_d3_latent_3d}
\end{figure}

\paragraph{Zero-shot generalization: effect of diffeomorphism order}
The zero-shot mode described in the main text uses a fitted polynomial diffeomorphism to map factors directly to latents for unseen systems. The fitted quadratic diffeomorphism used in the main text takes the explicit form
\begin{equation}
    \begin{bmatrix}
    c_x \\
    c_y
    \end{bmatrix}
    =
    \begin{bmatrix}
    -0.08 & -3.76 & 0.75 & 12.75 & -4.49 & -0.75\\
    -0.22 & 2.01 & 0.57 & -1.53 & 0.68 & -0.78
    \end{bmatrix}
    \begin{bmatrix}
    1\\
    k \\
    \beta \\
    k^2 \\
    k\beta \\
    \beta^2
    \end{bmatrix}
    \label{eq:DiffusionReaction-diffeo}
\end{equation}
fitted by least squares on the 9 training SIs.

To assess whether the nonlinear structure of this map is essential, we compare polynomial diffeomorphisms of increasing order, all fitted on the same 9 training SIs via least-squares:

\begin{itemize}
    \item \textbf{Order 1 (affine):} Features $[1,\,k,\,\beta]$ (3 coefficients per latent dimension).
    \item \textbf{Order 2 (quadratic):} Features $[1,\,k,\,\beta,\,k^2,\,k\beta,\,\beta^2]$ (6 coefficients; Eq.~\ref{eq:DiffusionReaction-diffeo} above).
    \item \textbf{Order 3 (cubic):} Features up to $k^3,\,k^2\beta,\,k\beta^2,\,\beta^3$ (10 coefficients---underdetermined with 9 training points, so ridge regularization is applied).
    \item \textbf{One-shot:} The encoder $f_\phi$ maps a single test trajectory to the latent space. No factor values are required; included as a reference baseline.
\end{itemize}

We evaluate all four approaches on the 16 held-out FN-RD systems (4 interpolated, 12 extrapolated), measuring $L_2$ relative error over 50-step rollouts (Table~\ref{tab:zeroshot}).

\begin{table}[hbtp]
    \centering
    \caption{Zero-shot generalization on FN-RD ($L_2$ relative error, 50-step rollout) as a function of diffeomorphism polynomial order. One-shot is included as a non-zero-shot reference.}
    \begin{tabular}{lccc}
        \toprule
        Method & Interpolated & Extrapolated & Overall \\
        \midrule
        Order 1 (affine)    & $0.0165{\pm}0.0047$ & $0.0219{\pm}0.0080$ & $0.0205{\pm}0.0077$ \\
        Order 2 (quadratic) & $0.0151{\pm}0.0045$ & $\mathbf{0.0188{\pm}0.0056}$ & $0.0179{\pm}0.0056$ \\
        Order 3 (cubic)     & $\mathbf{0.0146{\pm}0.0043}$ & $0.0189{\pm}0.0054$ & $\mathbf{0.0178{\pm}0.0055}$ \\
        \midrule
        One-shot            & $0.0285{\pm}0.0106$ & $0.0239{\pm}0.0073$ & $0.0250{\pm}0.0085$ \\
        \bottomrule
    \end{tabular}
    \label{tab:zeroshot}
\end{table}

Results are summarized in Table~\ref{tab:zeroshot} and visualized in main-text Fig.~4d. The affine map already achieves reasonable zero-shot accuracy, but the quadratic diffeomorphism reduces overall error by 13\% ($0.0179$ vs.\ $0.0205$), confirming that the nonlinear cross-terms $k^2,\,k\beta,\,\beta^2$ capture meaningful structure in the factor-to-latent mapping. Adding cubic terms yields negligible further improvement ($0.0178$), and the fit becomes underdetermined with only 9 training points, suggesting that the quadratic form already captures the essential geometry. The one-shot baseline ($0.0250$) is notably worse, with the gap attributable to the generalization limit of the encoder when applied to out-of-distribution systems.

We further visualize the decoder output when sampling latents on a uniform grid spanning the training-set latent space (Fig.~\ref{fig:zeroshot_grid}). The decoder produces physically plausible dynamics at every sampled location, confirming that the learned manifold supports continuous, smooth generation of valid PDE solutions across the entire latent space.

\begin{figure}[htbp]
    \centering
    \includegraphics[width=\linewidth]{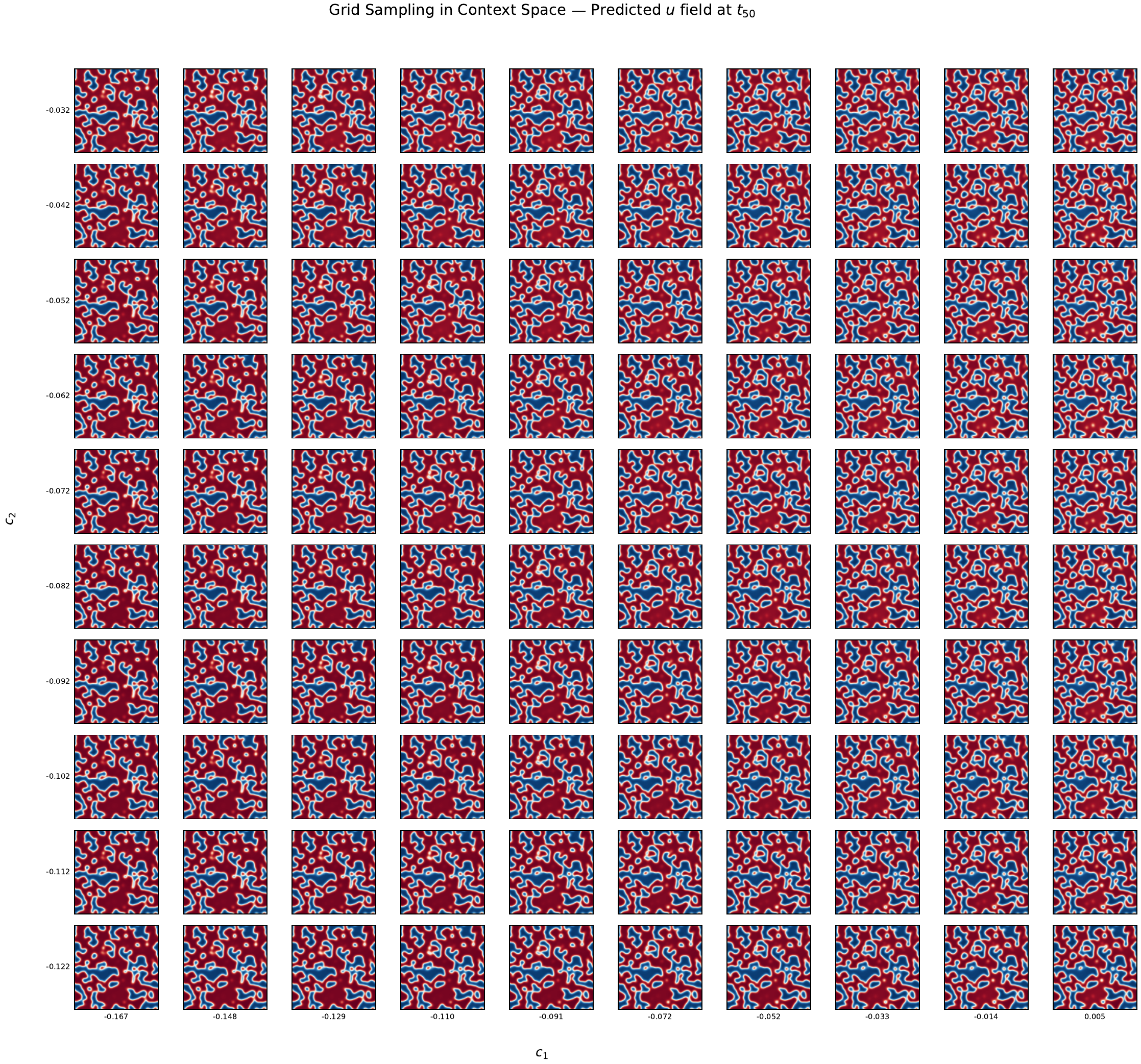}
    \caption{Grid sampling in latent space. Each cell shows the predicted FN-RD field at $t{=}50$ for a uniformly sampled latent $\mathbf{c}$, demonstrating that the decoder generates physically plausible dynamics across the entire learned manifold.}
    \label{fig:zeroshot_grid}
\end{figure}

\subsection{Parameter-supervised reference (PNO) performance}\label{sec:supp_pno_reference}

The parameter-supervised neural operator (PNO) receives the true governing factor of each system instance as an explicit input and shares the decoder architecture of NODnet within each case. PNO is not a peer baseline: it relies on factor labels that the NOD setting does not provide. Its role is to set a reference for the prediction error attainable when the per-instance factor is known, against which the gap incurred by factor inference can be read off. Calibration details (e.g., the $\sqrt{\nu}$ transform used for the Burgers' case) are reported in Section~\ref{sec:supp_burgers}.

\begin{table}[!htbp]
    \centering
    \caption{\textbf{Parameter-supervised reference (PNO) performance.} $L_2$ relative error ($\times 10^{-2}$, mean$\pm$std) at 1-, 5-, and 50-step rollout horizons across the same splits used in main-text Table~1. The PNO row receives the true governing factor as input; values are single-seed.}
    \begin{tabular}{lcccc}
        \toprule
        Model & Subset & 1-step & 5-step & 50-step \\
        \midrule
        \multicolumn{5}{l}{\textit{Burgers' equation}} \\
        \midrule
        PNO-DeepONet$^{\uparrow}$ (5.0M)  & ID            & $3.89{\pm}1.74$ & $6.04{\pm}2.58$ & $13.43{\pm}6.39$ \\
                                          & OOD           & $4.03{\pm}1.76$ & $5.79{\pm}2.43$ & $12.85{\pm}6.20$ \\
                                          & OOD-inviscid  & $4.19{\pm}1.19$ & $7.68{\pm}3.12$ & $16.90{\pm}5.70$ \\
        \midrule
        \multicolumn{5}{l}{\textit{Cylinder flow wake}} \\
        \midrule
        PNO$^{\uparrow}$ (4.5M)           & ID            & $3.65{\pm}1.51$ & $12.27{\pm}5.53$ & $36.52{\pm}12.26$ \\
                                          & OOD           & $2.02{\pm}1.51$ & $5.01{\pm}3.71$  & $28.11{\pm}13.58$ \\
        \midrule
        \multicolumn{5}{l}{\textit{FitzHugh--Nagumo reaction--diffusion}} \\
        \midrule
        PNO-Hier$^{\uparrow}$ (2.5M)      & ID        & $0.77{\pm}0.06$ & $1.16{\pm}0.09$ & $1.83{\pm}0.28$ \\
                                          & OOD-Intra & $0.74{\pm}0.04$ & $1.10{\pm}0.06$ & $1.68{\pm}0.16$ \\
                                          & OOD-Extra & $0.76{\pm}0.08$ & $1.23{\pm}0.14$ & $2.33{\pm}0.55$ \\
        \bottomrule
    \end{tabular}
    \label{tab:supp_pno}
\end{table}

%%%%%%%%%%%%%%%%%%
\section{Extended Discussion}\label{sec:extended_discussion}

This section gathers extended discussion in four groups: empirical validation of design choices (\textsection~\ref{sec:supp_tds} TDS ablation, \textsection~\ref{sec:supp_efficiency} efficiency vs.\ iMODE); theoretical positioning (\textsection~\ref{sec:supp_perspectives} alternative perspectives on the diffeomorphism, \textsection~\ref{sec:supp_identifiability} connections to identifiability theory); comparisons with alternative paradigms (\textsection~\ref{sec:supp_foundation_comparison} foundation-model surrogates, \textsection~\ref{sec:supp_sr_comparison} symbolic regression); and cross-case synthesis (\textsection~\ref{sec:supp_structural_slack} structural slack), closing with a practitioner's checklist (\textsection~\ref{sec:supp_workflow}).

\subsection{Ablation on Trajectory-Decoupled Sampling}\label{sec:supp_tds}

Trajectory-Decoupled Sampling (TDS) ensures that the conditioning trajectory $\mathbf{x}_{0:t}$ (fed to the encoder) and the prediction trajectory $\mathbf{x}'_{0:s}$ (predicted by the decoder) are randomly sampled from different trajectories of the same system instance. Table~\ref{tab:tds-comparison} compares convergence speed and final test-set accuracy with and without TDS across three cases.

\begin{table*}[t]
\centering
\caption{Ablation on Trajectory-Decoupled Sampling. Model convergence is evaluated on the test dataset at 1-step prediction. All pairs share the same optimizer and learning rate schedule. CFW uses unseen systems as the test set; the ablation retains the SupCL regularization.}
\begin{tabular}{lcccccc}
\toprule
\multicolumn{1}{c}{} & \multicolumn{3}{c}{\textbf{w/ TDS}} & \multicolumn{3}{c}{\textbf{w/o TDS}} \\
\cmidrule(lr){2-4} \cmidrule(lr){5-7}
\textbf{Case} & Epochs & MSE & L2RE\% & Epochs & MSE & L2RE\% \\
\midrule
1D Burgers' Equation & 1.5e4 & 2.85e-4 & 0.56 & 3e4 & 4.71e-4 & 0.92 \\
FN-RD & 67 & 3.02e-5 & 0.92 & 208 & 6.02e-5 & 1.96 \\
CFW & 422 & 0.101 & 1.60 & 976 & 0.490 & 5.61 \\
\bottomrule
\end{tabular}
\label{tab:tds-comparison}
\end{table*}

TDS consistently accelerates convergence (2--3$\times$ fewer epochs) and improves final test-set accuracy across all cases. Without TDS, the encoder attempts to encode trajectory-specific transient information (e.g., initial conditions, temporal phase) into the latent $c$, producing noisy optimization signals and prolonging training. TDS prevents this by forcing the encoder to extract only the invariant governing dynamics shared across trajectories of the same SI. The resulting improvement in generalization (lower L2RE on held-out data) confirms that TDS acts as an implicit regularizer against overfitting to trajectory-level features.

\subsection{Efficiency comparison between NODnet and iMODE}\label{sec:supp_efficiency}
We substitute the neural-ODE process in iMODE with a fixed-interval predictor $f$ for like-for-like comparison with \ngs{} in computational complexity. The iMODE formulation is
\begin{align}
    \textrm{Prediction}&: \hat{\mathbf{x}}_{t+l;\,i} = f_{\theta}^{(l)}(\mathbf{x}_t, \eta_i^{(j)})\\
    \textrm{Loss}&: \mathcal{L}_{i}(\theta, \eta_i^{(j)}) = \frac{1}{k}\sum_{h=1}^k \mathrm{MSE}(\hat{\mathbf{x}}_{t+h;\,i}, \mathbf{x}_{t+h;\,i})\\
    \textrm{Initialization}&: \eta^{(0)}_i = \eta^*,\quad \eta_i \equiv \eta_i^{(p)} \\
    \textrm{Inner update}&: \eta_i^{(j+1)} = \eta_i^{(j)} - \alpha\,\nabla_{\eta}\, \mathcal{L}_{i}(\theta, \eta_i^{(j)})\\
    \textrm{Outer update}&: \theta \leftarrow \theta - \frac{\beta}{N} \sum_{i=1}^N \nabla_{\theta}\, \mathcal{L}_i(\theta, \eta_i)
\end{align}
where $p$ is the number of inner update steps and $m$ is the number of recurrent neural-network evaluations. The NODnet formulation is
\begin{align}
    \textrm{Latent}&: c_i = f_{\phi}(\mathbf{x}_{s:s+m;\,i}) = o\bigl(\Phi^{(m+1)}_{\phi}(\mathbf{x}_{s:s+m;\,i})\bigr)\\
    \textrm{Prediction}&: \hat{\mathbf{x}}_{t+l;\,i} = g_{\theta}^{(l)}(\mathbf{x}_t, c_i)\\
    \textrm{Loss}&: \mathcal{L}_i = \frac{1}{k}\sum_{h=1}^k \mathrm{MSE}(\hat{\mathbf{x}}_{t+h;\,i}, \mathbf{x}_{t+h;\,i}) \\
    \textrm{Update encoder}&: \phi \leftarrow \phi - \frac{\alpha}{N}\sum_{i=1}^N \nabla_{\phi}\, \mathcal{L}_i(\phi, \theta)\\
    \textrm{Update decoder}&: \theta \leftarrow \theta - \frac{\alpha}{N}\sum_{i=1}^N \nabla_{\theta}\, \mathcal{L}_i(\phi, \theta)
\end{align}
In practice, neural-network backpropagation takes $\beta$ times longer than the forward pass (typically $\beta > 1$) because of intermediate-state storage. The computational complexities of iMODE and \ngs{} are
\begin{equation}
    \begin{split}
        \mathcal{O}_{\text{iMODE}} &= N(p+1)(\beta+1)\,k\,\mathcal{O}_{\theta},\\
        \mathcal{O}_{\text{NODnet}} &= N(\beta+1)\bigl(m\,\mathcal{O}_{\phi} + k\,\mathcal{O}_{\theta}\bigr),
    \end{split}
    \label{compare}
\end{equation}
giving the ratio
\begin{equation}
    \frac{\mathcal{O}_{\text{iMODE}}}{\mathcal{O}_{\text{NODnet}}} = \frac{k(p+1)\,\mathcal{O}_{\theta}}{m\,\mathcal{O}_{\phi} + k\,\mathcal{O}_{\theta}}.
\end{equation}

We compare the training of iMODE and \ngs{} on the bistable system case where $k=m=p=5$. The training costs for both methods are shown in Table~\ref{table:iMODEcomparison}. iMODE consumes approximately the same amount of memory as \ngs{} because only first-order gradient information is stored. However, the repetitive inner loop optimizations end up with 3 times computational complexity than \ngs{}. 

\begin{table*}[hbtp]
    \centering
    \caption{Efficiency comparison between \ngs{} and iMODE, on an RTX Ada 6000 NVIDIA GPU. Note that time* is simply calculated as Epoch $\times$ Computation time per epoch for reference.}
    \begin{tabularx}{\textwidth}{ >{\centering\arraybackslash}X 
   >{\centering\arraybackslash}X 
   >{\centering\arraybackslash}X }
        \toprule
        Metrics & NODnet & iMODE \\
        \midrule
        Computation time per epoch & 8.4 s & 24.7 s \\
        % \hline
        GRAM per batch & 2360 MB & 2852 MB \\
        % \hline
        Epoch reaching error 1e-3 & 31 & 12\\
        % \hline
        Time* reaching error 1e-3 & 260.4 s & 296.4 s \\
        % \hline
        Epoch reaching error 1e-4 & 800 & 500 \\
        % \hline
        Time* reaching error 1e-4 & 6720 s & 12350 s\\
        \bottomrule
    \end{tabularx}
    \label{table:iMODEcomparison}
\end{table*}

\subsection{Alternative Perspectives for Diffeomorphic Latent Space}\label{sec:supp_perspectives}

In this section, we offer alternative perspectives to understand the diffeomorphic structure of latent space.

\paragraph{GAN perspective}
The original generative adversarial network (GAN) solves the following min-max optimisation problem,
\begin{equation}
     \min_\mathbf{G}\max_\mathbf{D} V(\mathbf{D},\mathbf{G})=\mathbb{E}_{\mathbf{x}} [\log \mathbf{D}(\mathbf{x})] + \mathbb{E}_{\mathbf{z}} [\log(1-\mathbf{D}(\mathbf{G}(\mathbf{z})))],
\end{equation}
where $\mathbf{x}\sim p_{\textrm{data}}$, $\mathbf{z}\sim p_{\mathbf{z}}$, $\mathbf{D}$ is a discriminator optimised to distinguish real samples $\mathbf{x}$ from synthetic ones $\mathbf{G}(\mathbf{z})$, and $\mathbf{G}$ is a generator that produces synthetic samples from latent input $\mathbf{z}$ and seeks to maximise $\mathbf{D}$'s evaluation.

\ngs{} admits an analogous reading: the encoder $f_\phi$ serves as the generator (producing latents from conditioning trajectories) and the decoder $g_\theta$ serves as the discriminator (using the latent to predict the next state and being scored by prediction error). For interpretation, in each training epoch we manually separate the latents into two sets, $c = \{c_x, c_z\}$, where $c_x$ acts as the real factors up to a diffeomorphism in the current epoch and $c_z$ are the synthetic ones. The NODnet discriminator (the decoder $g_\theta$) is optimised by
\begin{equation}
   \max_{g_\theta}\, -\textbf{MSE}\bigl(g_\theta(\mathbf{x}_t, c_x), \mathbf{x}_{t+1}\bigr) \equiv \min_{g_\theta}\, \textbf{MSE}\bigl(g_\theta(\mathbf{x}_t, c_x), \mathbf{x}_{t+1}\bigr).
\end{equation}
The NODnet generator (the encoder $f_\phi$) is optimised by
\begin{equation}
    \min_{f_\phi}\, \textbf{MSE}\bigl(g_\theta(\mathbf{x}_t, c_z), \mathbf{x}_{t+1}\bigr) = \textbf{MSE}\bigl(g_\theta\bigl(\mathbf{x}_t, f_\phi(\mathbf{x}_{s:s+m;\,z})\bigr), \mathbf{x}_{t+1}\bigr).
\end{equation}
$f_\phi$ and $g_\theta$ are simultaneously optimised by
\begin{equation}\label{eq:GANtotal}
\begin{aligned}
    \min_{f_\phi}\max_{g_\theta} \;&\bigl[-\textbf{MSE}\bigl(g_\theta(\mathbf{x}_t, c_x), \mathbf{x}_{t+1}\bigr) + \textbf{MSE}\bigl(g_\theta\bigl(\mathbf{x}_t, f_\phi(\mathbf{x}_{s:s+m;\,z})\bigr), \mathbf{x}_{t+1}\bigr)\bigr] \\
    \equiv \min_{f_\phi,\, g_\theta} \;&\textbf{MSE}\bigl(g_\theta\bigl(\mathbf{x}_t, f_\phi(\mathbf{x}_{s:s+m})\bigr), \mathbf{x}_{t+1}\bigr).
\end{aligned}
\end{equation}
In words, $g_\theta$ is optimised for prediction accuracy given the reference latents $c_x$, while $f_\phi$ is optimised so that the inferred latents $c_z = f_\phi(\mathbf{x}_{s:s+m;\,z})$ produce predictions consistent with $g_\theta$. Equation~\eqref{eq:GANtotal} gives a GAN-inspired interpretation of the coupled encoder--decoder training objective, rather than a formal equivalence to adversarial training.

\paragraph{Partial-ordering perspective}
To further explain the ordering in the latent space, we formulate:
\begin{equation}
    \mathcal{S}(\mathbf{x}_{t}\mid p_i)=g_\theta(\mathbf{x}_{t}\mid c_i) = \mathbf{x}_{t+1}
\end{equation}
where $\mathcal{S}$ is the true physics solution operator. In classical physics, $\mathcal{S}$ is generally continuous and differentiable. Since neural operator $g_\theta$ is usually also continuous and differentiable, it follows that:
\begin{equation}
    c_i = g^{-1}_\theta\mathcal{S}(p_i),
\end{equation}
where the operator $g^{-1}_\theta\mathcal{S}(\cdot)$ is also continuous and differentiable. 
For any triplet of factors $(p_i, p_j, p_k) \in \mathcal{P}$, continuity of the mapping suggests that if $p_j$ lies between $p_i$ and $p_k$, the discovered latent $c_j$ should take a corresponding intermediate position between $c_i$ and $c_k$. This behavior is related to Brouwer's Invariance of Domain~\cite{spanier2012algebraic}: under the additional assumptions of continuity, injectivity, and matched manifold dimension, the mapping preserves neighborhood structure. Trajectory-decoupled sampling encourages the encoder $f_\phi$ to assign a consistent latent to each system instance; when this empirical map is approximately injective and smooth, the observed latent space $\mathcal{C}$ behaves as a smooth image of the factor manifold $\mathcal{P}$.

\subsection{Connections to identifiability theory in latent-variable models}\label{sec:supp_identifiability}

NOD's recovery of per-instance factors from heterogeneous trajectories sits adjacent to several lines of identifiability theory in latent-variable models. The comparison clarifies what NOD assumes and does not assume, and why TDS is the natural training objective in this regime.

\paragraph{Nonlinear ICA with auxiliary variables.}
Khemakhem et al.~\cite{khemakhem2020variational} establish that nonlinear ICA is identifiable up to component-wise transformations when the latents $z$ are conditionally distributed given an observed auxiliary variable $u$ and $p(z\mid u)$ lies in a sufficiently rich exponential family; the auxiliary $u$ breaks the rotational symmetry that otherwise renders nonlinear ICA non-identifiable. NOD's instance grouping (knowing which trajectories belong to the same system instance) supplies exactly such an auxiliary: each grouping label $u_i$ indexes a distinct conditional $p(z\mid u_i)$, and $p(z\mid u_i)\neq p(z\mid u_j)$ whenever the underlying factors differ ($p_i\neq p_j$). The formal subtlety is that this conditional is degenerate (a point mass at the deterministic factor $p_i$) rather than a continuous parametric family. NOD therefore lies in the limit where Khemakhem's auxiliary-variable identifiability degenerates to deterministic-factor identifiability, with the symmetry-breaking role preserved and the identification enforced through TDS rather than through likelihood maximisation over a parametric conditional.

\paragraph{Disentanglement under weak supervision.}
Locatello et al.~\cite{locatello2019challenging} prove that fully unsupervised disentanglement is non-identifiable: for any generative model, an alternative model with entangled latents is observationally equivalent. NOD escapes this negative result because instance grouping provides weak supervision. The closer reference is Locatello et al.~\cite{locatello2020weakly}: under pair sampling where two observations share $K$ of $d$ generative factors, the latent is identifiable up to component-wise reparametrization. NOD's TDS instantiates this with $K=d$ (the conditioning and supervision trajectories share all factors, only initial conditions differ), the maximally-shared case, giving identifiability up to permutation and sign of the latent coordinates without further structural assumptions.

\paragraph{Temporal structure as auxiliary.}
Hyv{\"a}rinen et al.~\cite{hyvarinen2019nonlinear} use the time index as an auxiliary variable for nonlinear ICA: segments at different time points have different conditional distributions over the latents, and the resulting contrast identifies them. Klindt et al.~\cite{klindt2021towards} extend this to nonlinear disentanglement of natural data via temporal sparse coding. NOD's TDS uses the same kind of structural index to break symmetry but swaps the contrast axis from time-segment to instance: the auxiliary indexes which system instance generated the trajectory, not which time window within one trajectory. The resulting latent identifies the per-instance factor rather than the per-segment temporal mode, and the contrast is across the instance ensemble rather than within a single trajectory.

\paragraph{What NODnet adds beyond these.}
Existing identifiability results constrain the latent representation but assume the latent dimension $d$ is given. NODnet's minimal conditioning principle adds dimension identification: $d^*$ is gauged from data as the smallest value at which prediction loss saturates. The recovered $d^*$ matches the number of physically independent governing factors in every case (Table~\ref{tab:supp_seven_cases}), giving an identifiability claim that is both factor-wise (latent coordinates correspond to factors) and dimension-wise (the number of recovered factors is correct). The decoder side is also distinct: $g_\theta$ is a full neural operator advancing states under continuous-time dynamics, not a static generative model, so identification is enforced through forward dynamical consistency rather than reconstruction of static observations. The identifiability claim NODnet supports is therefore: under instance grouping (auxiliary in the Khemakhem sense, weak supervision in the Locatello~\cite{locatello2020weakly} sense), TDS pairing with $K=d$ shared factors, and dimension gauging at $d^*$, the recovered latent is unique up to permutation, sign, and component-wise reparametrization, with the dimension of the recovered representation matching the dimension of the governing factor space.

\paragraph{From extrapolation to extrapolation range.}
Identifiability ensures the latent-factor map is unique on the training support; it does not specify how far the map remains valid beyond the training hull. NODnet's case studies exhibit useful extrapolation in every instance (Burgers' crosses the inviscid singularity $\nu{=}0$, CFW generalizes outside the trained Reynolds-number range, FN-RD generalizes to factor pairs outside the training $(k,\beta)$ box). Characterizing the \emph{radius} of stability of the diffeomorphism, the maximum factor distance from the training hull at which one-shot or zero-shot inference still recovers the correct dynamics, is one step beyond identifiability and is not formally addressed here. Empirically, generalization fails when the test factor leaves the smooth latent-factor manifold (e.g.\ at bifurcation thresholds or topology changes in the dynamics); a formal theory linking the recovered Jacobian condition number and the loss landscape's local curvature to the resulting stability radius would close this gap.

\subsection{Foundation-model surrogates: scope of comparison}\label{sec:supp_foundation_comparison}

Foundation-model-style surrogates~\cite{cao2024vicon,hao2024dpot,chen2025flow,mccabe2023multiple} predict the next state from a long window of past states, with the conditioning produced inside a single forward pass through a large transformer-class decoder. Two independent design axes contribute to their reported accuracy: (i) the conditioning structure, an unstructured past-state window with no per-instance latent, and (ii) the decoder capacity, a foundation-scale backbone trained across many system families.

A direct head-to-head between NODnet and a full foundation surrogate (MPP, DPOT, ViCon, Multi-Physics Pretraining) would conflate these two axes: a larger decoder typically wins regardless of conditioning structure, so a side-by-side test would not isolate the contribution of the minimal conditioning principle. We therefore use ANO$^{\downarrow}$ as the controlled foundation-style baseline at matched per-case decoder capacity, with the conditioning channel replaced by the same window-of-past-states structure that foundation surrogates use (given history, predict next). The ablation between NODnet and ANO$^{\downarrow}$ thereby measures the conditioning-structure effect with decoder capacity held constant; the comparison against PNO$^{\uparrow}$ (true factors as input) and SNO ($d{=}0$, no encoder) on the same axis completes the conditioning sweep from oracle to absent.

NODnet's encoder--decoder split is itself architecture-agnostic: the foundation-scale decoders cited above could be substituted for the per-case backbone in $g_\theta$, with the minimal conditioning channel preserved between $f_\phi$ and $g_\theta$ (a foundation-scale $g_\theta$ with a $d^*$-dim latent on the side). Such a hybrid is a natural extension complementary to foundation-scale training, not a competitor. The principle prescribes how the conditioning channel is structured, not how the decoder is scaled; holding decoder capacity fixed in the present comparison ensures that the gap between NODnet and ANO$^{\downarrow}$ is the conditioning-structure effect rather than a backbone-scale effect.

\subsection{Comparison with symbolic regression}\label{sec:supp_sr_comparison}

Symbolic regression (SR) and NODnet address overlapping problems but sit at different points on the expressivity--interpretability frontier; the comparison clarifies when each is the appropriate tool.

\paragraph{What each method returns.}
SR returns a closed-form symbolic equation, typically a sparse combination of hand-designed primitives such as polynomials, partial derivatives, and trigonometric functions~\cite{udrescu2020ai,brunton2016discovering,cranmer2023interpretable}. NODnet returns a learned operator pair $(f_\phi, g_\theta)$ together with a $d^*$-dimensional latent space empirically diffeomorphic to the per-instance factor space; the form of the underlying law is not recovered.

\paragraph{Scope of dynamics.}
SR's expressivity is bounded by its primitive library. Systems whose governing equations involve primitives outside the library, or whose effective dynamics are not sparsely representable, fall outside SR's reach. The cylinder-flow wake on an unstructured 32k-node mesh is one such regime: the spatial discretization, the boundary geometry, and the multi-scale wake structure jointly resist sparse symbolic representation. NODnet's neural-operator decoder has no analogous bound and recovers a usable surrogate in this regime (main-text Fig.~3, Table~1). The same scope distinction applies to experimentally observed systems where the underlying physics is itself partially unknown: NODnet requires only trajectories.

\paragraph{Family recovery.}
Modern SR variants extend the original sparse-regression framework to families with varying coefficients, e.g.\ parametric SINDy~\cite{rudy2019data}. When the parametric dependence is itself sparse and library-expressible, these variants recover the family. NODnet recovers operator and per-instance factor jointly in one training run, without requiring the parametric form to be analytically clean; the gauged $d^*$ identifies the correct number of factors directly from data.

\paragraph{Inference cost.}
A symbolic equation must be numerically integrated to make predictions, with cost set by the discretization required for the recovered equation (CFL-bounded $\Delta t$, fine spatial grids, stiffness-dependent solver choice). For 1D PDEs this is modest (milliseconds to seconds per trajectory); for 2D PDEs the cost reaches seconds to minutes on CPU; 3D pushes higher. NODnet inference is a forward pass of $g_\theta$ rolled forward from an initial state, costing milliseconds on GPU for the cases here. The practical speedup ranges from $\sim$10$^2\times$ for 1D systems to $\sim$10$^4$--$10^5\times$ for 2D problems with fine resolution.

\paragraph{What SR offers that NODnet does not.}
A recovered closed-form law generalises beyond the training distribution to arbitrary initial and boundary conditions, composes naturally with other equations in multi-physics settings, and provides physical insight from the equation form itself. NODnet generalises only as far as its training distribution and the diffeomorphism extends, and its interpretability is at the factor-structure level rather than equation-structure level.

\paragraph{When to pick which.}
SR is appropriate when the dynamics are within library expressivity, a closed-form law is the deliverable, and the inference budget tolerates numerical integration. NODnet is appropriate when the dynamics are too complex for library expressivity (e.g.\ the cylinder wake), only trajectory data is available (experimental settings), or inference must be fast. The two are complementary rather than substitutes.

\subsection{Structural slack: cross-case synthesis}\label{sec:supp_structural_slack}

The minimal conditioning principle predicts that test error worsens above $d^*$ because the encoder's spare dimensions absorb initial-condition-specific signal that the decoder then propagates as drift through the rollout. The three case studies provide complementary evidence of this structural-slack mechanism, with the strength of the signal modulated by data availability and augmentation.

\paragraph{Burgers' (S2.1).}
Seed-mean Train MSE rises from $1.01$ at $d{=}1$ to $1.42$ at $d{=}2$ (Table~\ref{tab:supp_dsweep}), and the 50-step rollout error rises from $\sim 17.6\%$ at $d{=}1$ to $\sim 21.7\%$ at $d{=}2$ on ID, with $d{=}2$ the worst plateau dimension. The encoder's extra dimensions absorb shock-front micro-variation that does not depend on the governing factor $\nu$, and the autoregressive loss amplifies this absorption. The $d{=}3, 4$ rows return closer to the $d{=}1$ envelope because at higher dimensions the slack scatters across multiple axes rather than concentrating in one, an effect also seen on CFW (next paragraph).

\paragraph{Cylinder flow wake (S2.2).}
The clearest geometric demonstration. PCA on the trained $d{=}2$ encoder's per-system mean latents returns variance ratios $(0.739, 0.261)$: the second axis is populated, not empty (Fig.~\ref{fig:supp_d2_latents_cfw}), and what populates it is within-system conditioning-window variability rather than $\nu$. The non-monotonic $d{>}1$ test error in Table~\ref{tab:supp_dsweep_cfw} is the rollout consequence: $d{=}2$ is worst because the slack is concentrated in one axis; $d{=}3,4$ partially recover by spreading the slack across multiple axes (PC3 contribution $0.083$, PC2 dropping from $0.261$ to $0.231$).

\paragraph{FN-RD (S2.3).}
The 50-step interpolation column shows a soft surge at $d{=}4$: seed-means $4.46 \to 3.05 \to 2.50 \to 3.15$ as $d$ grows from $1$ to $4$, with Train MSE worsening from $4.74$ at $d{=}3$ to $5.06$ at $d{=}4$. The surge is delayed by one step relative to Burgers' and smaller in magnitude because the $10\times$ O(2) data augmentation acts as an additional regularizer, suppressing the slack effect within $d{\leq}3$ and partially absorbing it at $d{=}4$.

\paragraph{The $d{=}0$ anchor.}
At long horizons in the cylinder wake the $d{=}0$ baseline (no encoder, no slack to fill) is competitive with the entire $d{>}1$ curve (Table~\ref{tab:supp_dsweep_cfw}, $H{=}50$ on OOD: $d{=}0{=}34.5\%$, $d{=}2{=}38.3\%$, $d{=}3{=}34.5\%$, $d{=}4{=}33.0\%$, all within seed envelope of one another and of $d^*{=}1$ at $31.0\%$). ``No encoder'' is therefore not strictly dominated by ``encoder with slack'', direct empirical evidence that slack rather than absent factor information is the dominant source of $d{>}d^*$ rollout error: once the rollout has time to amplify the IC absorption, the bottleneck-free baseline catches up because it lacks the slack channel that drives drift.

\subsection{Applying NODnet to a new dataset}\label{sec:supp_workflow}

This subsection condenses the workflow for applying NODnet to a new heterogeneous-trajectory dataset, drawing on the seven reported cases.

\paragraph{Data.}
Organise as $\mathcal{D} = \{\mathcal{T}_{ij}\}$ with $i$ indexing the system instance and $j$ indexing trajectories within an instance. NODnet requires (i) at least two trajectories per instance under distinct initial conditions (for trajectory-decoupled sampling), and (ii) trajectories long enough that a conditioning window $\mathbf{x}_{0:t}$ captures factor-bearing dynamics. The instance grouping is the only supervision NODnet uses; per-instance factor labels are not required at any stage.

\paragraph{Backbone selection.}
Choose $f_\phi$ and $g_\theta$ to match the observation structure: CNN/U-CNN for regular grids, Point Transformer for irregular meshes, recurrent (LSTM/GRU) for time series, FNO for resolution-agnostic recovery, Neural ODE for differential decoders. Table~\ref{tab:supp_seven_cases} maps observation structure to backbone choice across the seven reported cases. The principle is architecture-agnostic; matching the backbone to the data is an engineering choice.

\paragraph{Dimension gauging.}
Sweep $d \in \{0, 1, \ldots, d_{\max}\}$ with $d_{\max}$ a few above the suspected factor count, training each $d$ over $\geq 3$ independent seeds under identical settings (same encoder for $d{\geq}1$, same decoder, same loss, same epoch budget); the $d{=}0$ run drops the encoder, recovering the SNO baseline. Apply the two-stage rule (Methods). \textit{Stage~1.} On a long-horizon training-side gauge --- Train MSE for autoregressive cases, Train $H{=}50$ $L_2$ relative error for cases where short-horizon Train MSE is overfit by the encoder-free baseline (CFW) --- identify the smallest $d$ at which the next-step relative change falls within the across-seed envelope, and admit it together with the next dimension as the candidate set. \textit{Stage~2.} Inspect PCA variance ratios on per-instance latent means at the candidate dimensions: select the smallest $d^*$ whose $d{+}1$ run either populates a spurious slack axis or collapses to a $d^*$-dimensional submanifold. Sections S2.1, S2.2, S2.3 give worked examples. Three seeds per $d$ sufficed across all reported cases, and the runs are independent so they parallelise across GPUs.

\paragraph{Trajectory-decoupled sampling.}
At each training step, sample the conditioning trajectory $\mathbf{x}_{0:t}$ from one initial condition of instance $i$ and the supervision trajectory $\mathbf{x}'_{0:s}$ from a different initial condition of the same instance. TDS integrates as a sampler choice in the data loader and adds no extra forward passes; the ablation in S3.1 shows 2--3$\times$ faster convergence and lower test error than naive single-trajectory sampling.

\paragraph{Diagnostics after training at $d^*$.}
Three checks confirm the principle has taken effect: (i) per-instance latent means $\bar{c}_i$ versus a factor proxy should form a smooth, monotonic curve (the empirical diffeomorphism signature, Figs.~\ref{fig:supp_d1_latents_cfw}, \ref{fig:supp_d3_latent_3d}); (ii) PCA on the per-instance latents should return rank matching $d^*$ to numerical noise, with a non-empty principal component beyond $d^*$ signalling structural slack; (iii) one-shot inference on held-out instances should yield a latent consistent with the manifold geometry.

\paragraph{Failure modes.}
(i) \emph{Encoder collapse}: the latent maps all instances to the same value. Causes include short conditioning windows, supervision-geometry mismatch (Burgers' NODnet-AR is the canonical case, Section S2.1), or insufficient instance diversity; diagnose by plotting per-instance latents; remedy by lengthening the conditioning window, adopting a full-field decoder, or adding instance diversity. (ii) \emph{Latent degeneration with sparse instances}: with very few instances ($\lesssim 16$), prediction loss alone may not separate the latents enough; add a manifold regularizer (the supervised-contrastive loss in S2.2 CFW handles 16 instances). (iii) \emph{Discrete or topologically nontrivial factor spaces}: NODnet's diffeomorphism premise assumes continuous factor variation; bifurcating regimes and topology-changing dynamics lie outside the present scope.

% \section{Robustness to measurement noise}
% \subsection{Measures for noise enhancement}
% Taking the bistable system as an example

% \subsection{Brief results from other systems}

% \vspace{2em}
% \noindent \textbf{\Large Supplemental movie}
% \begin{enumerate}
%     \item .
% \end{enumerate}

\bibliographystyle{sn-aps}
\bibliography{../Refs}

%%%%%%%%%%%%%%%%%%%%%%%%%%%%%%%%%%%%%%%for recycling
\begin{comment}

\end{comment}

\endgroup
\end{document}